\documentclass[sn-basic,iicol]{sn-jnl}

\usepackage{hyperref}
\usepackage{url}
\usepackage{amssymb}       
\usepackage{algpseudocode}
\usepackage{algorithm}
\usepackage{amsfonts,amsthm}       
\usepackage{graphicx}
\usepackage{subfigure}
\usepackage{natbib}

\usepackage{enumitem}
\usepackage{mathtools}
\usepackage{csquotes}
\usepackage{wrapfig}
\usepackage{placeins} 
\usepackage{subcaption}
\usepackage{textcase}
\usepackage{comment}
\usepackage{xcolor}

\newcommand{\Vect}{\text{LoCE}}
\newcommand{\Vects}{\text{LoCEs}}
\newcommand{\locVect}{\text{LoCE}}
\newcommand{\locVects}{\text{LoCEs}}
\newcommand{\globVect}{\ensuremath{\text{GloCE}}}
\newcommand{\globVects}{\ensuremath{\text{GloCEs}}}
\newcommand{\subglobVect}{\ensuremath{\text{SGloCE}}}
\newcommand{\subglobVects}{\ensuremath{\text{SGloCEs}}}
\newcommand{\multiLocVect}{\text{ML-LoCE}}
\newcommand{\multiLocVects}{\text{ML-LoCEs}}

\newcommand*{\Reals}{\mathbb{R}}
\DeclareMathOperator*{\argmin}{arg\,min}
\newcommand*{\Concept}{\ensuremath{\mathbf{c}}}
\newcommand*{\ConceptClass}{\mathbf{C}}
\newcommand*{\ConceptTerm}[1]{\ensuremath{\text{\texttt{\small #1}}}}
\newcommand*{\vectOptimLoss}{\mathcal{L}}

\newcommand*{\shortparagraph}[1]{\par\noindent\textbf{#1.}~}

\newcommand{\Result}[1]{\textbf{\textcolor{blue!50!black}{#1}}}

\newcommand{\appendixsection}[2]{
    \section*{Appendix #1. #2}  
    \addcontentsline{toc}{section}{Appendix #1. #2}  
    \renewcommand{\thetable}{#1\arabic{table}}  
    \renewcommand{\thefigure}{#1\arabic{figure}}  
    \setcounter{table}{0}  
    \setcounter{figure}{0}  
}

\usepackage{tabularx}
\usepackage{multirow}
\usepackage{float}
\usepackage{morefloats}
\usepackage{subcaption}

\usepackage[capitalize]{cleveref}
\crefname{section}{Sec.}{Sec.}
\Crefname{section}{Section}{Sections}
\crefname{subsection}{Sec.}{Sec.}
\Crefname{subsection}{Section}{Sections}
\crefname{table}{Tab.}{Tab.}
\Crefname{table}{Table}{Tables}

\usepackage[algo2e,boxed,vlined,ruled]{algorithm2e}

\AtBeginDocument{\PassOptionsToPackage{authoryear,round}{natbib}}
\let\cite\citep

\theoremstyle{definition}
\newtheorem{definition}{Definition}
\newtheorem{assumption}[definition]{Assumption}

\BeforeBeginEnvironment{definition}{\par\medskip}
\AfterEndEnvironment{definition}{\par\medskip}
\BeforeBeginEnvironment{assumption}{\par\medskip}
\AfterEndEnvironment{assumption}{\par\medskip}

\usepackage{aliascnt}

\begin{document}


\title[Local Concept Embeddings]{Local Concept Embeddings for Analysis of Concept Distributions in Vision DNN Feature Spaces}


\author*[1,2]{\fnm{Georgii} \sur{Mikriukov}}
\email{georgii.mikriukov@continental.com}

\author[3]{\fnm{Gesina} \sur{Schwalbe}}
\email{gesina.schwalbe@uni-luebeck.de}

\author[1]{\fnm{Korinna} \sur{Bade}}
\email{korinna.bade@hs-anhalt.de}

\affil*[1]{\orgname{Hochschule Anhalt}, \orgaddress{\country{Germany}}}

\affil[2]{\orgname{Continental AG}, \orgaddress{\country{Germany}}}

\affil[3]{\orgname{University of Lübeck}, \orgaddress{\country{Germany}}}





\abstract{
Insights into the learned latent representations are imperative for verifying deep neural networks (DNNs) in critical computer vision (CV) tasks. Therefore, state-of-the-art supervised Concept-based eXplainable Artificial Intelligence (C-XAI) methods associate user-defined concepts like ``car'' each with a single vector in the DNN latent space (concept embedding vector). In the case of concept segmentation, these linearly separate between activation map pixels belonging to a concept and those belonging to background. Existing methods for concept segmentation, however, fall short of capturing implicitly learned sub-concepts (e.g., the DNN might split car into ``proximate car'' and ``distant car''), and overlap of user-defined concepts (e.g., between ``bus'' and ``truck''). In other words, they do not capture the full distribution of concept representatives in latent space. For the first time, this work shows that these simplifications are frequently broken and that distribution information can be particularly useful for understanding DNN-learned notions of sub-concepts, concept confusion, and concept outliers. To allow exploration of learned concept distributions, we propose a novel local concept analysis framework. Instead of optimizing a single global concept vector on the complete dataset, it generates a local concept embedding (LoCE) vector for each individual sample. We use the distribution formed by LoCEs to explore the latent concept distribution by fitting Gaussian mixture models (GMMs), hierarchical clustering, and concept-level information retrieval and outlier detection. Despite its context sensitivity, our method's concept segmentation performance is competitive to global baselines. Analysis results are obtained on three datasets and six diverse vision DNN architectures, including vision transformers (ViTs). The code is available at \url{https://github.com/continental/local-concept-embeddings}.
}




\maketitle

\begingroup
\renewcommand\thefootnote{}\footnote{Accepted for publication in the \textit{International Journal of Computer Vision (IJCV)}. This is the author’s accepted manuscript. The final version will be published by SpringerLink. To cite this work, please refer to the final journal version once published.}
\addtocounter{footnote}{-1}
\endgroup

\section{Introduction}
\label{sec:intro}

\begin{figure*}[htb]
  \centering
  \includegraphics[width=\linewidth]{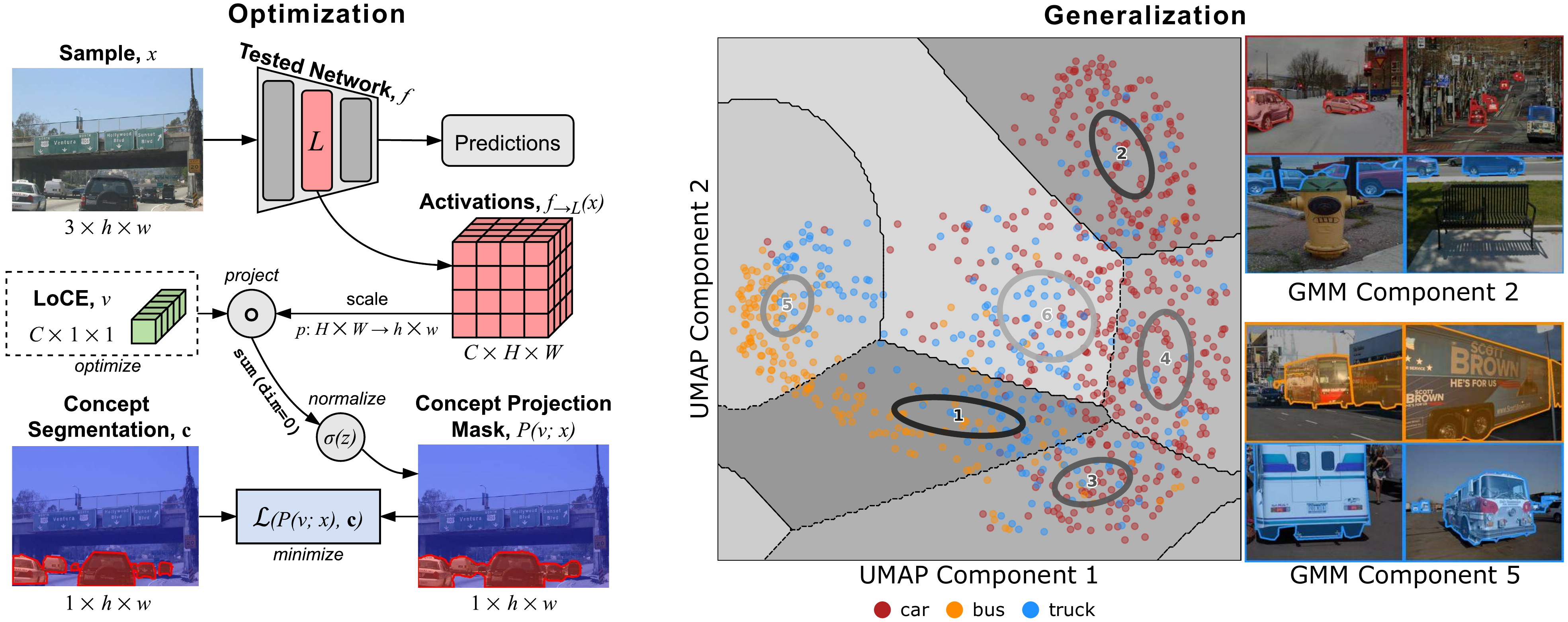}
  \caption{(\textit{left}) \locVect{} optimization for an image-concept pair: \locVect{} $v$ represents the optimal convolutional filter weights that \enquote{project} Sample's $x$ Activations $f_{\to L}(x)$ from layer $L$ into the Concept Projection Mask $P(v;x)$, aiming to reconstruct the target Concept Segmentation $\Concept$ with minimal loss $\vectOptimLoss(P(v;x), \Concept)$. (\textit{right}) Distribution of 2D UMAP-reduced \locVects{} demonstrating the confusion of \ConceptTerm{car}, \ConceptTerm{bus}, and \ConceptTerm{truck} concepts in \texttt{model.encoder.layers.1} of DETR. Gaussian Multinomial Mixture (GMM) is fitted to \locVects{} to highlight the structure. Additionally, some samples from GMM components 2 and 5 are demonstrated.} 
  \label{fig:method-gcpv-optimization}
\end{figure*}


Opaque machine learning models, like convolutional DNNs (CNNs) \cite{cong2023review}
or vision transformers (ViTs) \cite{khan2023survey}, find widespread use in critical domains like automated driving \cite{turay2022performing,fingscheidt2022deep,zhu2024systematic} or medical diagnostics \cite{altaf2019going,xu2024vision}.
Hence, their safety and fairness assessability have become pressing concerns~\cite{longo2024explainable,houben2022inspect,arrieta2020explainable,aihleg2020assessment}, manifested in legal acts~\cite{gdpr,madiega2021artificial} and industrial standards~\cite{iso21448,iso8800,iso5469,iso24028}.
An important ingredient to assessability
\cite{zhang2020testing,houben2022inspect} and
debugging~\cite{adebayo2020debugging} is access to the model's learned
behavior, such as feature
importance~\cite{%
  lundberg2017unified,%
  ribeiro2016should,%
  liangExplainingBlackboxModel2021%
}
and logical structure \cite{%
  hailesilassie2016rule,
  rabold2018explaining,
  hohman2020summit
},
and to its \emph{learned knowledge}, such as learned features \cite{%
  olah2017feature,
  kim2018interpretability,
  fong2018net2vec,
  fel2023craft,
  posada2022eclad
}.
In case intrinsically interpretable models~\cite{rudin2019stop} fail to be applicable, this learned knowledge is opaquely encoded in the DNN feature spaces, and requires post-hoc explainability techniques for further insights.

\paragraph*{C-XAI}
For verification of CV DNNs, the atomic notion to specify requirements are \emph{(visual semantic) concepts} \cite{schwalbe2022concept,poeta_concept-based_2023,our_excv_paper}: These are human-understandable attributes from natural language that are applicable to images or image regions, such as object classes like \ConceptTerm{bus} or \ConceptTerm{truck} \cite{bau2017network}.
Thus, one crucial question for verification is:
\emph{Does and how does a model represent information about a user-defined concept?}
The sub-field of post-hoc supervised C-XAI tackles this question \cite{schwalbe2022concept,poeta_concept-based_2023,lee_neural_2024,our_excv_paper}.
The distributed nature of DNN representations, however,  makes it non-straight-forward to provide an answer. Neurons respectively filters in DNNs are typically polysemantic, i.e., correspond to more than one semantic concept, and a concept is distributed over several neurons/filters \cite{dreyer2024pure,bricken2023monosemanticity,bau2017network}.
%
Respective methods from C-XAI share a core and elegant idea: They specify the concept via labeled example images and represent the DNN-learned information as a classifier on latent space \emph{vectors} that allows to separate between concept and non-concept examples \cite{%
  bau2017network,
  fong2018net2vec,
  kim2018interpretability,
  lucieri2020explaining,
  graziani2020concept,
  crabbe2022concept
}.
As downstream applications, these classifiers can then, e.g., be analyzed offline to find spurious concept similarities \cite{fong2018net2vec,schwalbe2021verification} or applied online during inference to identify spurious or faulty sightings of concepts \cite{schwalbe2022enabling}.

Amongst such classifiers, linear models are preferable for explainability purposes \cite{kim2018interpretability}. They associate a concept with a (single) weight vector (so-called \emph{concept (embedding) vector}) in latent space, pointing in the direction of the concept.
Prominent and well-received examples are the initial approach NetDissect by \citet{bau2017network}, and its successors Net2Vec by \citet{fong2018net2vec} and TCAV by \citet{kim2018interpretability}. TCAV employs only image-level concept classification. Instead, the others \emph{segment} concepts by classifying each activation map pixel as belonging to a concept or the background using filter instead of neuron weights. This also allows analysis of object concepts in object detection DNNs \cite{schwalbe2021verification,mikriukov2023evaluating}.

\paragraph*{C-XAI Limitations: What about Distributions?}
Existing methods for post-hoc concept segmentation well answer \emph{whether} information about a concept was learned. However, they share a crucial shortcoming when it comes to \emph{how} the concept is represented: They represent concepts as single vectors, e.g., as done by \citet{bau2017network,fong2018net2vec,kim2018interpretability}; or as strictly separable regions in latent space as pursued by \citet{crabbe2022concept} in TCAR.
This may fail to capture the complexity of concept representation in state-of-the-art CV latent spaces in at least two points, as illustrated in \cref{fig:method-gcpv-optimization} \textit{(right)}:
(1) A DNN may not learn a concept like \ConceptTerm{car} sharply (= representable by a single vector). As pointed out by \citet{crabbe2022concept}, it may learn several \textbf{sub-concepts} like \ConceptTerm{proximate car} and \ConceptTerm{distant car} that only together (roughly) constitute the human-interpretable pendant (cf.\ examples in \cref{fig:subconcepts}).
(2) But even if relaxing the linear to a more general separation model as done by \citet{crabbe2022concept}, the strict separability condition might be broken by \textbf{overlap} of learned representations for user-defined concepts like that of \ConceptTerm{car} and \ConceptTerm{truck} (cf.\ examples in \cref{fig:confusion-detr}).

This results in the following claim that is here investigated in detail for the first time:
\smallskip
\begin{description}[font=\normalfont\bfseries, leftmargin=1em]
\item[Hypothesis.]~\\
  \emph{Typical DNN latent space structures are complex and consistently break the simplifying assumptions of existing supervised post-hoc concept segmentation approaches, missing out on valuable information about sub-concepts and concept confusion.}
\end{description}
\smallskip
In other words, the \emph{distribution} of information about concepts in latent space is more complex than existing C-XAI methods can capture with their single-vector and separable-region representation models. This challenge has only recently been formulated by \citet{our_excv_paper}, and has, to our knowledge, not been tackled so far.

\paragraph*{Approach: From Single Points to Distributions}
We here, for the first time, employ a shift of perspective
to alleviate these issues and complement existing C-XAI tooling. Instead of single vectors or sharp regions, we aim to represent concepts as \emph{distributions} of latent space vectors. These distributions may both be (1) multi-modal (accounting for sub-concepts), as well as (2) overlapping (accounting for concept confusion).
Additionally, they allow a more principled way to identify concept examples that are outliers for the DNN.
Global methods like Net2Vec would natively give rise to a distribution over concept training subsets by training a concept vector each. However, linear models are low-variance classifiers. Already training datasets as small as 50--100 samples will average out smaller subconcepts (details in \autoref{sec:appendix-b}).

Thus, we here suggest to change from the \enquote{global} approach to a \enquote{local} one:
Instead of optimizing a single concept vector on the \emph{complete} concept dataset (global), we optimize one vector for \emph{each small subset} of concept segmentation
labels (local).
To maximally reduce the risk of averaging out subconcepts without changing the training data distribution, we consider subsets of size 1. That way, a local concept embedding vector (\locVect{}) encodes the concept-against-background separation features that are local to a single image, see~\cref{fig:method-gcpv-optimization} (\textit{left}).
Further, we generalize the information about \locVect{} distribution to get valuable insights on representation learning in the feature space, like concept outlier detection, subconcept discovery, or concept confusion identification, see~\cref{fig:method-gcpv-optimization} (\textit{right}).

\paragraph*{Why image-wise \locVects{}?}
The loss of subconcepts is inherent to global methods, and image-wise \locVects{} are a natural first step to avoid this with minimal changes.
Note that single images may still contain more than one concept instance and, thus, several subconcepts. A tempting alternative to \emph{image}-wise \locVects{} could, therefore, be \emph{instance}-wise \locVects{}, which eliminates subconcept mixing. As a drawback, they would require costly instance segmentation labels and data augmentation to avoid distorting the background distribution (cf.\,\cref{fig:subconcepts}). 
The image-wise approach used here maintains the comparability to global methods and their low computational costs (find details in \cref{sec:method-complexity}). Here, we show that despite subconcepts co-occurring on images, our approach unravels subconcepts sufficiently to be detected.
Another noteworthy property of image-wise \locVects{} is that they can capture---and thus allow to analyze---background bias, other than way more costly background-randomized alternatives \cite{mikriukov2023evaluating}.
Also, the high model bias of the linear \locVect{} models avoids overfitting to their small training set of few activation map pixels.
This makes image-wise \locVects{} a suitable first step towards assessing concept representation distribution.


\paragraph*{Contributions}
Our key contributions are:
\begin{enumerate}[label=(\arabic*), nosep]
    \item A novel local post-hoc supervised \textbf{feature space analysis technique} for concept segmentation that captures the \textbf{distribution} of learned concept feature information. 
    Our broad ablation and baseline comparison study attests \textbf{valid results} of the extracted information that is \textbf{competitive with existing supervised global C-XAI techniques for concept segmentation}.
    \item For the first time, a detailed analysis of concept distributions in CV DNN latent spaces: Our diverse experiments show that the \textbf{feature information distribution for semantic concepts is far more complex than previously addressed in the literature}. Specifically, it exhibits subconcepts and concept overlap.
    \item Detailed demonstration of applicability to several debugging problems, notably
    \begin{itemize}[nosep]
        \item learned \textbf{sub-concept} identification,
        \item \textbf{concept confusion} measurement, and
        \item concept-level \textbf{outlier detection} and \textbf{information retrieval}.
    \end{itemize}
    \item A thorough discussion of potential future research directions leveraging the distribution perspective.
\end{enumerate}





\paragraph*{Outline}
After a discussion of related work (\cref{sec:related}), \cref{sec:method-optimization} introduces our method to obtain the local \locVects{}.
\cref{sec:method-generalization} then details several techniques how these can be used to analyze concept feature distributions and even recover a global concept vector.
Using the experimental setup detailed in \cref{sec:setup}, we first conduct an ablation study investigating influences on the performance of \locVects{}. Then, we show that typical concept distributions are complex but can be captured by our method, even competitive with respect to concept segmentation baselines (\cref{sec:experiments}). After that, we demonstrate how interesting insights can be obtained for model debugging (\cref{sec:applications}). We finalize with a discussion on future work (\cref{sec:discussion}) and a conclusion (\cref{sec:conclusion}).

\section{Related Work}
\label{sec:related}

\subsection{XAI}
The field of explainable artificial intelligence (XAI) is vast by now \cite{schwalbe2023comprehensive,vilone2020explainable,guidotti_principles_2021,linardatos2021explainable,burkart2021survey}, covering both ante- and post-hoc approaches \cite{linardatos2021explainable,schwalbe2023comprehensive,bodria2023benchmarking}.
While ante-hoc methods like concept bottleneck models~\cite{koh2020concept,wang2021interpretable,chen2020concept,losch2019interpretability,chen2019this} promise intrinsically transparent representations, they also require design- and training-time modifications and often fail to achieve full feature space interpretability \cite{kazhdan2021disentanglement,marconato2022glancenets,hoffmann2021this}.
Thus, we focus on \emph{post-hoc} methods, allowing us to reveal what an already trained black-box model has learned.
Here, \emph{local} methods like feature importance \cite{%
lundberg2017unified,
ribeiro2016why,
petsiuk2018rise,novello2022making,fel2021look, simonyan2013deep,springenberg2014striving,smilkov2017smoothgrad,sundararajan2017axiomatic,
selvaraju2017grad,zhou2016learning}
give insights why a \emph{single} output was obtained. 
In contrast, \emph{global} approaches aim to derive \emph{global} properties or interpretable surrogates \cite{hohman2020summit,zhang2018interpreting,rabold2020expressive} to explain how the model works. Only a few methods generalize local insights to global ones, allowing the assessment of how insights from single-samples are distributed over a dataset, and existing ones are restricted to generalizing feature importance, such as SpRAy~\cite{lapuschkin2019unmasking}, which clusters local feature saliency masks.
Unfortunately, besides fidelity issues~\cite{adebayo2018sanity,ghorbani2019interpretation},
local feature importance methods 
do not uncover the knowledge encoded in latent spaces. 
This lead to the (global) concept analysis methods that are discussed next.

\subsection{Post-hoc C-XAI}
Concept-based methods aim to connect (visual) semantic concepts---i.e., human-interpretable attributes that can be assigned to regions in the input (image) \cite{kim2018interpretability,fong2018net2vec,ghorbani2019towards}---with numerical representations in the feature space.
C-XAI as a subfield of XAI has only recently found further interest, with several specialized survey papers newly published \cite{schwalbe2022concept,poeta_concept-based_2023,lee_neural_2024,our_excv_paper}.
Ante-hoc approaches here insert concept associations by design, e.g., in the form of concept bottlenecks \cite{koh2020concept,chen2019this}
but are criticized for leaking non-semantic information \cite{hoffmann2021this} and require access to the DNN design and training. Due to the latter reason, we here consider post-hoc C-XAI methods.
Amongst these, post-hoc \emph{unsupervised} methods provide qualitative, not necessarily interpretable, insights into the recurring activation patterns and features the model has learned.
They employ clustering of activations~\cite{ghorbani2019towards,ge2021peek} and activation pixels~\cite{posada2022eclad,posada2023scalable}, general matrix factorization~\cite{zhang2021invertible,fel2023craft,leemann2023when} or subspace clustering~\cite{vielhaben2023multidimensional} to the latent space.
%
%
Instead, post-hoc \emph{supervised} methods, as considered here, associate latent space representations to \emph{user-pre-defined} concepts in a dedicated training phase
~\cite{schwalbe2022concept,lee_neural_2024,poeta_concept-based_2023}. 
%
One of the first techniques applied here was NetDissect by \citet{bau2017network}. They investigated the polysemanticity of neurons by finding the most associated concepts for each filter. Successively,
\citet{kim2018interpretability} found that \emph{linear} information encoding, i.e., a associating a concept to a linear combination of neurons, is the most desirable choice for explanations (other than, e.g., invertible neural networks as introduced by \citet{esser2020disentangling}).
The basic idea for linear post-hoc supervised analysis was first and simultaneously introduced with TCAV~\cite{kim2018interpretability} for concept \emph{classification} and Net2Vec~\cite{fong2018net2vec} for concept \emph{segmentation} as considered here.
Later works adhere to this general idea, extending to
regressive concepts \cite{graziani2018regression,graziani2020concept},
larger models like object detectors \cite{schwalbe2021verification,mikriukov2023evaluating},
from single concepts to complete concept sets \cite{yuksekgonul2022posthoc}, 
and from vector representation to feature space regions \cite{crabbe2022concept}.
%
%
To our knowledge, the only approach assigning local (gradient-based) concept vectors was by \citet{zhang2018examining} for concept classification, which was used to obtain a global vector via averaging.
Hence, existing post-hoc concept analysis methods only assign global concept representations and, thus, do not account for the distribution of concepts in latent spaces like disparate clusters of sub-concepts and overlap of different concepts, potentially hiding unexpected anomalies and outliers. Our approach tackles this for the first time, with minimal changes to the baseline of global methods.
The quality of concept representations is typically evaluated via performance of the linear concept models parametrized by the concept vectors. In our case, more accurate concept segmentation reconstructions on a test set mean better concept-to-vector alignment \cite{fong2018net2vec,schwalbe2021verification,lucieri2020explaining}.
As baseline for concept segmentation we chose NetDissect, top-$k$ Net2Vec (set Net2Vec weights of all but top $k$ contributing filters to zero), and classical Net2Vec. The latter still is state-of-the-art in the C-XAI domain \cite{our_excv_paper}, and successors in particular did not target improved concept reconstruction.

\subsection{Image and Outlier Retrieval}
Information retrieval (IR) in the context of image retrieval involves obtaining relevant images from a large database in response to a query~\cite{dubey2021decade}. This process, also known as content-based image retrieval (CBIR), involves representing the query as a vector in a feature space. This vector can be obtained through various methods, including traditional feature extractors like SIFT~\cite{lowe2004distinctive} and HOG~\cite{dalal2005histograms}, as well as CNNs. The latter learn high-dimensional, semantically rich feature embeddings from images, significantly enhancing retrieval accuracy~\cite{dubey2021decade, zhang2023content}. The retrieval process involves comparing the query vector to the vectors of database images using distance metrics such as Euclidean distance or cosine similarity, which quantify the similarity between the query and the images in the database~\cite{radenovic2018fine}.

In addition to retrieving similar images, IR techniques can be adapted for outlier detection. Outliers are data points that deviate significantly from most samples in a dataset~\cite{chalapathy2019anomaly}. Outlier detection is accomplished by analyzing the distribution of distances between a sample and the other samples in the dataset~\cite{chandola2009anomaly,campos2016evaluation}.
The use of DNNs for outlier detection has proven effective, as these networks intrinsically cluster similar images together and make detecting isolated or distant images more efficient~\cite{hendrycks2018anomaly,uhlemeyer2022towards}.

Here, we utilize IR to measure concept confusion and for concept-based outlier detection.

\section{Obtaining Local Concept Embeddings (\locVects)}
\label{sec:method-optimization}

\begin{figure*}[tbh]
  \centering
  \includegraphics[width=\linewidth]{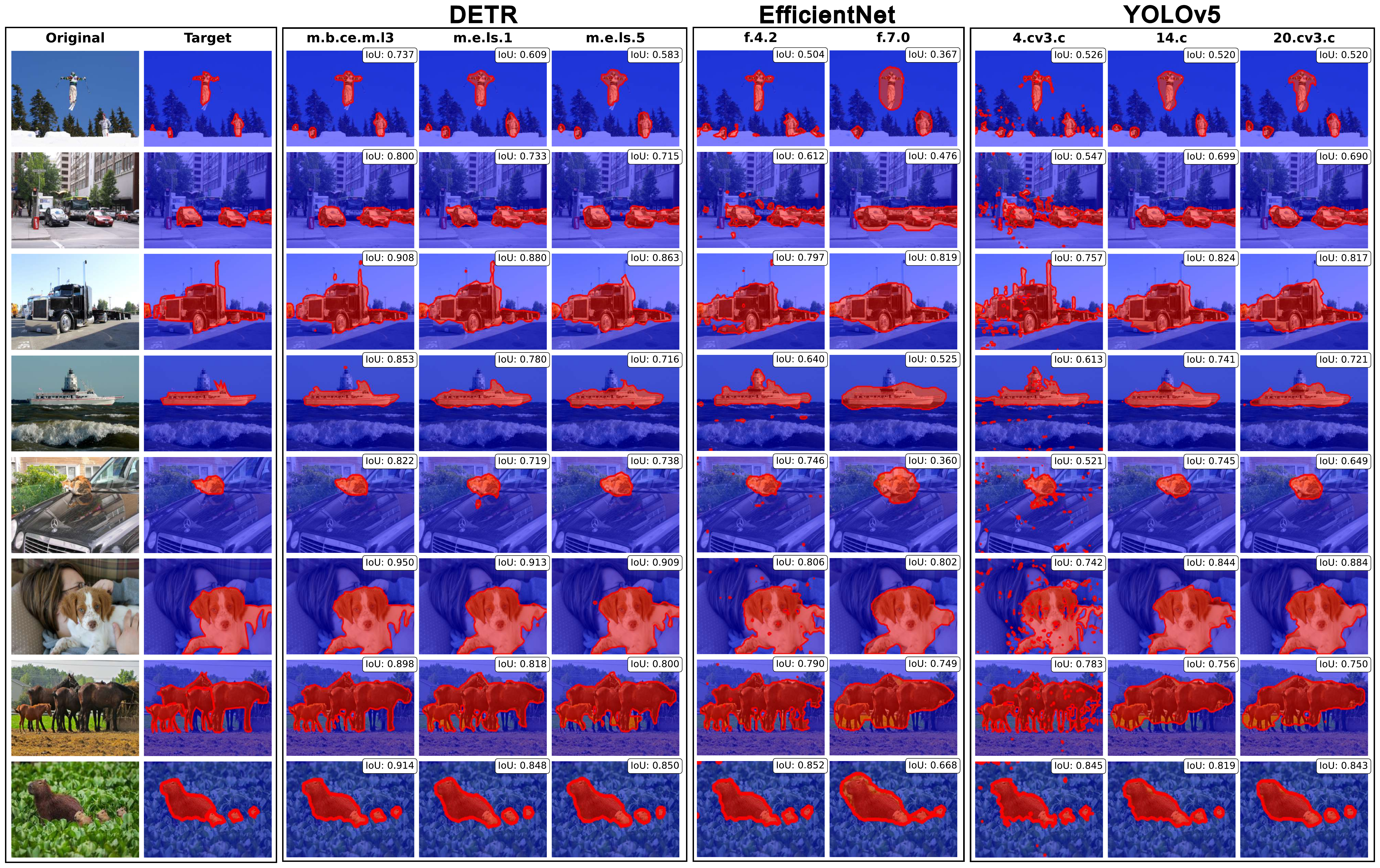}
  \caption{ MS COCO \& Capybara Dataset: Exemplary concept segmentation masks predicted by optimized image-local concept models for different trained DNNs and layers (\emph{columns}), and different original input image and target concept segmentation mask (\emph{rows}). Examples chosen randomly. Find more results in \cref{fig:gcpv-optimization-results-2}.}
  \label{fig:gcpv-optimization-results-1}
\end{figure*}

This section presents the core idea of our framework for local concept-based latent space analysis: how to associate a concept's segmentation mask in a single image with a latent space vector that captures those features which distinguish the concept from its background. 
The output of this step is a set of \locVects{}, i.e., local concept vectors, each representing the features of its concept extracted from \emph{one} image.

Per-image vectors also pose the main difference to global C-XAI methods
~\cite{bau2017network,fong2018net2vec,schwalbe2022enabling}
: These
represent a concept with a single vector for the \emph{entire dataset}—effectively averaging out important distributional details
. Instead, our
local \locVect{} focuses on analyzing the \emph{concept's 
distribution}
in latent space.
By examining 
that distribution
we can discover sub-concepts (as high-density regions), identify cases of concept confusion (as distribution overlaps), and detect outliers where the model exhibits uncertainty (\cref{sec:method-generalization}). While this approach trades off some global fidelity (\cref{sec:experiments}), it provides deeper insights into the concept’s structure and behavior in varying contexts (\cref{sec:applications}).

\textbf{\locVects{} can be viewed and explained as a natural generalization of the Net2Vec}~\cite{fong2018net2vec}.
To maintain comparability between, we keep changes to their training framework minimal, apart from changing from global to local focus, as detailed in this section.
Lastly we discuss implications of single-image \Vects{} on capturing background-bias (\cref{sec:method-context}).

\subsection{Common Idea for Obtaining \Vects{}}
\label{sec:method-idea}
Assume we are given an input image $x$ of height and width $h\times w$ from a training data set $X$, a concept label $\Concept$ as a $h\times w$-sized binary segmentation mask that localizes the concept occurrences in the image, and a layer $L$ of interest of the CNN $f$, in which information about the concept label should be found.
Now we want to find out whether and how information about $\Concept$ is distributed over the channels of the layer's activation maps $f_{\to L}(x)\in\Reals^{C\times H\times W}$ ($C$, $H$, $W$ are channel, height, width dimensions).
Note that each activation map channel can be scaled to represent a non-binary mask for the input using a respective scaling function $p\colon H\times W\to h\times w$. 
We here say the channel's information is projected to the input.
The idea now is: If these scaled channel masks can be used to reconstruct the ground truth concept mask $\Concept$ in a simple manner like linear combination, then (1) the CNN stores information about the concept, (2) the achievable reconstruction quality tells how well it does so, and (3) the reconstruction routine---if comparable---serves as a representation of concept information within the CNN.

While this applies directly to CNNs, ViTs require an additional spatial rearrangement to obtain activation maps, as detailed in \cref{sec:method-transformers}.


\subsection{Optimization}

\paragraph*{Local versus Global Optimization Goals}
The aim thus is to reconstruct the channel masks via \emph{linear combination} of (rescaled) activation channels
(see \cref{fig:method-gcpv-optimization}): Find a vector $v\in\Reals^{C\times 1\times 1}$ of $C$ weights (one for each of the $C$ channels) such that the resulting \emph{concept projection mask}
\begin{align}
    P(v; x) &\coloneqq p(v^T f_{\to L}(x)) 
    \nonumber\\ &= p({\textstyle\sum_{k \leq C}} v_k \cdot f_{\to L}(x)_k) \in \Reals^{h\times w},
    \label{eq:opt-concept-projection}
\end{align}
optimally coincides with the concept label mask $\Concept$ as defined by a loss $\vectOptimLoss(P(v;x), \Concept)$,
which measures the divergence of the masks (details below). 
The optimal vector $v$ 
\begin{align}
    v=\argmin_{v\in\Reals^{C\times 1\times 1}} \sum_{x\in X}\vectOptimLoss(P(v;x), \Concept_x) 
    \label{eq:opt-net2vec-vector}
\end{align}
captures the features that discriminate any occurrence of $\Concept$ in samples from $X$ against all backgrounds occurring in $X$.

\shortparagraph{Global case (Net2Vec)}
Here, $X$ is the complete available concept training dataset. It captures the features that discriminate any instance of $\Concept$ against any context (even if the concept instance would not occur in that kind of context).

\shortparagraph{Local case: Our \locVects{}}
To obtain the local case, we set the training set $X=\{x\}$ to a single image. 
The thus obtained \locVect{} captures the features discriminating the \textbf{single concept instance against its context}:
\begin{align}
    v=\argmin_{v\in\Reals^{C\times 1\times 1}} \vectOptimLoss(P(v;x), \Concept{}). 
    \label{eq:opt-gcpv-vector}
\end{align}
Just as in the global case, this vector can be used to predict a concept segmentation: Calculating the dot product between $v$ and each activation map pixel\footnote{This is equivalent to a $1\times 1$ convolution.} yields a non-binary mask indicating which activation map pixels belong to the concept and which not (cf.~\cref{fig:method-gcpv-optimization}).
This is useful for formulating the loss (see below) and evaluating how much information $v$ captured about the concept label (see metrics of experimental setup in \cref{sec:setup-metrics}).
However, other than in the global case, one cannot expect this local concept model to well generalize to other images. Instead, $v$ captures image-local information, as used, e.g., in image retrieval.

\paragraph*{\locVect{} Loss Formulation}
%

Recall that the activation maps have a different, typically smaller, resolution than the input. So, rescaling is necessary prior to comparing the ground truth mask to the predicted one.
To balance the tradeoff between speed and performance, we rescale (using) the activation maps and segmentations to a resolution of $100\times100$ (for more info, see ablation study results in \cref{sec:experiments-ablation}).
%
Similar to \cite{fong2018net2vec}, before calculating the difference between concept ground truth mask $\Concept$ and \locVect{} projection $P(v; x)$, the projection is normalized using an element-wise sigmoid $\sigma(z) = 1/(1+\exp(-z))\in (0,1)$.
%
Further, we employ the same per-pixel pseudo-BCE loss as Net2Vec\footnote{See \href{https://github.com/ruthcfong/net2vec/blob/master/src/linearprobe_pytorch.py\#L94}{source code} and comments in \cite[p.\,7]{schwalbe2020concept}.} to estimate $\vectOptimLoss(P(v;x), \Concept)$ in a per-sample manner. It penalizes a low agreement between ground truth and predicted concept mask but weighs the negative and positive areas respectively to compensate for the lower occurrence probability of positive mask areas:
\begin{align}
    \vectOptimLoss(P(v;x), \Concept) &
    \nonumber\\
    = \frac{1}{hw}{\textstyle\sum_{i,j}}\big( & \alpha\sigma(P_{i,j})\Concept_{i,j} +
    \nonumber\\
    & (1 - \alpha)(1-\sigma(P_{i,j}))(1-\Concept_{i,j})
    \big)
    \label{eq:opt-loss}
\end{align}
where $\alpha=1-\frac{\left|\Concept\right|}{hw}$ is the sample-specific hyperparameter that balances the influence of foreground pixels (of count $\left|\Concept\right|$) and background pixels (of count $hw-|\Concept|$).

This loss proved to produce good results in the original Net2Vec work. However, it should be noted that compared to standard binary cross-entropy it sacrifices the convexity of the linear model's optimization. This and early stopping can result in suboptimal convergence. In an ablation study we find that \locVect{} $v$ initialization with zeros at 50 epochs diminishes these issues (cf.\,\cref{tab:ablation-total-init}). Future work could instead depart from the Net2Vec setup and go for BCE or Dice loss instead, which also are reported to yield promising results \cite{rabold2018explaining,schwalbe2022enabling}.

%
%


\subsection{Discussion of Context-sensitivity}
\label{sec:method-context}
In terms of extracted features, existing global C-XAI concept segmentation methods completely decouple concept and background (alias context): E.g., which learned features make up a \ConceptTerm{cat} against \emph{any} background? Meanwhile, many debugging tasks require to answer the local contrastive~\cite{guidotti2022counterfactual} question, which learned features make up a \ConceptTerm{cat} against \emph{a specific} background, e.g., \enquote{on a fluffy sofa}; and also how these differ from features of a \ConceptTerm{cat} when \enquote{on a fluffy sofa next to a \ConceptTerm{dog}}.
This is particularly relevant since CNNs do express concepts as a combination of other concepts \cite{%
fong2018net2vec,
mikolov2013linguistic
}. For example, $\ConceptTerm{tree}\approx\ConceptTerm{wood}+\ConceptTerm{green}$ \cite{fong2018net2vec}; and similarly $\ConceptTerm{cat}\approx \ConceptTerm{fluffy} - \ConceptTerm{sofa} - \ConceptTerm{dog}$.
These sub-components might only become visible locally: In an image without a dog, the features of a \ConceptTerm{fluffy} in contrast to the \ConceptTerm{sofa} might be more prevalent to determine the \ConceptTerm{cat} ($\ConceptTerm{cat}\approx\ConceptTerm{fluffy}-\ConceptTerm{sofa}$), other than in presence of a \ConceptTerm{dog} ($\ConceptTerm{cat}\approx\ConceptTerm{fluffy}-\ConceptTerm{dog}$).

Capturing such background-bias is valuable for tasks like concept-vs-context-sensitive image retrieval (here called \emph{concept retrieval}). This means retrieving images from a given image database that fulfill both the semantic constraint of what concept should be in there, as well as what context it is contrasted against.
Similarly, the local concept information can be used for \emph{outlier detection} \cite{chalapathy2019anomaly} that also respects background biases for determining outliers. Additionally, referring to concepts relative to their context better accommodates the context-sensitivity of CNNs \cite{ribeiro2016why}. 

However, due to the context-sensitivity, one should not expect \locVects{} features to exactly generalize (e.g., via taking a mean) to the optimal Net2Vec features. This would require randomizing the background of the single instances, possibly sacrificing realism or faithfulness to the training data.

\section{Using Local Concept Information}
\label{sec:method-generalization}

Like other C-XAI methods we here make the following established assumption \cite{kim2018interpretability,fong2018net2vec,ghorbani2019towards}:

\begin{assumption}\label{ass:similarity}
    In well-learned feature spaces, samples from the same semantic category tend to activate similar convolutional filters~\cite{fong2018net2vec}. Consequently, \locVects{} 
    representing similar concepts respectively concept-vs-context pairs are expected to have more similar values than those representing unrelated concepts. 
\end{assumption}


Existing global methods assume the ideal case, namely that this leads to a unimodal distribution with highly similar \locVects{} for instances of the same concept.
However, this assumption may easily be broken, e.g., by high variance within a concept (e.g., \ConceptTerm{proximate}, \ConceptTerm{mid-distance}, and \ConceptTerm{distant}  \ConceptTerm{car} as in \cref{fig:subconcepts}), by concept labeling errors, or incorrect prior assumptions. This is particularly likely in shallow network layers, where the merging of various variants of concept representations has not yet occurred.
This underlines why using a single global vector representation of concepts in feature space may be overly simplistic and not always yields accurate results~\cite{koh2020concept,mikriukov2023evaluating}.

Thus, this paper investigates how far deviations from the no-subconcepts-assumption manifest in \emph{multi-modal distributions} of concept instance information in the feature space, 
and how this information can be used for introspection purposes, like sub-concept and outlier retrieval. 
In this section, we propose two alternative clustering techniques to assess the distributions in \cref{sec:method-generalization-loctoglob} (with a discussion of complexity in \cref{sec:method-complexity}).
Before going into detail on these, \cref{sec:method-applications} provides an overview of applications.

\paragraph*{Single- vs. Multi-concept Distributions}
One intrinsic benefit of our local concept representation is that we can both inspect the (conditional) distribution of a single concept (only \locVects{} of that concept type, \enquote{pure} \locVects{}), as well as the joint distribution of all concepts (\locVects{} of any of the chosen concept types) respectively any mixtures, like the distribution of a subset of concepts.
The effect of using pure or mixed \locVects{} is similar to supervised and unsupervised C-XAI:

\shortparagraph{Pure Concepts}
By restricting the \locVects{} distribution to one concept, we can impose a strong human prior: outliers and sub-concepts of what a human considers concept $\Concept$.

\shortparagraph{Mixed Concepts}
The joint distribution of several concepts can reveal what are clusters/sub-concepts from point of perspective of the DNN, possibly slightly sacrificing explainability.

\subsection{\locVect{} Applications for DNN Debugging}
\label{sec:method-applications}

In the following, we explain how \locVects{} are naturally usable to
(1) recover modes of multi-modal distributions (i.e., \emph{sub-concepts});
(2a) find and inspect regions of concept representatives that are overlapping with other concepts (i.e., \emph{concept confusion}) or (2b) of low density (i.e., \emph{outliers}); and even
(3) \emph{retrieve images} that are similar with respect to the concept-to-context relation.
For experimental results, refer to \cref{sec:experiments}.

\subsubsection{Identification of Sub-concepts}
If multiple high-density regions represent a concept, it likely encompasses meaningful sub-concepts, i.e., distinct concept variants. For example, the concept 
\ConceptTerm{car} might be subdivided into \ConceptTerm{proximate}, \ConceptTerm{mid-distance}, and \ConceptTerm{distant} (cf.~\cref{fig:subconcepts});
\ConceptTerm{person} might be subdivided into \ConceptTerm{crowd of people} and \ConceptTerm{single person}, or \ConceptTerm{truck} into \ConceptTerm{van}, \ConceptTerm{pickup} and \ConceptTerm{semi-trailer truck}. This subdivision reveals the hierarchical structure within the network, showing how the DNN merges simple concepts into more complex ones.

More precisely, sub-concepts are groups of similar samples whose \locVects{} together establish a high-density region in the feature space related to that concept. 
%
By clustering \locVects{}, we can find these groups. This allows us to derive distinct sub-global \locVects{} (\subglobVects{}) for each discovered sub-concept. However, rather than relying on pre-defined sub-concepts, our method discovers these sub-concepts directly from the data by clustering \locVects{}.

\begin{definition}[Global and Sub-global \Vects{}]
The centroid, e.g., the mean, of a sub-cluster $Q$ of \locVects{} can serve as a single representing vector of that cluster.
We term such a centroid a sub-global \Vect{} (\subglobVect{}):
\begin{align}
    \mathbf{v} = \frac{1}{|Q|}\sum_{v\in Q} v.
    \label{eq:generalization-centroid}
\end{align}
The centroid of the cluster containing all \locVects{} is called the global \Vect{} (\globVect).
\end{definition}

Being generalizations, both sub-global and global \Vects{} allow inference on new samples, serving as an alternative to concept vectors obtained by Net2Vec (find a comparison of complexities for our clustering methods in \cref{sec:method-complexity}).
\Cref{fig:subconcepts} visualizes how the pure concept \ConceptTerm{car} breaks down into sub-clusters for \ConceptTerm{proximate}, \ConceptTerm{mid-distance}, and \ConceptTerm{distant}, by showing exemplary inference results for the sub-global \Vects{}.
For further results, refer to \cref{sec:applications-sub-concepts}.

\subsubsection{Identification and Quantification of Concept Confusion}
%


As discussed in \cref{ass:similarity}, well-learned feature spaces of models should distinguish between different concepts.
Concept confusion occurs when different concepts, from a human perspective, are encoded similarly in the feature space. This often manifests as an overlap in the high-density regions of \locVects. For example, our qualitative inspections suggested concept labeling errors as a prominent cause of concept confusion: If two instances both receive the label \ConceptTerm{truck} and \ConceptTerm{car}, resulting in two near-to-identical concept masks and respective \locVects{} that are of different concepts.
To identify and quantify concept confusion between two concepts, we suggest two approaches: One quantifies the purity after multi-concept clustering, and the other the cluster separation when considering each concept to be a single cluster.

\paragraph*{Cluster Purity in Multi-concept Distributions}
As one means to measure confusion between two concepts, we suggest the following procedure:
\begin{enumerate}[label=(\arabic*)]
    \item Cluster their joint set of \locVects{} into sub-clusters (see \cref{sec:method-generalization-loctoglob});
    \item Assess the \emph{purity} of sub-clusters as in \cref{eq:method-single-cluster-purity}, i.e., how much different concepts are mixed in the cluster;
    \item Select impure clusters, which indicate confusion between their most frequently contained concepts.
\end{enumerate}

\begin{figure*}[tbh]
  \centering
  \includegraphics[width=\linewidth]{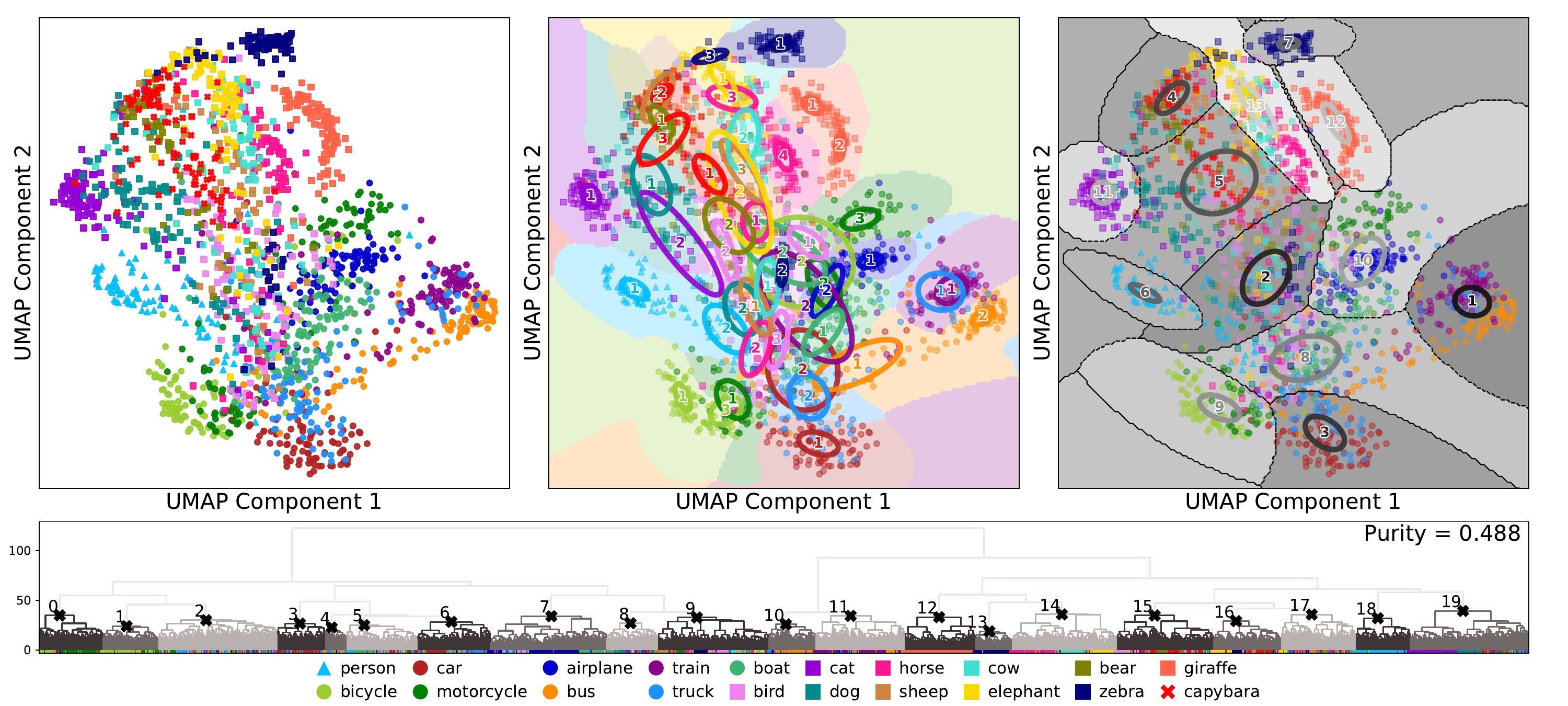}
  \caption{MS COCO \& Capybara Dataset: Generalization of all tested concepts \locVects{} in \texttt{model.encoder.layers.1} of DETR: 2D UMAP-reduced \locVects{} of every tested category (\emph{top-left}), GMMs fitted for samples with regard to their labels (\emph{top-middle}), GMMs fitted for all samples regardless of their labels (\emph{top-right}), and \locVects{} dendrogram with clusters identified with \cref{alg:adaptive-clustering} (\emph{bottom}). Find similar visualizations for further models in \cref{fig:gmm-dendrogram-efficientnet-f7,fig:gmm-dendrogram-swin-f7} (MS COCO \& Capybara Dataset) and \cref{fig:gmm-dendrogram-detr-e1-voc,fig:gmm-dendrogram-efficientnet-f7-voc,fig:gmm-dendrogram-swin-f7-voc} (PASCAL VOC).}
  \label{fig:gmm-dendrogram-detr-e1}
\end{figure*}

\Cref{fig:gmm-dendrogram-detr-e1} (top right) uses clustering on dimensionality reduced \locVects{} (later detailed in \cref{sec:method-clustering-gmm}) to illustrate how interwoven \locVects{} can be for different concepts. For example, in impure cluster 9, there is confusion between the \ConceptTerm{bicycle} and \ConceptTerm{motorcycle} categories, while cluster 7 shows confusion between \ConceptTerm{capybara} and \ConceptTerm{bear}.

Central to this approach is a notion of purity. We here measure the purity of a single cluster as follows.
\begin{definition}[Cluster Purity]
Cluster purity for a given cluster $Q$ is in this work simply defined as the maximum proportion of \locVects{} $v_{\Concept}$ with $\Concept$ from a common concept class $\ConceptClass\in\text{Concepts}$:
\begin{align}
\text{ClusterPurity}(Q) =& \tfrac{1}{|Q|} \max_{{\ConceptClass\in\text{Concepts}}}
    \left| Q \cap V_\ConceptClass\right|
    \label{eq:method-single-cluster-purity}
\end{align}
with $V_\ConceptClass$ the set of all \locVects{} belonging to category $\ConceptClass$.
It conveniently takes values in $[0,1]$.
\end{definition}

%
We generalize purity of a single cluster as follows to a global purity criterion that captures the homogeneity of a partition $\mathbf{Q}=\{Q_1,\dots,Q_n\}$ of a set $V$ of \locVects{}.
\begin{definition}[Purity]
Global Purity (further referred to as Purity) of a partition $\mathbf{Q}$ of \locVects{} $V$ is defined as
\begin{align}
    \text{Purity}(\mathbf{Q}) =
    \tfrac{1}{|V|} \sum_{Q \in \mathbf{Q}} \max_{\mathrlap{\ConceptClass \in \text{Concepts}}} |Q \cap V_\ConceptClass|
    \in [0,1]
    \label{eq:method-cluster-purity}
\end{align}
It calculates as the weighted mean of the per-cluster purities (\cref{eq:method-single-cluster-purity}), which each measure the homogeneity of data labels within a single cluster.
\end{definition}

This metric provides a numerical estimate of how well concepts of interest are isolated in the feature space. High global purity indicates that semantically similar samples are well-grouped, making it a useful tool for comparing the quality of feature space representations across different layers and models.



\paragraph*{Concept Separation in Single-concept Distributions}

Given several sets of \locVects{}, each representing a different concept, we aim to quantify the separation of these concepts—that is, identify how distinct concepts are within the given feature space.

To evaluate this, we propose using a classical metric from cluster analysis, namely the Separation Index (Dunn Index)~\cite{dunn1973fuzzy}. It estimates how effectively a model distinguishes between concepts in a given feature space (high separation) and identifies areas of potential overlap or confusion (low separation) where high-density regions of different concepts may intersect.
As a similar alternative, the Overlap Ratio, detailed in \cref{sec:appendix-overlap}, can be utilized.

In the following we assume that several concept representations are each given by a set of \locVects{}, interpreted each as a cluster.
The Separation Index quantifies how well-separated different concept representations are relative to their internal compactness, i.e., how differently concepts are encoded. It is defined as the ratio of concept compactness to distance between concepts and can be computed in two ways:
\begin{itemize}
    \item \textit{Absolute}: Evaluates the separation of a single concept $\ConceptClass_i$ relative to all other concepts.
    \item \textit{Pairwise}: Evaluates the separation between two specific concepts $\ConceptClass_i$ and $\ConceptClass_j$.
\end{itemize}

\begin{definition}[{Absolute Separation}]
Let $(\ConceptClass_i)_i$ be a set of concept representations, each a cluster of \locVects{}.
Then \textit{Absolute (Concept) Separation} of concept $\ConceptClass_i$ calculates as
\begin{align}
    \SwapAboveDisplaySkip
    \text{Separation}(\ConceptClass_i) = \frac{\text{\small Inter}(\ConceptClass_i, \bigcup_{j\neq i}\ConceptClass_j)}{\text{\small MeanIntra}(\ConceptClass_i)}
    \label{eq:separation-index-absolute}
\end{align}
where the pairwise \textit{Inter-Category Distance}, $\text{\small Inter}(\ConceptClass, \ConceptClass')$, measures the minimum distance between any \locVect{} in $\ConceptClass$ and any \locVect{} in $\ConceptClass'$:
\begin{align}
    \text{\small Inter}(\ConceptClass, \ConceptClass') = \min_{\mathclap{x \in \ConceptClass, x' \in \ConceptClass'}} \|x - x'\|_2
    \label{eq:pairwise-inter-class-dist}
\end{align}
and the \textit{Mean Intra-Category Distance}, $\text{\small MeanIntra}(\ConceptClass)$, measures the average pairwise distance between \locVects{} within concept $\ConceptClass$:
\begin{align}
    \text{\small MeanIntra}(\ConceptClass) = \tfrac{1}{|\ConceptClass| \cdot (|\ConceptClass| - 1)} \sum_{\mathclap{x, x' \in \ConceptClass, x \neq x'}}\|x - x'\|_2
    \label{eq:intra-cat-dist}
\end{align}
\end{definition}

\begin{definition}[{Pairwise Separation}]
With $\text{Inter}$ as above the \textit{Pairwise (Concept) Separation} of two concepts $\ConceptClass_i,\ConceptClass_j$ calculates as
\begin{align}
    \SwapAboveDisplaySkip
    \text{\small Separation}(\ConceptClass_i, \ConceptClass_j) = \frac{\text{\small Inter}(\ConceptClass_i, \ConceptClass_j)}{\text{\small MaxIntra}(\ConceptClass_i \cup \ConceptClass_j)}
    \label{eq:separation-index-pairwise}
\end{align}
and the \textit{Maximum Intra-Category Distance}, $\text{MaxIntra}(\ConceptClass_i \cup \ConceptClass_j)$, measures the maximum pairwise distance between any two \locVects{} in the combined categories:
\begin{align}
    \SwapAboveDisplaySkip
    \text{\small MaxIntra}(\ConceptClass) = \max_{\mathclap{x, x' \in \ConceptClass}} \|x - x'\|_2
    \label{eq:pairwise-intra-class-dist}
\end{align}

\end{definition}

\subsubsection{Detection of Concept-level Outliers}
Outliers at the concept level can arise from various factors, such as corrupted or low-quality labeling or a change with respect to concept or context style (e.g., blur, jitter; rare poses or backgrounds). These novel style or label combinations cause unusual features (thus \locVects{}) to be relevant to discriminate the concept from its background. The assumption is that the model is also prone to fail on such samples.
Unfortunately, if both concept and background individually are not new to the model, they may be hard to find using existing out-of-distribution detection methods, and if found, the source (which concept was an outlier) is hard to determine.

We thus propose using a distance
metric on \locVects{} to identify such outliers (see \cref{sec:applications-outliers} for experimental results).
Here, we calculate the cumulative $L_2$ distance ($\Sigma L_2$) of each \locVect{} from other vectors within the same category. The most remote samples can be identified, inspected, and flagged as outliers by sorting these distances.
\begin{definition}[Cumulative $L_2$ Distance]\label{def:outliers-sigma-l2}
    The cumulative $L_2$ distance of a \locVect{} $v$ within the set of \locVects{} $V$ is defined as
    \begin{align}
    \Sigma L_2(v) \coloneqq \sum_{\substack{v' \in V_\ConceptClass}\setminus\{v\}} \| v - v' \|_2
    \;.
    \label{eq:outliers-sigma-l2}
\end{align}

\end{definition}


\subsubsection{Concept-based Image retrieval}
\locVects{} encode how a specific concept contrasts against its background. Following \cref{ass:similarity},
\locVects{} vectors corresponding to concepts-to-background settings should exhibit high similarity, i.e., a low distance from each other. Consequently, \locVects{} can be effectively used to retrieve similar concepts in comparable contexts by querying those images with similar \locVects{} from an \locVect{}-bank.

\shortparagraph{Quantitative Evaluation}
As a further application, the retrieval quality 
provides another quantitative measure of how well similar concepts are separated in the feature space.
Typical metrics from the image retrieval domain, such as mAP@k (detailed in \cref{sec:setup-metrics}), can serve for this quantitative analysis.

\begin{figure*}[tbh]
  \centering
  \includegraphics[width=\linewidth]{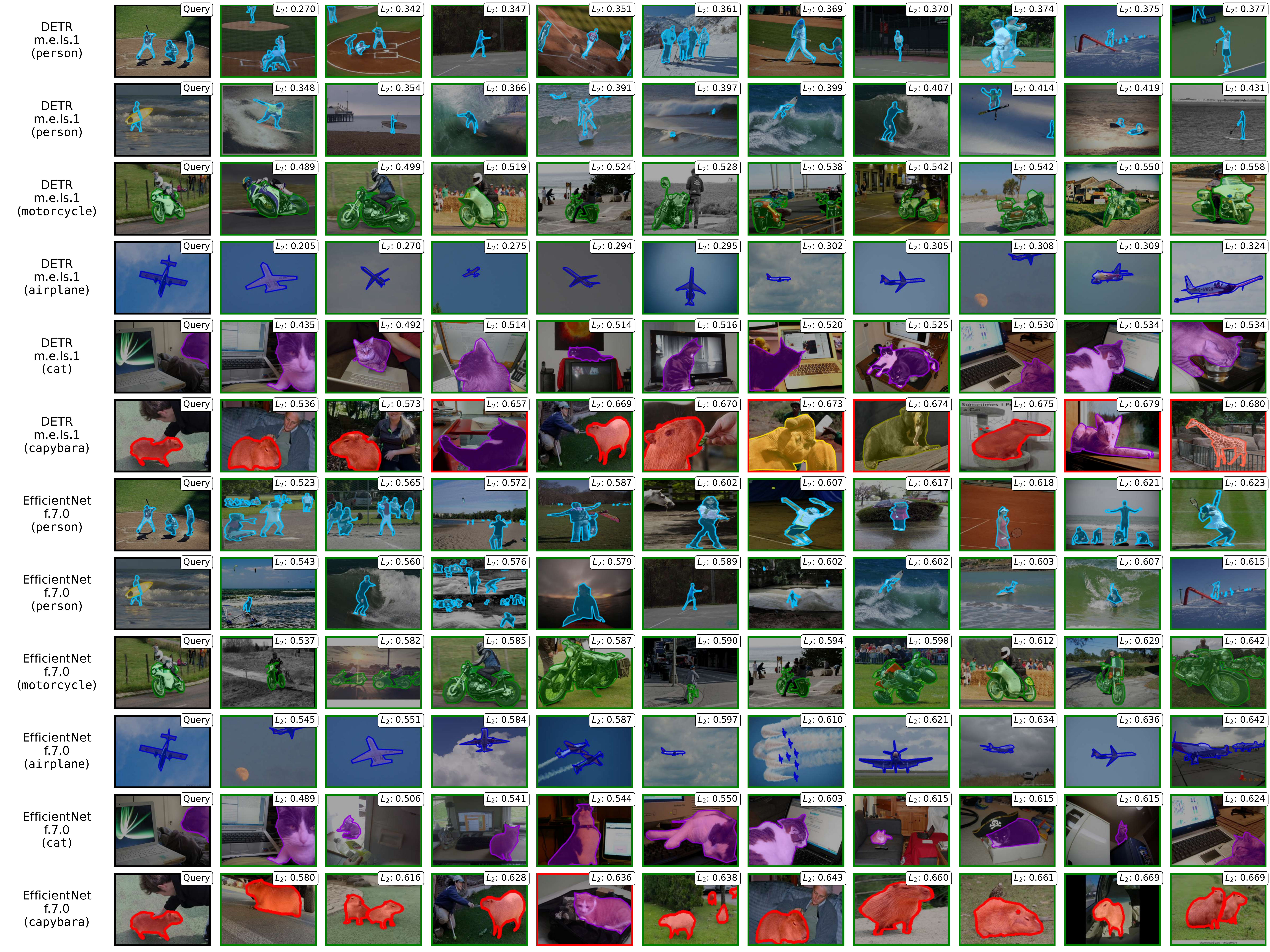}
  \caption{MS COCO \& Capybara Dataset: Top 10 retrieval results according to $L_2$-distance (\emph{columns}) for random \locVect{} queries (\emph{black frame, leftmost column}) of different categories in best model DETR (\emph{top}) and worst model EfficientNet (\emph{bottom}). \emph{Green frame} - relevant sample. \emph{Red frame} - irrelevant sample. Unique concepts are color-coded. Find similar visualizations for further models in \cref{fig:retrieval-qualitative-extra1,fig:retrieval-qualitative-extra2}.}
  \label{fig:retrieval-qualitative}
\end{figure*}

\shortparagraph{Qualitative Evaluation}
Qualitatively, the performance of retrieval can be evaluated by inspecting the retrieved samples. For example, in a well-trained feature space, a query with a \locVect{} encoding a \ConceptTerm{cat} with a monitor in the background should retrieve results where the majority corresponds to cats with monitors in the background (see \cref{fig:retrieval-qualitative}).
Inspection may, on the other side, also reveal cases in which human understanding of concept features do not align with those the DNN has learned. For example, \cref{fig:retrieval-qualitative} first row shows how people playing baseball is matched with any kind of persons in sportive activities like skiing, regardless of the stadium setting.

For detailed experimental results see \cref{sec:applications-retrieval}.

\subsection{%
Clustering of \locVects{}
}
\label{sec:method-generalization-loctoglob}

The previously introduced analysis techniques all require a way to generalize the set of local \locVects{} to some distribution or (sub-)cluster information.
We here introduce two techniques to analyze \locVect{} distributions via clustering:
(1) by means of hierarchical agglomerative clustering on the original \locVects{} (\cref{sec:method-clustering-hierarchical}),
and 
(2) by means of Gaussian mixture model fitting on dimensionality reduced \locVects{} for better visualization (\cref{sec:method-clustering-gmm}).

\begin{figure*}[tbh]
  \centering
  \includegraphics[width=\linewidth]{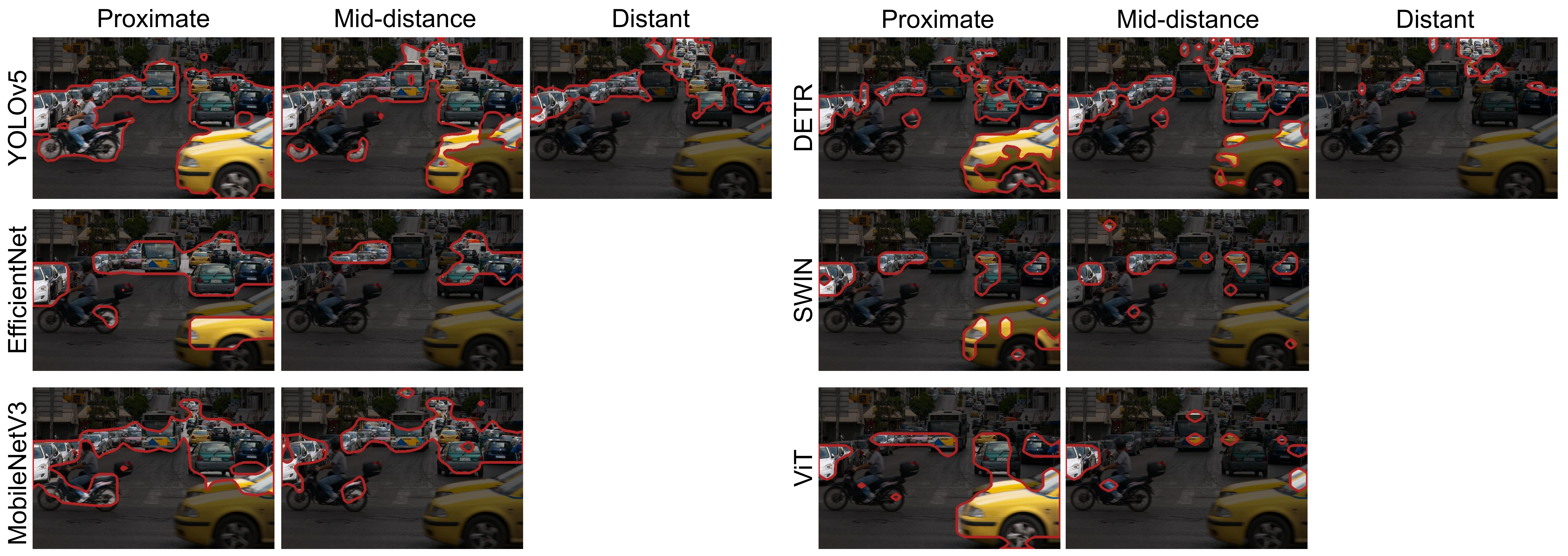}
  \caption{MS COCO: Detected distance sub-concepts of \ConceptTerm{car} (\emph{columns}) in different networks (\emph{rows}). Sub-concept segmentations were obtained by projecting activations with sub-concept cluster centroids (\subglobVects, \cref{eq:generalization-centroid}).  Sub-concept clusters were discovered in \locVect{} dendrograms using manual thresholding in \textbf{bold layers} of \cref{tab:layers}. Find more results in \cref{fig:subconcepts-extra}}
  \label{fig:subconcepts}
\end{figure*}



\subsubsection{Hierarchical clustering}
\label{sec:method-clustering-hierarchical}

Our goal is to find sets of \locVects{} that each together represent a high-density region of the concept's distribution. One technique we propose for this is agglomerative hierarchical clustering, i.e., bottom-to-top merging of sub-clusters of \locVects{} into larger (sub-)clusters. 
An advantage of hierarchical clustering is that it by definition yields not only a set of sub-clusters, but a (deep) hierarchy. Hence it captures not only high-density regions, but also their similarity and further sub-division.
Thus, it does not require
dimensionality reduction to interpret results, but can be illustrated using hierarchy dendrograms as done in the bottom plot of \cref{fig:gmm-dendrogram-detr-e1}.
%

We here conduct two steps to obtain interesting sub-clusters respectively sub-global \locVects{}: (1) apply hierarchical clustering to obtain the full hierarchy tree of \locVect{} clusters, and (2) from these select the largest coherent sub-clusters (prune the hierarchy).

\paragraph{Obtain Full Cluster Hierarchy}
We suggest using the common Ward's linkage agglomerative clustering\footnote{\tiny\url{https://docs.scipy.org/doc/scipy/reference/generated/scipy.cluster.hierarchy.ward}}~\cite{ward1963hierarchical}, which minimizes variance within each cluster and requires Euclidean distance as the distance metric. Alternatively, a combination of complete (or furthest neighbor) linkage clustering\footnote{\tiny\url{https://docs.scipy.org/doc/scipy/reference/generated/scipy.cluster.hierarchy.complete}} and cosine distance may also yield good results.
The result of the clustering is a full hierarchy binary tree (dendrogram) from single \locVects{} as leaves to the full set of \locVects{} as root, in the form of a linkage table of nodes.

\paragraph{Hierarchy Pruning: Selecting Sub-global \Vects{}}
The next step now is to prune the lower part of the hierarchy tree such that a manageable amount of sub-clusters remains 
(cf.\ bottom plots in \cref{fig:gmm-dendrogram-detr-e1,fig:gmm-dendrogram-efficientnet-f7,fig:gmm-dendrogram-swin-f7,fig:gmm-dendrogram-detr-e1-voc,fig:gmm-dendrogram-efficientnet-f7-voc,fig:gmm-dendrogram-swin-f7-voc}).
We employ the following strategies for cluster selection: 
(1) selecting branches based on linkage distance and, only for mixed clustering, (2) a new simple adaptive cluster selection strategy detailed in the following (\cref{alg:adaptive-clustering}).

\begin{figure}[tbh]
    \centering
    \begin{algorithm}[H]
        \SetAlgoLined
        \fontsize{8pt}{10pt}\selectfont
        \SetKwFunction{SelectCluster}{SelectCluster}
        \SetKwFunction{DendrogramToClusters}{DendrogramToClusters}
        \SetKw{ElsIf}{else if}
        \SetKwProg{Fn}{function}{:}{}

        $\textup{\texttt{CPT = 0.8}}$
        
        $\textup{\texttt{CST = 0.05 * TotalNodes}}$

        \Fn{\DendrogramToClusters{$\textup{\texttt{Node}}$}}{
            \uIf{$\textup{\texttt{Node.ClusterPurity > CPT}}$}{
                \SelectCluster{$\textup{\texttt{Node}}$}\;
            }
            \uElseIf{$\textup{\texttt{Node.Size < CST}}$}{
                \SelectCluster{$\textup{\texttt{Node}}$}\;
            }
            \Else{
                \DendrogramToClusters{$\textup{\texttt{Node.Left}}$}\;
                \DendrogramToClusters{$\textup{\texttt{Node.Right}}$}\;
            }
        }
        \caption{Adaptive Cluster Selection}
        \label{alg:adaptive-clustering}
    \end{algorithm}
\end{figure}

\shortparagraph{Adaptive Cluster Selection}
Assume one is given a dendrogram of mixed concept \locVects{}. Good clusters (sub-trees) are considered those that are pure, i.e., mostly contain \locVects{} of one concept.
\Cref{alg:adaptive-clustering} traverses the dendrogram from root to leaves. 
A sub-tree is recognized as a valid cluster if it meets any of the following two conditions: (1) its purity exceeds the cluster purity threshold (\texttt{CPT}; \cref{eq:method-single-cluster-purity}), meaning it contains a high proportion of samples from the same category; or (2) further subdivision would result in clusters smaller than the cluster size threshold (\texttt{CST}). In experiments, we set \texttt{CPT} to 0.8 and \texttt{CST} to 5\% of the test samples.

\subsubsection{Generalization with UMAP-GMM}
\label{sec:method-clustering-gmm}
An alternative to processing vectors directly in their original manifold is applying dimensionality reduction methods. These techniques are advantageous because they enable the visualization of high-dimensional data in lower dimensions, making it more comprehensible for humans. Methods such as UMAP~\cite{mcinnes2018umap} locally preserve distribution density, which helps maintain fidelity for future generalizations.

In our experiments, we (1) reduce the dimensionality of the \locVects{} set to 2 dimensions using UMAP\footnote{\tiny\url{https://github.com/lmcinnes/umap}}, and (2) fit a multivariate Gaussian mixture model (GMM)\footnote{\tiny\url{https://scikit-learn.org/stable/modules/generated/sklearn.mixture.GaussianMixture}} to the resulting 2d-embeddings. The optimal number of GMM components is determined using the Bayesian information criterion (BIC).

The two ways to cluster also apply to GMMs, namely: (a) to the entire set of UMAP-embeddings without considering labels of their original \locVects{}, or (b) to label-specific subsets (top-mid and top-right plots in \cref{fig:gmm-dendrogram-detr-e1,fig:gmm-dendrogram-efficientnet-f7,fig:gmm-dendrogram-swin-f7,fig:gmm-dendrogram-detr-e1-voc,fig:gmm-dendrogram-efficientnet-f7-voc,fig:gmm-dendrogram-swin-f7-voc}). Their effect is as follows: The first approach helps identify concept-level outliers and determine the number of sub-concepts (i.e., high-density regions) within a single concept label. The second approach reveals potential concept confusion and data labeling errors within the feature space (see \cref{sec:method-applications,sec:applications}). 

\subsection{Time and Memory complexity of \Vects{} and  Global \Vects{}}
\label{sec:method-complexity}

The time and memory complexity of \locVect{} and Net2Vec are comparable during the initial optimization phase, both requiring $O(e \times n \times C \times H \times W)$ for time and $\Omega(n \times C \times H \times W)$ for memory, where $n$ is the number of samples and $e$ denotes the optimization epochs.

However, during the generalization phase, \locVect{} incurs additional complexity. For example, hierarchical clustering has the time complexity of $O((n \times C)^2)$ and memory complexity of $\Omega((n \times C)^2)$. Please refer to \cref{sec:appendix-a} for a detailed analysis.

\section{Experimental Setup}
\label{sec:setup}

This section describes the experimental setup common to experiments in \cref{sec:experiments,sec:applications}.

\subsection{\Vect{} settings}
\label{sec:setup-concept-vector}

Unlike prior work~\cite{fong2018net2vec}, we recommend initializing the entries of the to-be-optimized \locVect{} $v$ to 0 (see ablation results in \cref{sec:experiments-ablation}). For optimization, we use the AdamW optimizer~\cite{loshchilov2017adamw} with a learning rate of $0.1$ and default parameters, running for $50$ epochs. The simplicity of the linear model minimizes the risk of overfitting, eliminating the need for early stopping.

Furthermore, we rescale activations and segmentations to $100 \times 100$ in all experiments, as this setting offers the best trade-off between speed and performance quality (see \cref{sec:experiments-ablation}).

\subsection{Models and Architectures}
\label{sec:setup-models}

To evaluate the applicability of our approach, we conduct experiments using diverse CNN and ViT classification and object detection models: MobileNetV3-L\footnote[4]{\label{footnote:torch-zoo}\tiny\url{https://pytorch.org/vision/stable/models}}~\cite{howard2019searching}, EfficientNet-B0\textsuperscript{\ref{footnote:torch-zoo}}~\cite{tan2019efficientnet}, YOLOv5s\footnote[5]{\tiny\url{https://pytorch.org/hub/ultralytics_yolov5/}}~\cite{yolov5}, with
residual backbone, and ViT-B-16\textsuperscript{\ref{footnote:torch-zoo}}~\cite{dosovitskiy2021vit}, SWIN-T\textsuperscript{\ref{footnote:torch-zoo}}~\cite{liu2021swin}, and DETR\footnote{\tiny\url{https://huggingface.co/facebook/detr-resnet-50}}~\cite{carion2020detr} with ResNet50~\cite{simonyan2014very} backbone. For brevity, we refer to models without specifying their variants (e.g., ViT instead of ViT-B-16). See \cref{tab:used-models} for the model categorization and sizes.

\begin{table}
    \centering
    \begin{tabular}{@{}|c|c|c|@{}}
        \hline
        \textbf{Type}
        & \textbf{Classifiers}
        & \textbf{Detectors} \\
        \hline
        \multirow{2}{*}{CNN} 
        & MobileNetV3 (5.5M)
        & \multirow{2}{*}{YOLOv5 (7.2M)} \\
        & EfficientNet (5.3M)
        &  \\
        \hline
        \multirow{2}{*}{ViT}
        & ViT (86.6M)
        & \multirow{2}{*}{DETR (41.0M)} \\
        & SWIN (28.3M)
        &  \\ 
        \hline
    \end{tabular}
    \caption{Tested CNN and ViT classifiers and detectors. Model sizes (millions of parameters) are specified in brackets.}
    \label{tab:used-models}
\end{table}

\begin{figure*}[ht]
  \centering
  \includegraphics[width=\linewidth]{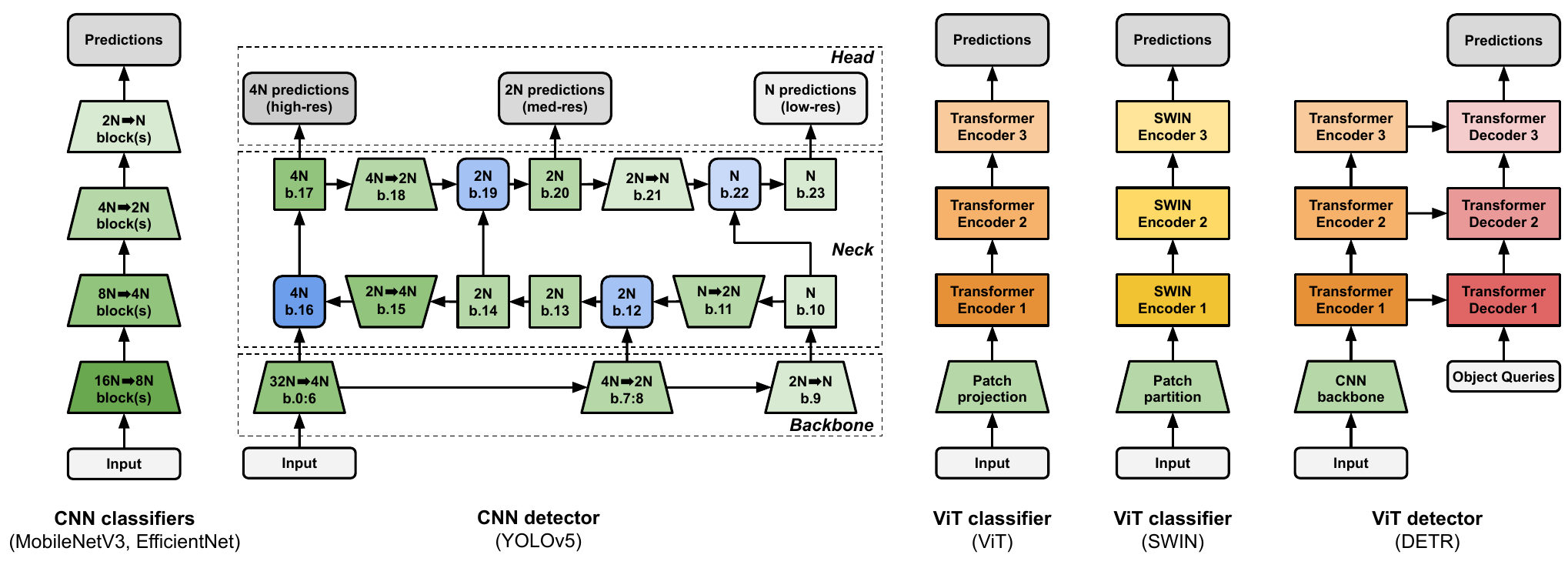}
  \caption{
Simplified architectures of tested networks. To maintain clarity, only key blocks and connections are shown, with fewer blocks displayed for simplicity. For CNNs, the relative resolution of activations is marked as \texttt{xN} (transformer blocks retain dimensionality). Upscaling and downscaling are denoted by $\rightarrow$. Blocks of YOLOv5 are labeled as \texttt{b.z}, with block sequences represented as \texttt{b.z:y}, where \texttt{z} and \texttt{y} refer to block indices.}
  \label{fig:network-architectures}
\end{figure*}

\cref{fig:network-architectures} illustrates the differences among the selected computer vision model architectures. The inverted residual CNN classifiers, EfficientNet and MobileNetV3, and the vision transformer classifiers, ViT (vanilla) and SWIN (with a hierarchical shifted-window mechanism) follow a linear block structure. These models, trained on ImageNet-1k~\cite{deng2009imagenet}, are designed explicitly for object-level image classification.

One-stage YOLOv5 and transformer DETR detectors are trained on more semantically complex MS COCO~\cite{lin2014microsoft}. YOLOv5 has several interconnected branches and detection heads of different resolutions (magnitudes of resolution are marked as \textbf{xN}). DETR employs an encoder-decoder architecture (encoders are similar to those in ViT). The encoder and decoder blocks are connected vertically and horizontally. Such architectural peculiarities of tested detectors create \enquote{shortcuts}, which influence the flow of semantics inside models (see \cref{sec:applications-shortcut-detection}).

\subsection{Datasets}
\label{sec:setup-data}

Our experiments focus on general-purpose datasets that primarily feature indoor and outdoor scenes, covering natural landscapes, urban settings, and scenarios related to the autonomous driving domain. While this scope allows us to evaluate our method under diverse yet relatable conditions, it excludes other specialized domains such as medical imaging (e.g., histopathology) or industrial applications (e.g., quality assurance in production lines). 

\shortparagraph{MS COCO}
Within this context, we utilize the validation subset of the MS COCO 2017\footnote{\tiny\url{https://cocodataset.org/}} dataset~\cite{lin2014microsoft}, which includes bounding boxes and segmentations for 80 concepts of high-level abstraction. We specifically focus on 19 concepts within the \ConceptTerm{person}, \ConceptTerm{vehicle}, and \ConceptTerm{animal} supercategories. In total, we used 3,212 unique segmentation instances across 2,382 unique images, as some images contain segmentations for multiple categories. MS COCO is a general-purpose dataset of high complexity, with a broader set of categories, diverse object sizes, and complex indoor and outdoor scenes that include occlusions and cluttered backgrounds.

\shortparagraph{PASCAL VOC}
To further evaluate the generalizability of our method, we include the validation subset of the PASCAL VOC 2012\footnote{\tiny\url{http://host.robots.ox.ac.uk/pascal/VOC/}}~\cite{everingham2010pascal}, which provides high-quality segmentations for 20 concepts within \ConceptTerm{person}, \ConceptTerm{vehicle}, \ConceptTerm{animal}, and \ConceptTerm{object} supercategories. The validation subset comprises 1,449 images and 2,147 segmentation instances. Like MS COCO, PASCAL
VOC is a general-purpose dataset that includes indoor and outdoor scenes. However, compared to MS COCO, PASCAL VOC features fewer categories and simpler annotation structures, making it a less semantically complex dataset. We hypothesize that our method will demonstrate superior performance on PASCAL VOC due to this reduced complexity, serving as a benchmark for testing generalizability and method robustness.

\shortparagraph{The Capybara Dataset}
To test the performance of \locVects{} in out-of-distribution conditions, we use the Capybara Dataset\footnote{\tiny\url{https://github.com/freds0/capybara_dataset}}, featuring images of capybaras (\textit{Hydrochoerus hydrochaeris}) unseen by the tested models. Our experiments use 165 manually segmented samples\footnote{\tiny\url{https://github.com/comrados/capybara_dataset/}}, each containing one or several capybaras, with potential co-occurrence of other entities such as \ConceptTerm{person}.

Consequently, most of our experiments are conducted on MS COCO or the MS COCO and the Capybara dataset pair, as these datasets provide valuable insights into model performance under challenging and out-of-distribution conditions.

\subsection{Layer Selection}
\label{sec:setup-layers}

For \Vect{} ablation and performance experiments, we selected 5 to 8 layers (depending on the number of blocks in each network) from various depths across the tested networks: output layers of convolutional blocks and Transformer encoders. For ViTs, we transform the outputs of block's last MLP layers into quasi-CNN activations, as described in \cref{sec:method-transformers}. A detailed list of the chosen layers is provided in \cref{tab:layers-ablation}.

\begin{table*}
    \centering
    \fontsize{8pt}{10pt}\selectfont
    \begin{tabular}{@{}|c|cccccccc|@{}}
        \hline
        \multirow{2}{*}{\textbf{Model}} & \multicolumn{8}{c|}{\textbf{Ablation and Performance Test Layers}} \\
         & $l_1$ & $l_2$ & $l_3$ & $l_4$ & $l_5$ & $l_6$ & $l_7$ & $l_8$ \\
        \hline
        YOLOv5 & 6.cv3.c & 8.cv3.c & 9.cv2.c & 12 & 16 & 17.cv3.c & 20.cv3.c & 23.cv3.c \\
        MobileNetV3 & f.9 & f.10 & f.11 & f.12 & f.13 & f.14 & f.15 & --- \\
        EfficientNet & f.4.2 & f.5.0 & f.5.1 & f.5.2 & f.6.0 & f.6.1 & f.6.2 & f.7.0 \\
        \hline
        DETR & m.b.ce.m.l3 & m.ip & m.e.ls.0 & m.e.ls.1 & m.e.ls.2 & m.e.ls.3 & m.e.ls.4 & m.e.ls.5 \\
        ViT & e.ls.el1 & e.ls.el3 & e.ls.el5 & e.ls.el7 & e.ls.el9 & e.ls.el11 & --- & --- \\
        SWIN & f.0 & f.1 & f.3 & f.5 & f.7 & --- & --- & --- \\
        \hline
    \end{tabular}
    \caption{Layers for ablation and performance study (features=f, layer=l, layers=ls, encoder=e,  model=m, backbone=b, conv=c, conv\_encoder=ce, encoder\_layer\_=el, input\_projection=ip).}
    \label{tab:layers-ablation}
\end{table*}

For more detailed testing and comparison with baselines, we study 3 layers (2 for SWIN, see \cref{sec:experiments-performance}) from each network. Due to the selection of concepts of high-level abstraction, we use mid-depth, deep, and last (pre-classification or pre-detection head) layers, see \cref{tab:layers}.

\begin{table}[tbh]
    \centering
    \begin{tabular}{@{}|c|ccc|@{}}
        \hline        
        \multirow{2}{*}{\textbf{Model}} & \multicolumn{3}{c|}{\textbf{Application Layers}} \\
        & $l_{mid}$ & $l_{deep}$ & $l_{last}$ \\
        \hline
        YOLOv5 & 6.cv3.c & \textbf{14.c} & 20.cv3.c \\
        MobileNetV3 & f.9 & f.11 & \textbf{f.14} \\
        EfficientNet & f.5.2 & f.6.2 & \textbf{f.7.0} \\
        \hline
        DETR & m.b.ce.m.l3 & \textbf{m.e.ls.1} & m.e.ls.5 \\
        ViT & e.ls.el3 & e.ls.el7 & \textbf{e.ls.el11} \\
        SWIN & --- & f.5 & \textbf{f.7} \\
        \hline
    \end{tabular}
    \caption{Layers for applications and comparison with baselines (features=f, layer=l, layers=ls, encoder=e,  model=m, backbone=b, conv=c, conv\_encoder=ce, encoder\_layer\_=el). Layers in \textbf{bold} used for sub-concept discovery and category separation tests.}
    \label{tab:layers}
\end{table}

For the DETR model, our analysis is limited to outputs of the convolutional backbone and outputs of linear layers in encoder blocks. This restriction arises due to the unique characteristics of DETR’s implementation: the decoder layers use abstract learnable object queries, which query information of encoder outputs in the cross-attention layer. Since the decoder outputs have different dimensions (i.e., $(B \times HW \times C)$ in the encoder and $(B \times Q \times C)$ in the decoder, where $Q$ is the number of object queries), they cannot be converted into quasi-activations (\cref{sec:method-transformers}). However, decoder cross-attention and encoder self-attention layers offer inherent explainability via their weights visualization\footnote{\tiny\url{https://github.com/facebookresearch/detr/blob/colab/notebooks/detr_attention.ipynb}}.

\subsection{Comparison Baselines}
\label{sec:setup-baselines}

In our experiments, we compare local and generalized concept vectors from the \Vect{} framework against three global concept-based frameworks: Net2Vec~\cite{fong2018net2vec}, Net2Vec-16~\cite{fong2018net2vec}, and NetDissect~\cite{bau2017network}. To ensure a fair comparison, we apply the same visualization and binary projection mask estimation rules to all baseline methods as we do for \Vect{} (see~\cref{sec:method-optimization}).

NetDissect uses a naive approach where a concept is represented by a distinct neuron (convolutional filter). Net2Vec trains a single global vector, which represents the weights of each convolutional filter, to embed the concept. Net2Vec-16 modifies this by assuming that only a small subset of convolutional filters is sufficient to encode a concept. This version is represented as a sparse concept vector, with only a few non-zero values. In our experiments, we set the number of non-zero filters to 16, following the recommendation in the original paper as optimal for explaining object concepts~\cite{fong2018net2vec}.

\subsection{Metrics}
\label{sec:setup-metrics}
\paragraph*{Benchmarking Concept Segmentation}
To measure the quality of obtained concept segmentations (which is the inference performance of the corresponding concept models), we follow prior work \cite{fong2018net2vec,schwalbe2021verification}
and use the Jaccard index, more precisely set intersection over union (IoU):
\begin{align}
    \text{IoU}(P, \Concept) = \frac{ \left| \sum_{i,j}  P_{i,j}^{(\text{bin})} \land \Concept_{i,j} \right| }{ \left| \sum_{i,j}  P_{i,j}^{(\text{bin})} \lor \Concept_{i,j} \right|}
    \label{eq:opt-iou}
\end{align}
where, $P_{i,j}^{(\text{bin})} = \begin{cases} 1 & \text{if } \sigma(P)_{i,j} > \frac{1}{2}, \\ 0 & \text{otherwise}. \end{cases}$ is the binarized predicted concept mask.

\paragraph*{Benchmarking Image Retrieval}
Here, we extract local context-sensitive encodings of object concepts from DNNs and show how they can be used for context-aware image retrieval.
The quality of image retrieval is used as a metric for purity in close-by neighborhoods of samples.

As a subfield of information retrieval, image retrieval retrieves the list of top $k$ images relevant to a query from an image bank. Relevance means, e.g., that the images match the queried class. A retrieval is considered good if the top retrieved samples are truly relevant.
Mean Average Precision (mAP) is a widely used metric for evaluating retrieval performance~\cite{dubey2021decade}. It is often assessed for a specific top-$k$ precision (mAP@k), which evaluates the precision of the top-$k$ retrieved items ranked by their distance from the query%
\footnote{mAP approximates the area under the precision-recall curve as the number of top-$k$ results from the retrieval list increases \cite{turpin2006user}.}:
\begin{align}
    \text{mAP@k} = \frac{1}{|Q|} \sum_{q\in Q} \frac{1}{r_q} \sum_{i=1}^{k} P_q(i) \times rel_q(i)
    \label{eq:retrieval-mAP}
\end{align}
where $Q$ is the set of queries, $r_q$ is the number of relevant items for query $q$, $k$ is the number of retrieved samples, $rel_q(k)$ is a binary indicator that equals 1 if the $i$-th retrieved item is relevant to the query and 0 otherwise, and $P_q(i)=\frac{1}{i}\sum_{j=1}^i rel_q(j)$ is the precision at the $i$-th position.

\paragraph*{Benchmarking Concept Robustness}
While cosine similarity is a widely used metric for assessing the similarity between data sets, its sensitivity to offsets limits its applicability in some scenarios. Normalized Cross-Correlation (NCC), also known as Pearson Correlation Coefficient (PCC), a mathematically equivalent form of mean-centered cosine similarity, addresses this limitation by focusing on the structural alignment of data while removing biases caused by offsets and overall trends.

NCC evaluates the preservation of structural patterns in data when subjected to variations such as noise or perturbations. NCC focuses on the relationships within data, isolating relative patterns while being invariant to scale and offset differences. By mean-centering the data, NCC removes global trends or biases, ensuring that the comparison emphasizes deviations from the mean rather than absolute magnitudes. This makes it particularly suited for evaluating the impact of noise on the integrity of underlying data structures.

For two sets of \locVects{}, $V$ (from the original data) and $V'$ (from the noisy, perturbed data), NCC is defined as:

\begin{align}
    \text{NCC}(V, V') = \cos(V_c, V'_c) =
    \frac{V_c \cdot V'_c}
    {\|V_c\| \|V'_c\|},
    \label{eq:cross-correlation}
\end{align}

where $V_c = V - \mu$ and $V'_c = V' - \mu'$ are the mean-centered versions of $V$ and $V'$, with $\mu$ and $\mu'$ denoting their respective means.

NCC values range from $-1$ to $1$. A value of $1$ indicates perfect preservation of structural patterns between the original and perturbed data, while $0$ signifies no correlation, indicating complete disruption of patterns. A value of $-1$ represents a perfect inverse relationship, reflecting structural distortion. 
%
Higher NCC values correspond to greater robustness, capturing scenarios where the internal relationships in the data are preserved even in the presence of noise or perturbations.

\section{Experiments: Concept Distribution in DNNs}
\label{sec:experiments}

Here, we present a series of experiments aimed at validating \Vect{} as a means for concept distribution analysis and using it to test how strongly typical DNN latent spaces exhibit subconcepts and concept confusion.

First, we conduct ablation studies to examine the sensitivity of \locVects{} to various hyperparameters and robustness to the input noise, revealing that while success is sensitive to these settings, with the right configuration \Result{one can achieve good results consistently and robustly across different models and layers}.

Next, several analyses of the \locVect{} distribution are used to uncover \Result{the inherent complexity of concept representations in neural networks, which exist as overlapping distributions rather than simple point estimates} (\cref{sec:experiments}). This core result emphasizes the limitations of global C-XAI methods and the importance of local distribution-based approaches.
In particular, a final comparison of \Vects{} to state-of-the-art global methods finds that \Result{\locVects{} deliver results competitive to baselines}, despite their context-sensitivity. A slight but consistent performance drop from \subglobVects{} to Net2Vec-like methods showcases the loss in concept insights when going from local to global representatives.
In summary, our method is the first one to effectively overcome the limitations of global C-XAI point estimates.

\subsection{Ablation Study}
We here determine optimal settings of hyperparameters (\cref{sec:experiments-ablation}) and the influence of concept, model, and layer (\cref{sec:experiments-performance}).
In particular, hyperparameter sets that achieve good results across models, layers, and concepts can be found, and other factors show similar influence as for existing C-XAI methods. Additionally, we confirm the robustness of our method with respect to noise in the input (\cref{sec:ablation-noise}).

\subsubsection{Ablation of \locVect{} Hyperparameters}
\label{sec:experiments-ablation}


Recall that the original Net2Vec loss used here sacrifices the optimization problem's convexity. We investigate how far convergence issues due to this and early stopping can be avoided via the choice of parameter initialization. Apart from that, the mismatch between input and activation map resolution results in a tradeoff between computational cost (comparison in low-resolution) and IoU performance (high-resolution).
Therefore, in our ablation studies, we assess the quality and success rate of \locVects{} convergence under different conditions: (1) varying initialization parameters of \locVect{} and (2) different resolutions of target masks and activations. We evaluate performance by measuring the mean IoU between the approximated projections and the target masks, and we also track optimization failures (when IoU = 0). These tests were conducted on a random selection of 950 samples from the MS~COCO dataset, with 50 samples chosen from each tested category. Importantly, while \Result{the success of \locVects{} is influenced by its hyperparameters, good results can be obtained with the right settings}. 

\begin{table*}[tbph]
	\centering
	\fontsize{8pt}{10pt}\selectfont
	\setlength{\tabcolsep}{2.5pt} 
	\begin{tabular}{@{}|c|c|cccccccc|cccccccc|@{}}
		\hline
		\multirow{2}{*}{\textbf{Model}}        & \multirow{2}{*}{\textbf{$v$ init}} &   \multicolumn{8}{c|}{\textbf{IoU}}                                                                                                      & \multicolumn{8}{c|}{\textbf{Optimization failures}, \%}                                                                             \\
                              &                           & $l_1$         & $l_2$         & $l_3$         & $l_4$         & $l_5$         & $l_6$         & $l_7$         & $l_8$         & $l_1$         & $l_2$         & $l_3$         & $l_4$        & $l_5$        & $l_6$         & $l_7$        & $l_8$        \\
            \hline
                            \multirow{4}{*}{YOLOv5}       & ones                      & 0.69          & 0.68          & 0.21          & 0.73          & 0.64          & 0.16          & 0.21          & 0.22          & \textbf{0.3}  & \textbf{0.9}  & 73.2          & \textbf{0.5} & \textbf{0.2} & \textbf{0.0}  & \textbf{0.0} & \textbf{0.2} \\
                            & normal                    & 0.68          & 0.66          & 0.52          & 0.62          & 0.62          & 0.52          & 0.57          & 0.50          & 2.6           & 6.0           & 23.3          & 15.3         & 12.1         & 16.5          & 14.4         & 23.3         \\
                            & uniform                   & \textbf{0.72} & \textbf{0.69} & 0.39          & \textbf{0.74} & 0.68          & 0.22          & 0.48          & 0.35          & 0.5           & 1.4           & 46.5          & 0.7          & \textbf{0.2} & 0.1           & 0.5          & 0.6          \\
                            & zeros                     & 0.70          & \textbf{0.69} & \textbf{0.55} & 0.69          & \textbf{0.70} & \textbf{0.62} & \textbf{0.65} & \textbf{0.59} & 0.4           & 1.7           & \textbf{22.2} & 5.3          & 3.1          & 4.5           & 2.1          & 5.0          \\
            \hline
                            \multirow{4}{*}{MobileNetV3}  & ones                      & 0.60          & 0.62          & 0.63          & 0.61          & 0.67          & 0.66          & 0.65          & ---           & 3.8           & 3.6           & 1.4           & 0.2          & 1.4          & 3.3           & 2.3          & ---          \\
                              & normal                    & 0.61          & 0.62          & 0.63          & 0.62          & 0.66          & 0.67          & 0.64          & ---           & 2.3           & 2.0           & 1.3           & 0.4          & 2.7          & 3.6           & 4.2          & ---          \\
                              & uniform                   & 0.63          & 0.65          & 0.65          & 0.64          & \textbf{0.68} & 0.68          & \textbf{0.66} & ---           & 2.7           & 2.4           & 0.6           & \textbf{0.1} & 0.6          & 1.5           & 2.7          & ---          \\
                              & zeros                     & \textbf{0.66} & \textbf{0.66} & \textbf{0.67} & \textbf{0.66} & \textbf{0.68} & \textbf{0.69} & 0.63          & ---           & \textbf{0.8}  & \textbf{0.6}  & \textbf{0.2}  & \textbf{0.1} & \textbf{0.3} & \textbf{0.6}  & \textbf{1.2} & ---          \\
            \hline
                                \multirow{4}{*}{EfficientNet} & ones                      & 0.64          & 0.63          & 0.65          & 0.65          & 0.64          & 0.61          & 0.59          & 0.60          & \textbf{2.0}  & 2.2           & 2.4           & 2.2          & 7.4          & 13.0          & 16.1         & 10.5         \\
                              & normal                    & 0.60          & 0.63          & 0.65          & 0.65          & \textbf{0.66} & \textbf{0.66} & \textbf{0.65} & 0.59          & 6.0           & 3.0           & 2.4           & 2.8          & 5.5          & 6.7           & 7.4          & 12.6         \\
                              & uniform                   & \textbf{0.66} & \textbf{0.66} & \textbf{0.67} & \textbf{0.68} & 0.65          & 0.64          & 0.61          & \textbf{0.61} & 3.6           & 2.5           & 3.5           & 2.7          & 7.9          & 9.0           & 14.2         & 10.1         \\
                              & zeros                     & 0.63          & \textbf{0.66} & \textbf{0.67} & 0.67          & 0.65          & 0.64          & 0.63          & 0.55          & 4.5           & \textbf{0.8}  & \textbf{0.8}  & \textbf{0.7} & \textbf{2.0} & \textbf{2.1}  & \textbf{2.6} & \textbf{2.7} \\
            \hline
                            \multirow{4}{*}{DETR}         & ones                      & 0.61          & 0.34          & 0.55          & 0.44          & 0.20          & 0.02          & 0.25          & 0.65          & \textbf{0.9}  & 59.1          & 24.2          & 31.9         & 69.4         & 96.6          & 62.5         & 0.7          \\
                              & normal                    & 0.75          & \textbf{0.66} & 0.66          & 0.64          & 0.66          & 0.62          & 0.60          & 0.61          & 3.6           & 7.1           & 7.2           & 6.3          & 5.5          & 9.3           & 9.7          & 6.0          \\
                              & uniform                   & 0.75          & 0.47          & 0.60          & 0.58          & 0.54          & 0.41          & 0.52          & 0.65          & 2.5           & 41.3          & 17.9          & 17.4         & 22.1         & 37.9          & 22.3         & 1.2          \\
                              & zeros                     & \textbf{0.77} & \textbf{0.66} & \textbf{0.70} & \textbf{0.70} & \textbf{0.70} & \textbf{0.68} & \textbf{0.68} & \textbf{0.66} & \textbf{0.9}  & \textbf{1.1}  & \textbf{0.8}  & \textbf{0.6} & \textbf{0.8} & \textbf{1.2}  & \textbf{0.8} & \textbf{0.2} \\
            \hline
                            \multirow{4}{*}{ViT}          & ones                      & 0.66          & 0.66          & 0.66          & 0.64          & 0.59          & 0.55          & ---           & ---           & 2.3           & 1.7           & 6.1           & 7.6          & 15.3         & 17.0          & ---          & ---          \\
                              & normal                    & 0.66          & 0.66          & 0.64          & 0.63          & 0.60          & \textbf{0.58} & ---           & ---           & 3.3           & 6.2           & 9.7           & 9.5          & 12.8         & 13.5          & ---          & ---          \\
                              & uniform                   & 0.67          & 0.68          & \textbf{0.67} & 0.64          & 0.59          & 0.55          & ---           & ---           & 1.8           & 2.0           & 5.3           & 7.5          & 13.6         & 16.3          & ---          & ---          \\
                              & zeros                     & \textbf{0.68} & \textbf{0.69} & \textbf{0.67} & \textbf{0.66} & \textbf{0.62} & 0.55          & ---           & ---           & \textbf{1.2}  & \textbf{1.7}  & \textbf{3.4}  & \textbf{3.3} & \textbf{6.8} & \textbf{10.2} & ---          & ---          \\
            \hline
                                \multirow{4}{*}{SWIN}         & ones                      & \textbf{0.18} & 0.30          & 0.28          & 0.51          & \textbf{0.62} & ---           & ---           & ---           & \textbf{54.3} & 44.8          & 58.1          & \textbf{3.1} & 9.2          & ---           & ---          & ---          \\
                              & normal                    & 0.10          & 0.38          & \textbf{0.51} & 0.64          & 0.53          & ---           & ---           & ---           & 71.7          & \textbf{23.4} & \textbf{18.3} & 9.7          & 19.6         & ---           & ---          & ---          \\
                              & uniform                   & 0.11          & 0.36          & 0.43          & 0.62          & 0.56          & ---           & ---           & ---           & 75.6          & 35.2          & 38.6          & 7.4          & 17.0         & ---           & ---          & ---          \\
                              & zeros                     & 0.02          & \textbf{0.41} & 0.35          & \textbf{0.65} & 0.58          & ---           & ---           & ---           & 96.7          & 24.9          & 47.6          & 9.8          & \textbf{6.7} & ---           & ---          & ---          \\
            \hline
	\end{tabular}
	\caption{Ablation study for different initialization strategies of \locVects{} optimization (cf.~\cref{sec:experiments-ablation}). Models and layers are chosen as described in \cref{tab:layers-ablation}. Best values in \textbf{bold}.}
	\label{tab:ablation-total-init}
\end{table*}

\begin{table*}[tbp]
	\centering
	\fontsize{8pt}{10pt}\selectfont
	\setlength{\tabcolsep}{2.5pt} 
	\begin{tabular}{@{}|c|c|cccccccc|cccccccc|@{}}
		\hline
		\multirow{2}{*}{\textbf{Model}} & \multirow{2}{*}{\textbf{Resolution}} & \multicolumn{8}{c|}{\textbf{IoU}} & \multicolumn{8}{c|}{\textbf{Optimization failures}, \%} \\
		                               &                    & $l_1$       & $l_2$       & $l_3$       & $l_4$       & $l_5$       & $l_6$       & $l_7$       & $l_8$       & $l_1$       & $l_2$       & $l_3$       & $l_4$       & $l_5$       & $l_6$       & $l_7$       & $l_8$       \\
            \hline
		                      \multirow{4}{*}{YOLOv5}       & $50 \times 50$              & 0.69          & 0.67          & 0.37          & 0.59          & 0.59          & 0.48          & 0.62          & 0.55          & 2.3           & 4.7          & 51.8         & 20.3         & 19.0         & 25.7         & 7.6          & 13.5         \\
                              & $100 \times 100$            & 0.70          & \textbf{0.69} & 0.55          & 0.69          & 0.70          & 0.62          & \textbf{0.65} & 0.59          & 0.4           & 1.7          & 22.2         & 5.3          & 3.1          & 4.5          & 2.1          & 5.0          \\
                              & $150 \times 150$            & \textbf{0.71} & 0.68          & 0.60          & 0.71          & 0.71          & \textbf{0.64} & \textbf{0.65} & \textbf{0.60} & 0.1           & 0.5          & 11.7         & 2.0          & 0.8          & 1.2          & 0.9          & 2.6          \\
                              & $200 \times 200$            & \textbf{0.71} & 0.68          & \textbf{0.62} & \textbf{0.72} & \textbf{0.72} & \textbf{0.64} & 0.64          & 0.59          & \textbf{0.0}  & \textbf{0.2} & \textbf{6.5} & \textbf{1.1} & \textbf{0.1} & \textbf{0.4} & \textbf{0.6} & \textbf{1.8} \\
            \hline
                                \multirow{4}{*}{MobileNetV3}  & $50 \times 50$              & 0.65          & \textbf{0.67} & \textbf{0.67} & \textbf{0.67} & \textbf{0.68} & \textbf{0.69} & \textbf{0.65} & ---           & 4.2           & 3.2          & 1.5          & 0.3          & 1.9          & 2.2          & 3.1          & ---          \\
                              & $100 \times 100$            & \textbf{0.66} & 0.66          & \textbf{0.67} & 0.66          & \textbf{0.68} & \textbf{0.69} & 0.63          & ---           & 0.8           & 0.6          & 0.2          & 0.1          & 0.3          & 0.6          & 1.2          & ---          \\
                              & $150 \times 150$            & 0.65          & 0.66          & 0.65          & 0.65          & \textbf{0.68} & 0.68          & 0.62          & ---           & 0.2           & \textbf{0.0} & \textbf{0.0} & \textbf{0.0} & \textbf{0.0} & 0.1          & 0.3          & ---          \\
                              & $200 \times 200$            & 0.64          & 0.65          & 0.64          & 0.63          & 0.67          & 0.67          & 0.60          & ---           & \textbf{0.0}  & \textbf{0.0} & \textbf{0.0} & \textbf{0.0} & \textbf{0.0} & \textbf{0.0} & \textbf{0.0} & ---          \\
            \hline
                                \multirow{4}{*}{EfficientNet} & $50 \times 50$              & 0.57          & 0.65          & 0.66          & 0.66          & \textbf{0.65} & \textbf{0.65} & \textbf{0.64} & \textbf{0.57} & 14.5          & 3.3          & 3.8          & 3.5          & 5.1          & 6.0          & 6.5          & 7.7          \\
                              & $100 \times 100$            & \textbf{0.63} & \textbf{0.66} & \textbf{0.67} & \textbf{0.67} & \textbf{0.65} & 0.64          & 0.63          & 0.55          & 4.5           & 0.8          & 0.8          & 0.7          & 2.0          & 2.1          & 2.6          & 2.7          \\
                              & $150 \times 150$            & \textbf{0.63} & 0.64          & 0.65          & 0.66          & 0.63          & 0.62          & 0.62          & 0.54          & 2.2           & \textbf{0.0} & \textbf{0.0} & 0.1          & 0.6          & 0.9          & 1.4          & 1.9          \\
                              & $200 \times 200$            & \textbf{0.63} & 0.63          & 0.63          & 0.64          & 0.61          & 0.61          & 0.60          & 0.52          & \textbf{0.4}  & \textbf{0.0} & \textbf{0.0} & \textbf{0.0} & \textbf{0.0} & \textbf{0.1} & \textbf{0.3} & \textbf{1.1} \\
            \hline
                                \multirow{4}{*}{DETR}         & $50 \times 50$              & 0.72          & \textbf{0.67} & 0.68          & 0.68          & 0.68          & 0.67          & 0.67          & \textbf{0.66} & 8.4           & 6.5          & 3.9          & 3.5          & 3.1          & 3.6          & 3.2          & 1.7          \\
                              & $100 \times 100$            & 0.77          & 0.66          & 0.70          & \textbf{0.70} & \textbf{0.70} & \textbf{0.68} & \textbf{0.68} & \textbf{0.66} & 0.9           & 1.1          & 0.8          & 0.6          & 0.8          & 1.2          & 0.8          & 0.2          \\
                              & $150 \times 150$            & 0.79          & 0.64          & \textbf{0.71} & \textbf{0.70} & 0.69          & \textbf{0.68} & 0.67          & \textbf{0.66} & 0.3           & 0.5          & 0.2          & 0.2          & 0.1          & 0.2          & \textbf{0.0} & \textbf{0.0} \\
                              & $200 \times 200$            & \textbf{0.79} & 0.62          & \textbf{0.71} & \textbf{0.70} & 0.69          & \textbf{0.68} & 0.67          & 0.65          & \textbf{0.0}  & \textbf{0.0} & \textbf{0.0} & \textbf{0.0} & \textbf{0.0} & \textbf{0.0} & \textbf{0.0} & \textbf{0.0} \\         
            \hline
                                \multirow{4}{*}{ViT}          & $50 \times 50$              & 0.66          & 0.67          & 0.63          & 0.64          & 0.59          & 0.52          & ---           & ---           & 4.0           & 5.9          & 11.4         & 8.6          & 13.9         & 15.6         & ---          & ---          \\
                              & $100 \times 100$            & \textbf{0.68} & \textbf{0.69} & 0.67          & \textbf{0.66} & \textbf{0.62} & 0.55          & ---           & ---           & 1.2           & 1.7          & 3.4          & 3.3          & 6.8          & 10.2         & ---          & ---          \\
                              & $150 \times 150$            & \textbf{0.68} & \textbf{0.69} & \textbf{0.68} & 0.65          & \textbf{0.62} & 0.56          & ---           & ---           & 0.5           & 0.6          & 1.6          & 1.2          & 3.9          & 7.3          & ---          & ---          \\
                              & $200 \times 200$            & 0.67          & \textbf{0.69} & 0.67          & 0.64          & \textbf{0.62} & \textbf{0.57} & ---           & ---           & \textbf{0.2}  & \textbf{0.1} & \textbf{0.3} & \textbf{0.5} & \textbf{1.5} & \textbf{4.2} & ---          & ---          \\
            \hline
                                \multirow{4}{*}{SWIN}         & $50 \times 50$              & 0.00          & 0.25          & 0.07          & 0.44          & 0.52          & ---           & ---           & ---           & 99.4          & 56.8         & 87.7         & 40.4         & 17.3         & ---          & ---          & ---          \\
                              & $100 \times 100$            & 0.02          & 0.41          & 0.35          & 0.65          & 0.58          & ---           & ---           & ---           & 96.7          & 24.9         & 47.6         & 9.8          & 6.7          & ---          & ---          & ---          \\
                              & $150 \times 150$            & 0.07          & 0.45          & 0.51          & 0.68          & \textbf{0.59} & ---           & ---           & ---           & 87.0          & 11.7         & 23.0         & 4.0          & 3.5          & ---          & ---          & ---          \\
                              & $200 \times 200$            & \textbf{0.15} & \textbf{0.46} & \textbf{0.60} & \textbf{0.69} & 0.58          & ---           & ---           & ---           & \textbf{71.7} & \textbf{3.4} & \textbf{7.8} & \textbf{1.7} & \textbf{2.1} & ---          & ---          & ---          \\ \hline
	\end{tabular}
	\caption{Ablation study for different rescaling resolutions for loss calculation in \locVects{} optimization (cf.~\cref{sec:experiments-ablation}). Models and layers are chosen as described in \cref{tab:layers-ablation}. Best values in \textbf{bold}.}
	\label{tab:ablation-total-resolution}
\end{table*}

\paragraph{Tuning \locVect{} Initialization}
In this ablation study, we test four different initialization methods for \locVect{} values:
\begin{itemize}
\item \textbf{\enquote{ones}}, where the vector is initialized with scalar values of 1 (resulting in the first projection being similar to working with the original activations); 
\item \textbf{random uniform} initialization (in the interval $[0, 1]$); 
\item \textbf{random normal} distribution; and 
\item \textbf{\enquote{zeros}}, where the vector is initialized with scalar values of 0 (placing the vector in a neutral state, allowing weights to easily adjust in either direction).
\end{itemize}
It's important to note that with \locVect{} \enquote{zeros} initialization, optimization is still feasible because the gradients will not be zero. After applying the sigmoid function, all zeroed concept projection values will equal $0.5$.

\shortparagraph{Results}
As shown in \cref{tab:ablation-total-init}, the differences in IoU mean values across various initialization methods are minimal. Further, we observe only marginal differences in IoU standard deviation values. This leaves the optimization failure rate as the key metric to compare initialization methods. In general, \Result{the \enquote{ones} initialization method performs the worst, while the \enquote{zeros} initialization method performs the best}. Since concept vector values can be positive and negative, initializing with \enquote{ones} sets some weights far from their optimal values, seemingly often leading to excessive failures. In contrast, \enquote{zeros} initialization provides a more balanced starting point, resulting in fewer failures and often the best performance.
Additionally, \Result{random uniform initialization performs worse than normal distribution}. Similar to the \enquote{ones} case, the weights are skewed in a positive direction. We hypothesize that the normal distribution performs worse than \enquote{zeros} because, due to randomness, the initial weight positions can be opposite to the optimal direction.

\paragraph{Tuning \locVect{} Projection Resolution}
The resolution chosen for reshaping activations and target segmentations significantly impacts the computational cost of optimizing \locVect{}. It's important to recognize that the optimal settings can differ between datasets due to variations in factors such as the distribution of relative segmentation sizes within images and the nature of the concepts involved. The following ablation results are tailored to MS COCO, a dataset that includes samples with very small concept segmentations.


Our goal is to find a balance between computational efficiency and performance across a wide range captured by four different resolutions: (1) $50 \times 50$, which serves as our baseline, (2) $100 \times 100$, requiring \texttt{4x} computational effort compared to that baseline, (3) $150 \times 150$, demanding \texttt{9x} computation, and (4) $200 \times 200$, which involves \texttt{16x} computation.

As shown in \cref{tab:ablation-total-resolution}, similar to the previous ablation experiment, the mean IoU differences (and their standard deviations) are minimal, so we focus on the optimization failure rate as the key metric. 
In all tested models, the \Result{performance at a $50 \times 50$ resolution is notably worse} than at higher resolutions, with a higher frequency of unsuccessful optimization cases. This underperformance is likely due to the very small target concept segmentations in MS COCO; after reshaping, some target masks may end up with zero \enquote{activated} pixels (i.e., no pixels with a value of 1). This makes further optimization pointless since the final IoU will always be 0. The \Result{differences in performance among the $100 \times 100$, $150 \times 150$, and $200 \times 200$ resolutions are minimal}, though $200 \times 200$ does perform slightly better. However, this marginal improvement comes at a substantial computational cost---requiring \texttt{4x} more computation than $100 \times 100$ and \texttt{1.8x} more than $150 \times 150$. Based on these observations, we select $100 \times 100$ as the optimal resolution for further experiments, balancing performance and computational efficiency.

\paragraph{Discussion of Optimization Failures}
%
%
We identified two primary reasons for \locVect{} optimization failures, i.e., very low performance after optimization:
\begin{itemize}
    \item \textbf{Small target segmentations}: Failures often occur when either (a) the resized binary concept segmentation contains no pixels with a value of 1, or (b) the test model was trained on samples of a different relative size, making it difficult for the model to identify very small objects. 
    
    \textit{Mitigation}: These issues can be mitigated by properly selecting the resize resolution for activations and target masks and carefully choosing the probing data.


    \item \textbf{Semantic complexity of the sample}: Failures can also arise due to (a) labeling errors or (b) significant differences in how the concept is represented within the sample.
    
    \textit{Mitigation}: These problems require corrections to the labeling and a careful selection of test data (see \cref{sec:applications-outliers}).


\end{itemize}
The remaining optimization failures can be attributed to issues in linear separability (the concept pixel vectors cannot be linearly separated) or the non-convexity of the chosen optimization procedure, which, in rare cases, may get stuck in local minima.

\subsubsection{Further Influence Factors on \locVect{} Performance}
\label{sec:experiments-performance}

This section uses the previously determined optimal settings to provide detailed qualitative and quantitative assessments of the performance of \Vect{} concept segmentation. Other than before, results here are broken down by \textbf{concept} (respectively concept category) for the different \textbf{models and layers}.
The main findings are that \Result{consistently good performance can be obtained across concepts and models}, and that the \Result{influence of layer choice is consistent with known model peculiarities}.

\paragraph{Results: Qualitative Performance}
\cref{fig:gcpv-optimization-results-1} (and \cref{fig:gcpv-optimization-results-2}) display the qualitative results of \locVect{} optimization, showing the projections of \locVects{} and their IoU scores compared to target concept segmentations (for MS COCO classes and the Capybara Dataset).
Note that the visual difference between IoU scores of 0.4 and 0.8 is barely noticeable to the human eye.

\Result{In the shallow layers of CNNs artifact patches resembling grains are often present} (e.g., \texttt{4.cv3.conv} of YOLOv5 or \texttt{features.4.2} of EfficientNet in \cref{fig:gcpv-optimization-results-1}). These layers can only work with very primitive concepts when reconstructing higher-level ones, and their receptive field is limited. However, \Result{this issue diminishes in deeper layers}, which supports focusing on mid and deep layers as described in \cref{sec:setup-layers} for the MS COCO dataset.

In contrast, \Result{Vision Transformer models can effectively attribute patches in self-attention starting from the shallowest encoders} and are not restricted by receptive field (\texttt{e.ls.el3} of ViT in \cref{fig:gcpv-optimization-results-2}), avoiding the issues encountered by CNNs. The only exception is the SWIN model, which has a hierarchical structure that acts like the scope in CNNs and uses a shifting window mechanism. This makes analysis feasible only in the deeper layers. More details are provided below.


\begin{figure*}[tbh]
  \centering
  \includegraphics[width=\linewidth]{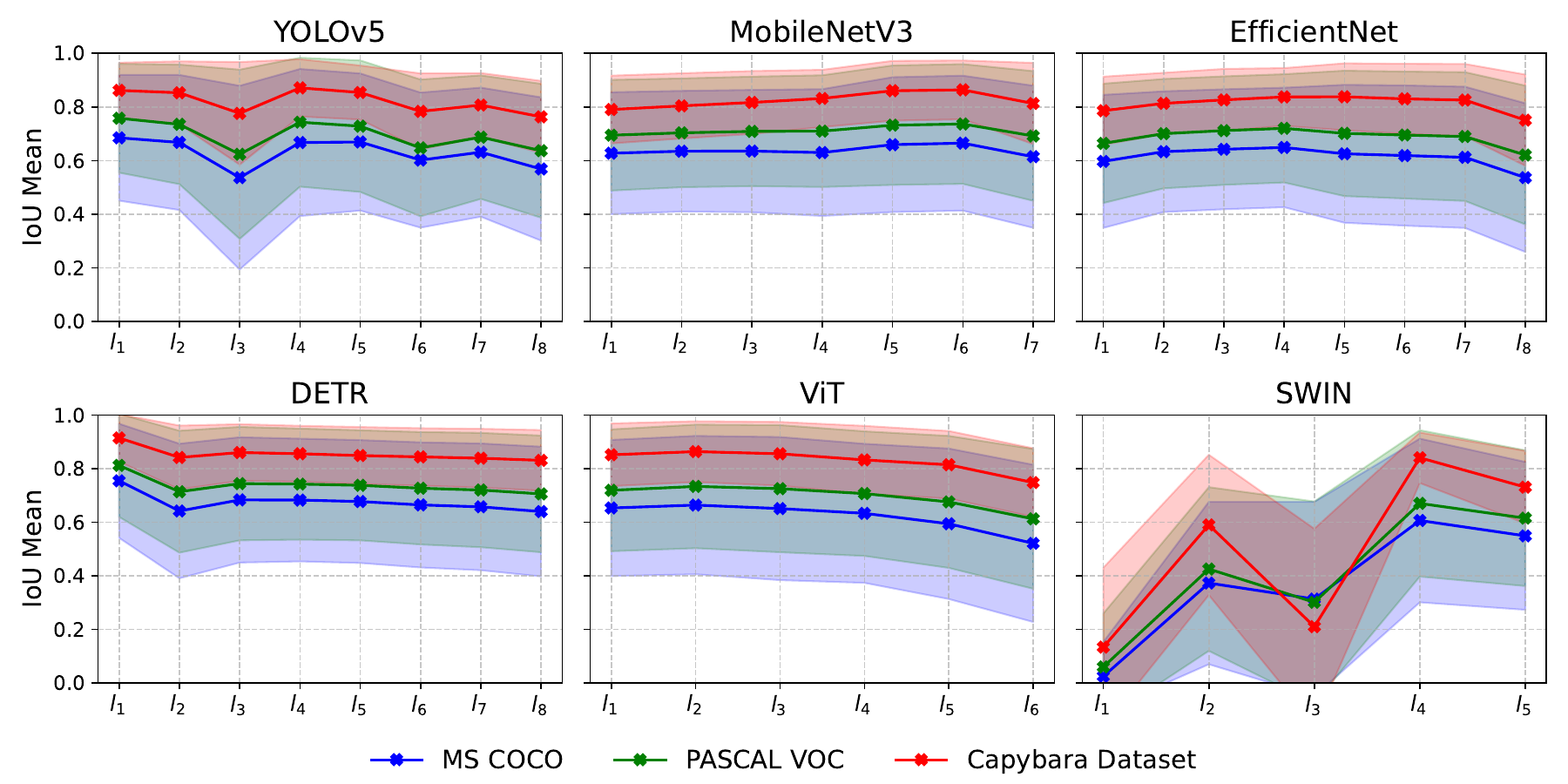}
\caption{MS COCO \& PASCAL VOC \& Capybara Dataset: Mean values and standard deviation of \locVect{} concept segmentation IoU performance across models and increasing layer depth. Layers chosen as described in \cref{tab:layers-ablation}, optimization settings as found in \cref{sec:experiments-ablation}.
  Find detailed results in \cref{tab:optimization-stats-aggregated-coco-voc-capy}.}
  \label{fig:gcpv-optimization-stats-coco-voc-capy}
\end{figure*}

\paragraph{Results: Quantitative Performance}
The quantitative results of \locVect{} optimization are summarized in \cref{fig:gcpv-optimization-stats-coco-voc-capy}, where we present the mean Intersection over Union (IoU) values and standard deviations aggregated across samples for all tested datasets. The results show that \Result{the optimization performance remains fairly consistent across layers and dependent on the data complexity}, with a few exceptions. Notably, the performance on each dataset aligns with its complexity: MS COCO, the most semantically complex dataset (featuring a high variety of object classes, challenging small object instances, and diverse scene contexts), demonstrates the lowest performance. In contrast, the Capybara dataset, characterized by relatively simple scenes with close-up object samples, achieves the highest performance. PASCAL VOC, which has moderate complexity (mix of natural and urban scenes), exhibits performance between the other two datasets.

%

\paragraph{Results: Impact of Model and Layer}
%
We found the \locVects{} concept segmentation \Result{performance results to coincide well with known peculiarities of the investigated DNNs}, as discussed in detail in the following. In particular, just as in Net2Vec \cite{schwalbe2021verification}, \Result{low resolution and low receptive field seem to have a negative impact} on \locVect{} performance.

\shortparagraph{SWIN}
SWIN is a hierarchical transformer that incrementally expands its \enquote{receptive field} (window) by incorporating surrounding patches at each stage, closely resembling the receptive field mechanism of CNNs. It is more computationally efficient than other transformers because self-attention is computed only within local windows. As a result, parts of target segmentations may initially reside in different windows. These local windows are merged as the layers deepen. This supposedly makes identifying high-level concepts in SWIN's shallow layers challenging, which coincides with our \locVects{} results ($l_1$, $l_2$ and $l_3$ in \cref{tab:optimization-stats-aggregated-coco-voc-capy}). The window splitting mechanism complicates finding a consistent set of \enquote{virtual filters} for reconstructing target segmentations, as these filters may represent different combinations in each patch. In contrast, ViT and DETR have global self-attention and do not employ hierarchical merging, allowing them to capture and reconstruct complex concepts even in shallow layers through straightforward feature processing.

\shortparagraph{YOLOv5}
In \cref{fig:gcpv-optimization-stats-coco-voc-capy} and \cref{tab:optimization-stats-aggregated-coco-voc-capy}, the reduced performance of the $l_5$ (\texttt{9.cv2.conv}) layer of YOLOv5 (the last convolutional layer of block 9, \texttt{b.9} in \cref{fig:network-architectures}) is consistent with several factors. First, (1) it operates at the lowest possible resolution (marked as \texttt{N}). Additionally, this layer is the final one in the backbone, which (2) is semantically shortcutted by earlier blocks (\texttt{b.6} and \texttt{b.8}). Finally, (3) its primary role is to transmit information to the low-resolution detection head (for very big objects, which are rare in MS COCO and can also be effectively detected by other heads) after block 23 (\texttt{b.23}). Combining these factors likely leads to less effective feature learning in this layer, reducing the overall performance of \Vect.

In contrast, YOLOv5 layers \texttt{12} and \texttt{16} ($l_4$ and $l_5$ in \cref{fig:gcpv-optimization-stats-coco-voc-capy} and \cref{tab:optimization-stats-aggregated-coco-voc-capy}) perform better because their blocks (\texttt{b.12}, \texttt{b.16}) concatenate features from different resolutions (after upscaling or downscaling), making them semantically richer.

\shortparagraph{DETR}
Aditionally, in \cref{fig:gcpv-optimization-stats-coco-voc-capy} and \cref{tab:optimization-stats-aggregated-coco-voc-capy}, in the DETR model, we observe a performance drop in $l_2$ (\texttt{m.ip}) layer, which is the \textit{input projection} layer. This layer reduces the number of output filters of ResNet50 backbone from 2048 to 256 \enquote{virtual filters} ($B \times 2048 \times H \times W \rightarrow  B \times 256 \times H \times W$), corresponding to the transformer embedding dimension, for further processing in the architecture. This reduction partially erases the information about learned features, leading to the observed performance drop.

\shortparagraph{ViT, MobileNetV3, EfficientNet}
In the remaining classifier models (see \cref{fig:gcpv-optimization-stats-coco-voc-capy}), we observe varying degrees of performance drop in the last, pre-head layers. This again coincides with a steady drop in resolution along the layers, but can also be explained by the following: (1) the classification task does not prioritize object localization, so only the attribution of certain features is emphasized; and (2) these models are trained on ImageNet-1k, which consists of object-level images, leading to smaller features potentially being ignored in the deeper layers.

\subsubsection{Robustness of \locVect{} to Noise}
\label{sec:ablation-noise}

\begin{figure*}[tbh]
  \centering
  \includegraphics[width=\linewidth]{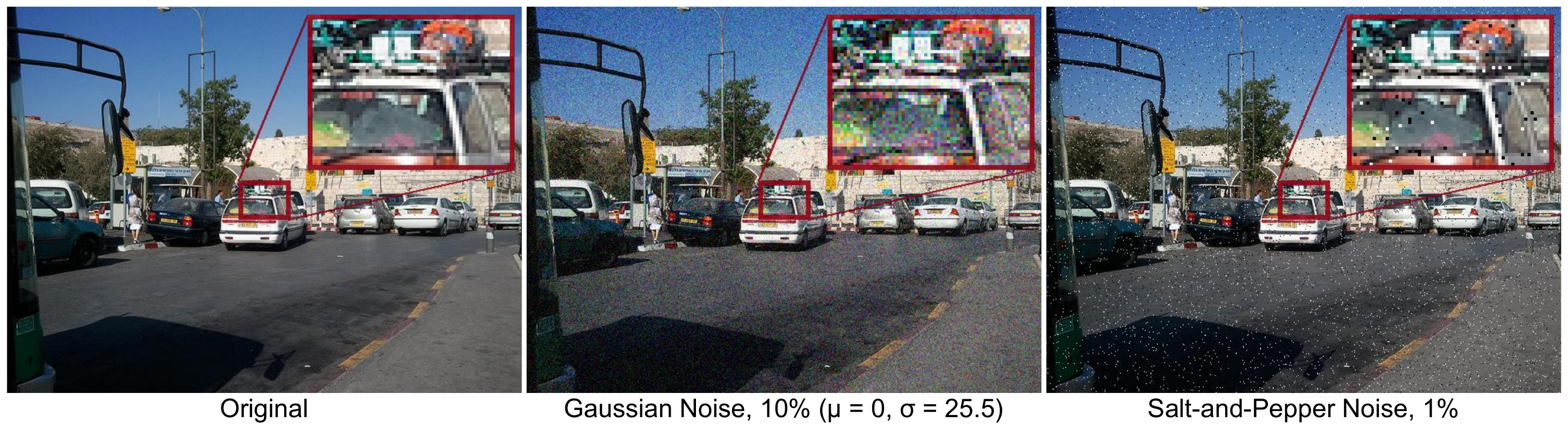}
  \caption{An image from the MS COCO dataset (\textit{left}) alongside its distorted versions: with 10\% Gaussian noise ($\mu=0$, $\sigma=25.5$; \textit{center}) and with 1\% Salt-and-Pepper noise (\textit{right}).
  A magnified patch from the highlighted region is shown in the top-right corner of each image to illustrate noise effects in detail.}
  \label{fig:noise-examples}
\end{figure*}

\begin{table*}
\centering
\fontsize{8pt}{10pt}\selectfont
\setlength{\tabcolsep}{3pt}
\caption{Robustness of \locVects{} to noise: mean and standard deviation values of NCC (\cref{eq:cross-correlation}) (mean-centered cosine similarities) between the \locVects{} of clean and noisy images. Gaussian Noise with $\mu = 0$ and $\sigma$ selected as $1\%$, $3\%$, $5\%$, and $10\%$ of the maximum pixel value (255). Salt-and-Pepper Noise replaces a specified percentage of pixels: half with white and half with black. Layers of models are chosen as described in \cref{tab:layers}.}
\label{tab:ablation-robustness}
\begin{tabular}{@{}|c|c|cccc|cccc|@{}}
\hline
\multirow{2}{*}{\textbf{Model}} & \multirow{2}{*}{\textbf{Layer}} & \multicolumn{4}{c|}{\textbf{Gaussian Noise} ($\mu = 0$)} & \multicolumn{4}{c|}{\textbf{Salt-and-Pepper Noise}} \\
&            &   $\sigma=2.55$ &   $\sigma=7.65$ &   $\sigma=12.75$ &  $\sigma=25.5$ &   0.1\% &  0.25\% &   0.5\% &   1.0\% \\
\hline
      \multirow{3}{*}{YOLOv5} &  $l_{mid}$ & 0.99±0.01 & 0.92±0.05 & 0.85±0.07 & 0.71±0.10 & 0.92±0.07 & 0.85±0.08 & 0.79±0.09 & 0.72±0.11 \\
                              & $l_{deep}$ & 0.99±0.02 & 0.90±0.08 & 0.82±0.10 & 0.67±0.14 & 0.83±0.14 & 0.74±0.17 & 0.67±0.18 & 0.59±0.18 \\
                              & $l_{last}$ & 0.97±0.04 & 0.84±0.09 & 0.74±0.11 & 0.58±0.13 & 0.77±0.15 & 0.67±0.16 & 0.60±0.16 & 0.53±0.15 \\
\hline
 \multirow{3}{*}{MobileNetV3} &  $l_{mid}$ & 0.97±0.03 & 0.88±0.08 & 0.80±0.10 & 0.67±0.12 & 0.92±0.07 & 0.85±0.10 & 0.78±0.11 & 0.71±0.12 \\
                              & $l_{deep}$ & 0.97±0.04 & 0.87±0.07 & 0.79±0.09 & 0.64±0.11 & 0.91±0.06 & 0.84±0.09 & 0.77±0.10 & 0.68±0.11 \\
                              & $l_{last}$ & 0.97±0.03 & 0.85±0.07 & 0.76±0.09 & 0.59±0.10 & 0.84±0.10 & 0.77±0.10 & 0.69±0.11 & 0.58±0.12 \\
\hline
\multirow{3}{*}{EfficientNet} &  $l_{mid}$ & 0.96±0.06 & 0.85±0.11 & 0.79±0.11 & 0.69±0.10 & 0.87±0.08 & 0.80±0.10 & 0.74±0.11 & 0.67±0.11 \\
                              & $l_{deep}$ & 0.95±0.07 & 0.83±0.10 & 0.76±0.09 & 0.65±0.09 & 0.80±0.10 & 0.75±0.10 & 0.69±0.10 & 0.62±0.10 \\
                              & $l_{last}$ & 0.95±0.06 & 0.81±0.09 & 0.74±0.09 & 0.62±0.09 & 0.79±0.10 & 0.73±0.10 & 0.67±0.10 & 0.59±0.11 \\
\hline
        \multirow{3}{*}{DETR} &  $l_{mid}$ & 0.97±0.02 & 0.87±0.06 & 0.78±0.07 & 0.62±0.09 & 0.84±0.09 & 0.74±0.10 & 0.64±0.11 & 0.53±0.12 \\
                              & $l_{deep}$ & 0.97±0.03 & 0.87±0.07 & 0.78±0.09 & 0.62±0.11 & 0.74±0.15 & 0.63±0.16 & 0.51±0.16 & 0.40±0.16 \\
                              & $l_{last}$ & 0.98±0.03 & 0.93±0.06 & 0.88±0.08 & 0.79±0.11 & 0.86±0.11 & 0.80±0.14 & 0.71±0.16 & 0.60±0.18 \\
\hline
         \multirow{3}{*}{ViT} &  $l_{mid}$ & 1.00±0.00 & 0.98±0.01 & 0.95±0.04 & 0.85±0.09 & 0.97±0.03 & 0.93±0.06 & 0.88±0.09 & 0.81±0.10 \\
                              & $l_{deep}$ & 0.99±0.01 & 0.96±0.03 & 0.91±0.06 & 0.79±0.09 & 0.93±0.06 & 0.87±0.08 & 0.82±0.09 & 0.75±0.09 \\
                              & $l_{last}$ & 1.00±0.00 & 0.97±0.03 & 0.93±0.05 & 0.84±0.08 & 0.95±0.05 & 0.91±0.06 & 0.87±0.07 & 0.82±0.08 \\
\hline
        \multirow{2}{*}{SWIN} & $l_{deep}$ & 0.91±0.06 & 0.83±0.07 & 0.77±0.08 & 0.66±0.10 & 0.88±0.07 & 0.83±0.08 & 0.78±0.08 & 0.72±0.09 \\
                              & $l_{last}$ & 0.79±0.09 & 0.67±0.09 & 0.59±0.10 & 0.47±0.11 & 0.74±0.09 & 0.67±0.10 & 0.61±0.10 & 0.53±0.11 \\
\hline
\end{tabular}
\end{table*}

The robustness evaluation results show that \Result{\locVects{} are consistently robust even against large amounts of noise}. However, \Result{robustness is slightly dependent on model type and layer depth}. For estimation, we used the Normalized Cross-Correlation (NCC, \cref{eq:cross-correlation}) between the \locVects{} of clean and noisy images. The test was conducted on 950 randomly selected MS COCO images (50 samples per category). Two noise types were applied to the input images (segmentations for \locVects{} stayed unchanged):

\begin{itemize}
    \item \textbf{Gaussian Noise} (\cref{fig:noise-examples}, \textit{center}), which distorts the content by adding random variations, blurring fine details while keeping the overall structure, and
    \item \textbf{Salt-and-Pepper Noise} (\cref{fig:noise-examples}, \textit{right}), which erases information by randomly replacing pixels with extreme values (black or white), leading to localized loss of content and visual discontinuities.
\end{itemize}

\paragraph{Results: Gaussian Noise}
Gaussian noise was added with a mean ($\mu$) of 0 and standard deviations ($\sigma$) of $1\%$, $3\%$, $5\%$, and $10\%$ ($2.55$, $7.65$, $12.75$, and $25.5$, respectively) of the maximum pixel value (255).

For Gaussian noise with increasing standard deviation ($\sigma$), we observe a gradual decline in NCC values across all models and layers (\cref{tab:ablation-robustness}). The mid-level layers ($l_{mid}$), \Result{the most shallow layers among those tested, consistently exhibit the highest robustness}, maintaining higher NCC values even at $\sigma = 25.5$ (10\% of the maximum pixel value). In contrast, the deeper ($l_{deep}$) and final layers ($l_{last}$) show a sharper decrease in robustness, with the $l_{last}$ layer being the most affected. This can be attributed to the accumulation of noise as it propagates through the network: deeper layers involve more operations, such as multiplications and non-linear transformations, which amplify the effects of input noise. Consequently, the signal-to-noise ratio diminishes with depth, reducing robustness in these layers.

\Result{For Gaussian noise, \locVects{} demonstrate robustness even at significant noise levels}, with NCC values generally exceeding 0.6 for a standard deviation of up to $10\%$ of the maximum pixel value ($\sigma = 25.5$). This suggests that even with considerable distortions, \locVects{} remain stable and retain high similarity to their clean counterparts. 

\paragraph{Results: Salt-and-Pepper Noise}
In this experiment, a specified percentage of pixels in the input images was randomly replaced with extreme values: half of the replaced pixels were set to white, and the other half to black. The noise levels (replacement rates) tested were 0.1\%, 0.25\%, 0.5\%, and 1.0\% of the total image pixels.

In results (\cref{tab:ablation-robustness}), we observe a similar trend: increasing noise levels lead to a gradual decrease in NCC. Here, mid-level layers ($l_{mid}$) generally outperform deeper and last layers ($l_{deep}$ and $l_{last}$) in robustness.

\Result{For Salt-and-Pepper noise, \locVects{} are robust to small amounts of pixel corruption}, maintaining NCC values around or above 0.6 for noise levels affecting up to $1\%$ of the pixels. This indicates that the feature space can effectively handle sparse, localized disruptions. 






\subsection{Complexity in DNN Concept Distributions}
\label{sec:experiments-generalization}
Here, we conduct diverse experiments to assess the occurrence of sub-concepts and concept confusion in \locVect{} distributions for different layers and DNNs.
To assess the confusion of concepts in DNN latent spaces we 
(1) we manually inspect GMM-based approximations of the distribution of dimensionality reduced \locVects{};
(2) we assess purity of larger \locVect{} neighborhoods (subclusters) obtained in hierarchical clustering; and
(3) use image retrieval techniques to measure how pure local \locVect{} neighborhoods are.
While the GMMs and hierarchical clustering already give insights into the frequency of subconcepts, this is quantified
(4) by directly comparing concept segmentation performance of \locVects{}, \subglobVects{}, and \globVects{} against each other and various C-XAI baselines. This serves as a measure for information loss when going from local to global.

Our experiments reveal a crucial insight: \Result{concept representations in neural networks are far more complex than simple point estimates, existing as intricate, often overlapping multi-modal distributions in the feature space}.
In particular, the results clearly show how state-of-the-art global vectors frequently fail to capture the complexity of concept distributions, such as sub-concepts or concept confusions.
These aspects strongly influence the alignment between human and DNN understanding, and, thus, explainability methods capturing such are needed; ours is the first one to do so.

\subsubsection{UMAP-GMM Clustering}
UMAP preserves the local density of \locVects{} distribution, offering a focused view of the data structure. Exemplary results are shown in the top-row diagrams in \cref{fig:gmm-dendrogram-detr-e1,fig:gmm-dendrogram-efficientnet-f7,fig:gmm-dendrogram-swin-f7,fig:gmm-dendrogram-detr-e1-voc,fig:gmm-dendrogram-efficientnet-f7-voc,fig:gmm-dendrogram-swin-f7-voc}. The first top-row diagrams show \locVect{} samples of different categories. The second top-row diagrams demonstrate GMMs fitted separately for samples from each concept category. In contrast, the third top-row diagrams display GMMs fitted across all samples, regardless of their labels.
Ellipses denote the $1\sigma$ regions of the fitted GMM components, and shades indicate which GMM components are dominant. The optimal number of GMM components is determined with the BIC criterion, using 1 to 40 components (40 being double the number of tested categories).

\paragraph{Results: UMAP-GMM}
A qualitative inspection of the shown and similar diagrams for other layers provided clear evidence of the complexity in concept representations:
\begin{itemize}
\item \Result{Multiple sub-concepts}: Many concepts are represented by multiple Gaussian components (cf.~consecutive numbering of components), suggesting the presence of sub-concepts within categories.
\item \Result{Overlapping distributions}: Concept categories often show significant overlap (i.e., concept confusion), indicating that simple point representations would fail to capture nuanced relationships.
\item \Result{Varying density}: The size and shape of Gaussian components vary considerably, reflecting differences in the concentration and spread of concept representations.
\end{itemize}


\subsubsection{Hierarchical Clustering}
%
Exemplary dendrograms resulting from hierarchical clustering of (mixed concept) \locVects{} are shown in the bottom row of \cref{fig:gmm-dendrogram-detr-e1,fig:gmm-dendrogram-efficientnet-f7,fig:gmm-dendrogram-swin-f7,fig:gmm-dendrogram-detr-e1-voc,fig:gmm-dendrogram-efficientnet-f7-voc,fig:gmm-dendrogram-swin-f7-voc}. Each dendrogram leaf represents a single sample, with leaves color-coded according to their respective categories. Leaves are sorted into branches based on their proximity. The number of clusters used here is identified via adaptive clustering (\cref{alg:adaptive-clustering}). In subsequent experiments, these clusters are then used to estimate \globVects{} and \subglobVects{}.

\paragraph{Qualitative Results}
As expected, the results of hierarchial clustering agree with the UMAP-GMM findings:
\begin{itemize}
\item \Result{Complex tree structures}: The dendrograms reveal intricate branching patterns, far from the simple structures expected if concepts were cleanly separable points.
\item \Result{Category confusion}: While some high-level separations are visible, many branches contain a mix of category labels, indicating complex relationships between concepts.
\item \Result{Varying cluster sizes}: The adaptive clustering algorithm often identifies a large number of clusters (see \cref{tab:gcpv-clustering-generalization}), further emphasizing the simplicity of single-point representations.
\end{itemize}
Notable observations from \cref{fig:gmm-dendrogram-detr-e1,fig:gmm-dendrogram-efficientnet-f7,fig:gmm-dendrogram-swin-f7,fig:gmm-dendrogram-detr-e1-voc,fig:gmm-dendrogram-efficientnet-f7-voc,fig:gmm-dendrogram-swin-f7-voc} are that the \ConceptTerm{animal} and \ConceptTerm{vehicle} concepts are located in distinct regions of the diagrams, being consistent with expectations of concept separation in feature space. However, within these regions, concept confusion happens: E.g., the out-of-distribution \ConceptTerm{capybara} concept is clustered with other \ConceptTerm{animal} samples and sometimes confused with \ConceptTerm{bear}, majorly represented by brown-fur bears in MS COCO. This reflects the meaningfulness of models' context-aware concept representations.

\begin{table*}
        \centering

        \setlength{\tabcolsep}{3pt} 

        \begin{tabular}{@{}|c|c|ccc|ccc|@{}}
                \hline
                \multirow{2}{*}{\textbf{Dataset}} & 
                \multirow{2}{*}{\textbf{Model}} & \multicolumn{3}{c|}{\textbf{Purity}} & \multicolumn{3}{c|}{\textbf{Clusters}} \\
                & & $l_{\text{mid}}$ & $l_{\text{deep}}$ & $l_{\text{last}}$ & $l_{\text{mid}}$ & $l_{\text{deep}}$ & $l_{\text{last}}$ \\
                \hline
                \multirow{6}{*}{MS COCO}
                        & YOLOv5 & 0.35±0.02 & \textbf{0.43}±0.02 & 0.31±0.02 & 17.3±1.3 & 17.8±1.0 & 17.5±1.0 \\
                        & MobileNetV3 & 0.26±0.01 & 0.26±0.01 & \textbf{0.30}±0.02 & 17.4±1.2 & 17.6±1.2 & 18.0±1.1 \\
                        & EfficientNet & 0.27±0.01 & 0.37±0.02 & \textbf{0.47}±0.02 & 17.5±1.2 & 18.0±1.3 & 18.7±1.2 \\
                        
                        & DETR & 0.40±0.02 & \textbf{0.48}±0.03 & 0.22±0.01 & 17.7±1.3 & 17.8±1.2 & 17.3±1.2 \\
                        & ViT & 0.25±0.01 & 0.33±0.02 & \textbf{0.45}±0.03 & 17.3±1.1 & 18.4±1.4 & 18.6±1.2 \\
                        & SWIN &  ---  & 0.46±0.02 & \textbf{0.61}±0.02 &  ---  & 18.7±1.1 & 20.8±1.4 \\
                \hline
                \multirow{6}{*}{PASCAL VOC}
                        & YOLOv5 & 0.43±0.02 & \textbf{0.56±0.02} & 0.41±0.02 & 25.3±1.4 & 25.2±1.3 & 25.1±1.3 \\
                        & MobileNetV3 & 0.25±0.01 & 0.27±0.01 & \textbf{0.38±0.02} & 24.4±1.3 & 24.9±1.2 & 25.7±1.5 \\
                        & EfficientNet & 0.28±0.01 & 0.44±0.02 & \textbf{0.60±0.02} & 24.8±1.3 & 25.7±1.3 & 26.3±1.5 \\
                        
                        & DETR & 0.52±0.02 & \textbf{0.68±0.02} & 0.30±0.01 & 25.1±1.4 & 24.8±1.1 & 24.6±1.3 \\
                        & ViT & 0.26±0.01 & 0.41±0.02 & \textbf{0.63±0.02} & 24.8±1.4 & 25.6±1.3 & 26.8±1.3 \\
                        & SWIN &  ---  & 0.55±0.02 & \textbf{0.72±0.02} &  ---  & 27.2±1.5 & 27.9±1.2 \\
                \hline                
        \end{tabular}
        \caption{MS COCO \& PASCAL VOC: Purity and Number of Clusters of Hierarchial Clustering Generalization. Estimated for 50 runs. Layer names in \cref{tab:layers}. Best results per model and dataset in \textbf{bold}.}
        \label{tab:gcpv-clustering-generalization}
\end{table*}

\paragraph{Purity Results (Concept Confusion)}
Quantiative results for purity estimation are shown in \cref{tab:gcpv-clustering-generalization}. The table aggregates 50 runs with a random sampling of category-related samples.  
We see that \Result{the total number of clusters is consistent among all networks}. 
\Result{For classifiers the last layers achieve the highest purity} ($l_{last}$ of MobileNetV3, EfficientNet, ViT, and SWIN), indicating superior concept separation. 
\Result{For object detectors the best purity is observed in deep layers} ($l_{deep}$ of DETR and YOLOv5); this is also consistent with findings from further experiments (see \cref{sec:applications-retrieval}).

MS COCO, with its higher complexity, broader category set, and more diverse scenes, exhibits lower purity scores across models compared to PASCAL VOC. In contrast, PASCAL VOC's simpler annotation structures and fewer categories allow for better concept separation, as reflected in consistently higher purity scores across all models. Thus, \Result{dataset complexity and annotation quality influence the clustering outcomes}, with simpler datasets facilitating more distinct concept grouping. Additionally, the consistency in the number of clusters between datasets demonstrates the robustness of the clustering approach despite differing data distributions.

\subsubsection{Information retrieval}
\label{sec:applications-retrieval}

As described in \cref{sec:method-applications}, a straightforward use of the concept-vs-context information encoded in \locVects{} is for image retrieval. Quality of the results can both be estimated qualitatively by analyzing retrieved samples and quantitatively via mAP@k (see \cref{eq:retrieval-mAP}).
However, mAP@k may also directly serve as an indicator for concept confusion: A low mAP@k means that \locVects{} often find \locVects{} of a different concept in their immediate neighborhood.
In our experiments, more than 20\,\% of immediate \locVect{} neighbors are of a different class.

\paragraph{Qualitative Retrieval Results}
\cref{fig:retrieval-qualitative} (also \cref{fig:retrieval-qualitative-extra1,fig:retrieval-qualitative-extra2}) demonstrate the retrieval process for in-distribution (\ConceptTerm{person}, \ConceptTerm{motorcycle}, \ConceptTerm{airplane}, \ConceptTerm{cat}) and out-of-distribution (\ConceptTerm{capybara}) samples in different models. Each figure presents a concept segmentation (which corresponds to optimized \locVects) of query samples alongside the top 5 closest (best) retrieved samples. We color code the relevance or irrelevance of the retrieved sample with green and red frames, respectively. 

The quality of retrieval is closely linked to how effectively a model learns features in its latent space. According to a qualitative inspection, \Result{all tested models in the demonstrated layers provided semantically meaningful retrieval results} (with minor exceptions). For instance, when querying the \ConceptTerm{person} concept with a surfboard, the retrieved samples consistently involved scenes related to surfing and water activities. Similarly, querying with an image of a \ConceptTerm{cat} next to a \ConceptTerm{monitor} or \ConceptTerm{laptop} led to the retrieval of images featuring cats in similar contexts, with monitors or laptops in the background.

Notably, the model successfully retrieved images with similar characteristics even with samples like \ConceptTerm{capybara}, which were not part of the model's training data. For example, querying \ConceptTerm{capybara} with a person in the background resulted in retrieving samples with capybaras and persons (see \cref{fig:retrieval-qualitative}, DETR).

The inspection can, however, also help to uncover inconsistencies between human understanding of a concept in context and the network's: When querying samples similar to one showing persons playing baseball, other sports activities not even taking place in a stadium are shown like skiing.
This illustrates how \Result{the inspection can help to find both consistencies and deviations between human understanding of concepts and DNN ones}, ultimately indicating missing examples and data augmentation needs.

\begin{figure*}[tbh]
  \centering
  \includegraphics[width=\linewidth]{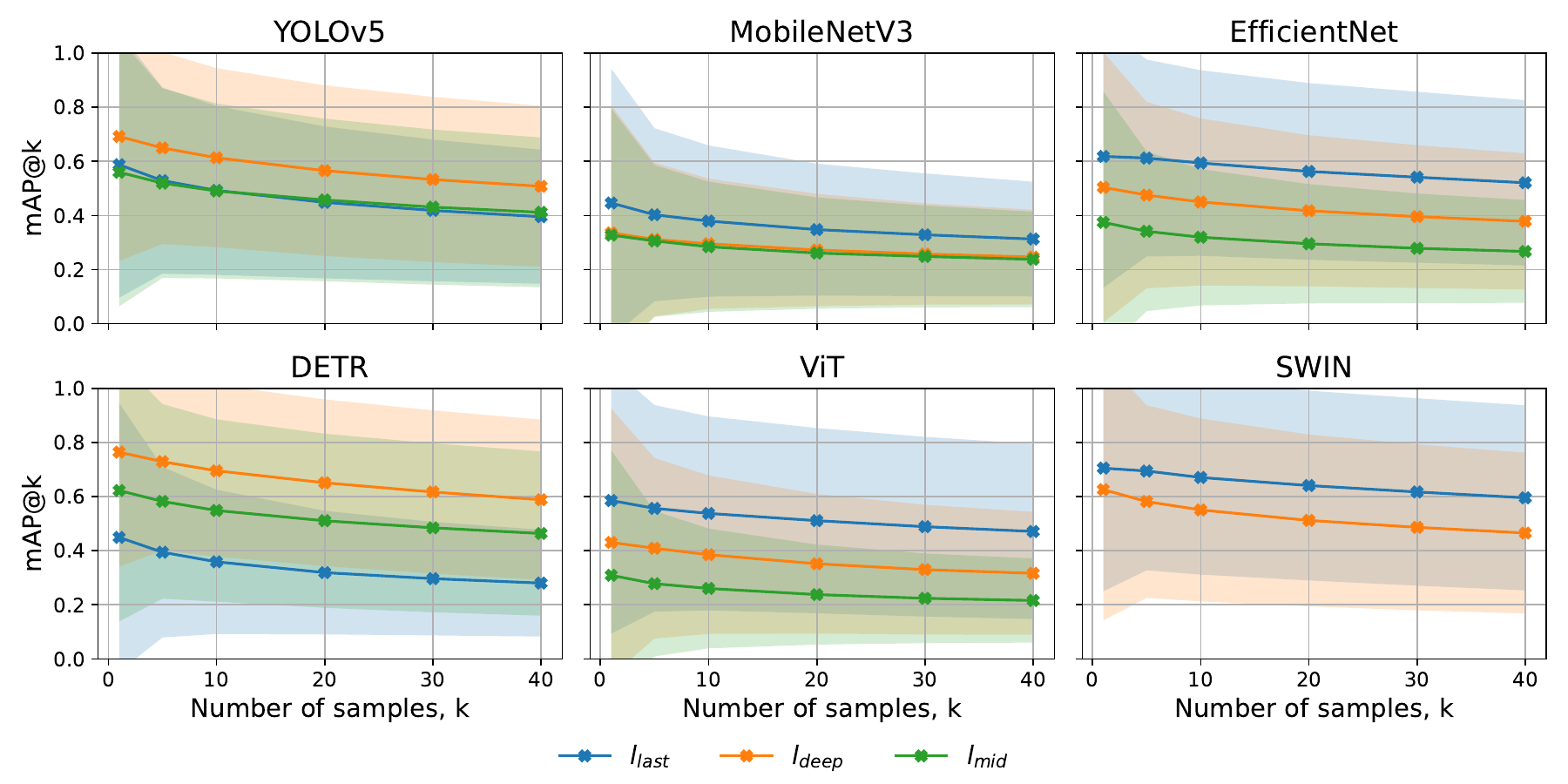}
  \caption{MS COCO \& Capybara Dataset: \locVect{} information retrieval mAP@k performance averaged across all categories in different models and increasing layer depth. Layers chosen as described in \cref{tab:layers}.
  Find detailed results in \cref{tab:retrieval-map-all}.
  }
  \label{fig:retrieval-mAP-plot}
\end{figure*}

\paragraph{Quantitative retrieval results}
Based on the qualitative retrieval observations discussed earlier, we can estimate the quality of feature space learning by calculating the mAP@k (mean Average Precision at k retrieved samples) metric. 
The mAP@k metric, particularly at higher k values, highlights how effectively concepts are separated in the feature space. Higher mAP values indicate superior retrieval quality and stronger concept differentiation within the model's latent representations: If a large amount of neighbors of each \locVect{} is from the same concept, concepts are both well separated, and mAP is high.
In our experiments, we calculated mAP@k for each of the 3,212 \locVects{} using them as queries against the remaining samples.


\cref{fig:retrieval-mAP-plot} display mAP@k curves (with standard deviations) for various typical values of $k \in \{1, 5, 10, 20, 30, 40\}$, aggregated across all categories. Additionally, mAP@k values are summarized in \cref{tab:retrieval-map-all} (for $k \in \{5, 10, 20, 40\}$), categorized by different super-categories. The curves are generalized across all tested categories and layers for each model. The color scheme---blue for the last tested layer $l_{\text{last}}$, green for the middle one $l_{\text{mid}}$, and orange for the deep layer $l_{\text{deep}}$---corresponds to the setup outlined in \cref{tab:layers}. 

In all cases, we observe the typical behavior of information retrieval curves: \Result{mAP@k gradually decreases as k increases}. 
%
For classifiers like SWIN, ViT, EfficientNet, and MobileNetV3, we observe that the last layer $l_{\text{last}}$ (blue line) typically performs the best. In contrast, object detectors like DETR and YOLOv5 exhibit different behavior due to their complex structures and block connections. In these detectors, the intermediate layer $l_{\text{deep}}$ (orange line) often outperforms others (see extended analysis in \cref{sec:applications-shortcut-detection}).

These quantitative estimations of \Result{retrieval using \locVects{} provide a meaningful numerical comparison of concept separation in feature spaces across models}. 
They might also help to identify the effectiveness of feature space learning within each model and assist in selecting the layers that possess the most relevant semantic knowledge.

\subsubsection{\Vects{} versus Baselines}
\label{sec:experiments-baselines}

\begin{figure*}[tbh]
  \centering
  \includegraphics[width=\linewidth]{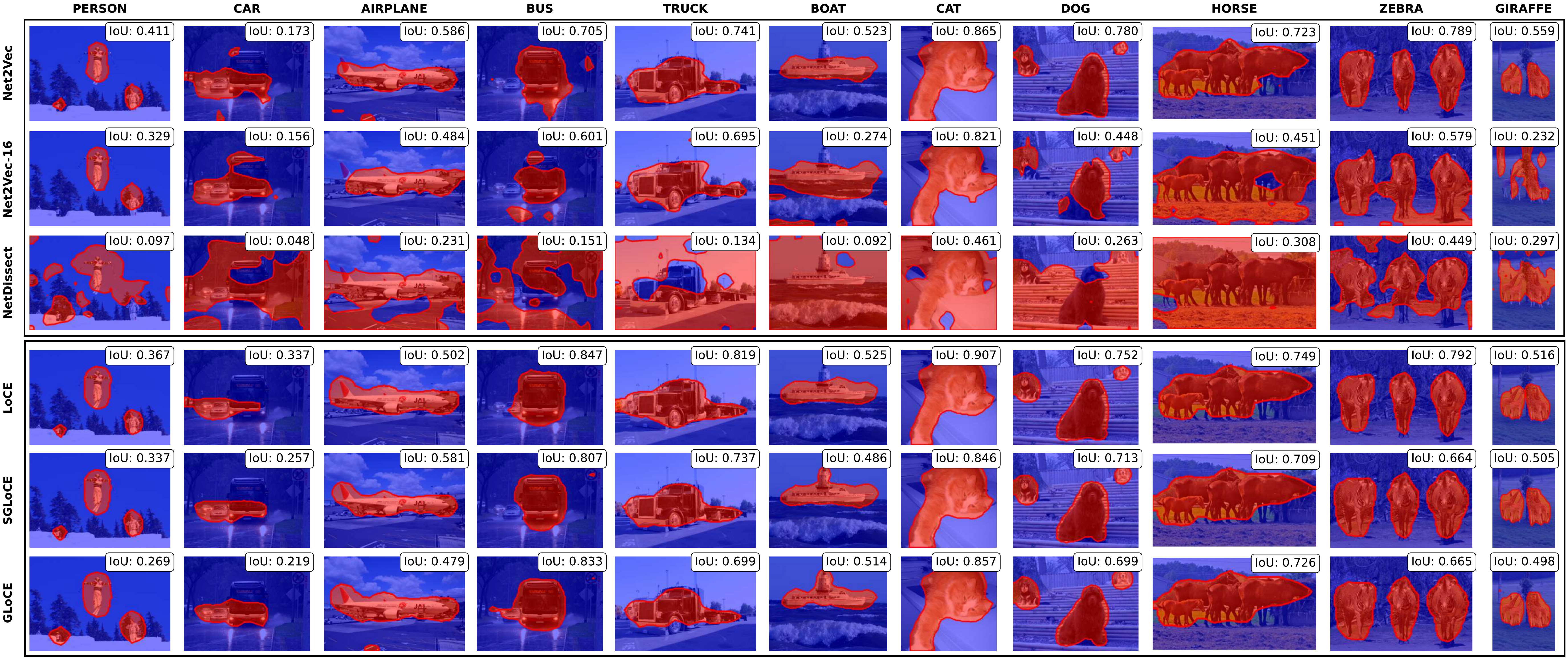}
  \caption{Concept segmentation of \locVects{}, \globVects{}, \subglobVects{} versus Baselines for EfficientNet features.7.0}
  \label{fig:gcpv-vs-baselines-efficientnet-f.7.0}
\end{figure*}

\begin{figure*}
  \centering
  \includegraphics[width=\linewidth]{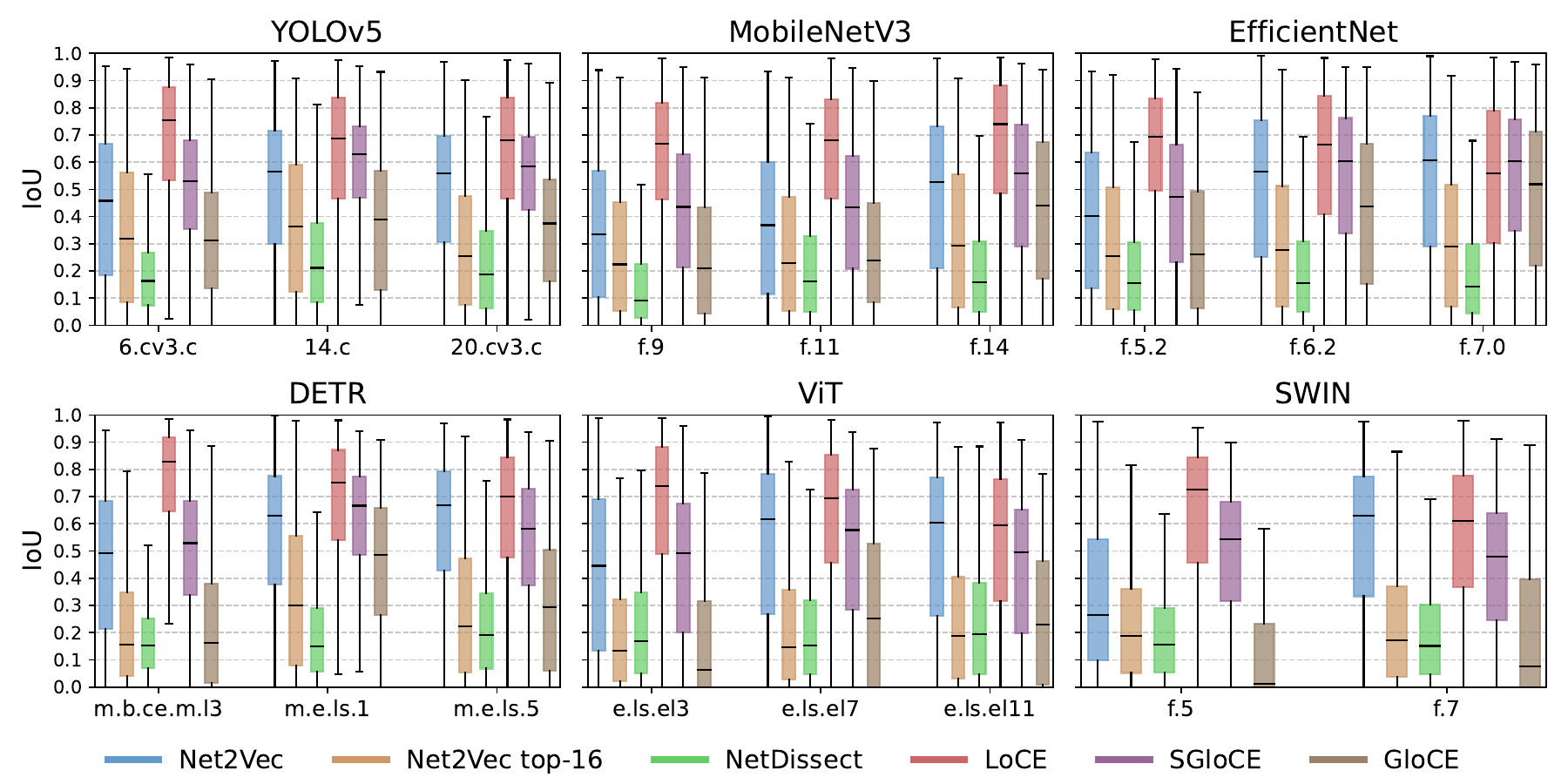}
  \caption{MS COCO: Concept segmentation performance comparison between the generalizations of \locVects{} to its (sub-)global variants (best matching \subglobVects{}, \globVects{}, see \cref{eq:generalization-centroid}), and direct global baseline methods (see \cref{sec:setup-baselines}).
  For reference, the average performance of individual \locVects{} on their respective training images is shown, marking the maximally achievable segmentation performance of our (sub-)global variants (in red).
  Layers are chosen as described in \cref{tab:layers}.
  Find detailed results in \cref{tab:gcpv-vs-baselines-final}.
  }
  \label{fig:gcpv-vs-baselines}
\end{figure*}

While previous results are promising, we want to validate how much information gets lost when considering local (context-specific) features instead of directly obtaining globally applicable feature vectors. For this, we compare the performance of local \locVect{} and its step-wise generalizations (\subglobVect{} and \globVect{}) against global C-XAI methods like Net2Vec, Net2Vec-16, and NetDissect. 
In this ordering, \locVects{} (which can only be tested on their respective local sample) are expected to contain the maximum amount of information about concept features in their image.
Their generalizations have larger scopes and are supposed to yield good results still on samples from the same sub-cluster (\subglobVects{}) respectively any sample.
The performance drop when going from local (\locVects{}) to global (\globVects{}), and comparison against directly global (Net2Vec etc.) methods shall give insights into the loss of information in the generalization process.
The main results are that \Result{an expected drop in performance occurs the more global one goes},
but that still \Result{\subglobVects{} and \globVects{} show competitive results to state-of-the-art}.


For fairness in comparison, we apply the same strategy for projection evaluation to NetDissect as we do with \Vect{} and Net2Vec, using sigmoid activation followed by thresholding at 0.5 for projection visualization.

\paragraph{Qualitative Results}
As shown in \cref{fig:gcpv-vs-baselines-efficientnet-f.7.0}, we observe concept projections in the last layer \texttt{features.7.0} (\texttt{f.7.0}) of EfficientNet. The \Result{visual difference between \locVect{}, Net2Vec, and \subglobVect{} is minimal}, with all three methods demonstrating strong performance. However, \Result{\globVect{} and Net2Vec-16 perform worse}. \globVect{} tends to overgeneralize by retaining only the most significant vector directions, erasing finer details during regularization.
This makes its performance similar to Net2Vec-16, which undergeneralizes compared to standard Net2Vec due to sparsity. Sparsity also leads to the loss of finer details. NetDissect shows the poorest performance, highlighting the \Result{oversimplification of searching a concept in a single filter}.

\paragraph{Quantitative Results}
\cref{fig:gcpv-vs-baselines} presents boxplots with the mean and standard deviations of IoU values aggregated for all categories and for all tested methods. As expected, in the shallower layers, \locVect{} substantially outperforms all baselines, possibly by leveraging more primitive context-specific features to discriminate higher-level concepts against their background. The \Result{performance of \subglobVect{} is consistently comparable to, or slightly better than, Net2Vec}. \Result{\globVect{} performs similarly to Net2Vec-16 but falls short compared to Net2Vec}. These results could be explained by the concept distribution's complexity and potential non-linearity in the feature space. Variations in performance may be attributed to the heterogeneity of certain concepts.
In particular, the drop in performance going from local to global remains as expected, not relevantly underperforming the non-context-sensitive baseline. Thus, one can conclude that \Result{relevant information can both be obtained using \locVects{}, and pertained to some extend when generalizing to \subglobVects{} and \globVects{}}.

As an application, by comparing the performance of global methods like Net2Vec and \globVect{}, we can infer how well a concept is generalized and isolated in the feature space. If a concept is densely packed within a specific region, the performance of Net2Vec and \globVect{} should be very similar. Ideally, separate, non-aggregated concepts shall be considered.

\section{Using Distributions: \locVects{} for Debugging}
\label{sec:applications}

In this section, we explore applications of \Vects{} for debugging DNNs. We uncover several key insights: \Result{concept representations in latent spaces are non-unimodal and overlapping}, leading to concept confusion, which can be identified. We also \Result{detect sub-concepts}, or variations within a concept, based on differences in concept distribution density, as well as \Result{concept-level outliers}, derived from density variations. Additionally, we demonstrate how \Vects{} can be used for \Result{concept-level information retrieval}, offering a quantitative measure of concept \enquote{purity} or \enquote{separation}. These findings are further validated by observing the \Result{alignment with the performance of DNN structural blocks} in the tested networks.

\subsection{Identification of sub-concepts}
\label{sec:applications-sub-concepts}

Previous experiments showed that \locVects{} exhibit varying distribution densities, with similar samples clustering more closely together. 
Sub-clusters of \locVects{} from a single concept can reveal learned sub-concepts---different feature sets that can individually make up the concept.
As an example, \cref{fig:subconcepts,fig:subconcepts-extra} illustrate how different networks exhibit sub-concepts of the concept \ConceptTerm{car} such as \ConceptTerm{proximate}, \ConceptTerm{mid-distance}, and \ConceptTerm{distant}. These sub-concepts were identified by clustering samples of the \ConceptTerm{car} category using dendrograms and then evaluating \subglobVects{} for these clusters (see \cref{eq:generalization-centroid}). We used linkage thresholding for the cluster selection (manual threshold value selection). The masks shown in the figure were generated by projecting activations with these \subglobVects{}.

We successfully identified the same three sub-concepts for object detectors (DETR, YOLOv5). However, for classifiers (MobileNetV3, EfficientNet, ViT, SWIN), only \ConceptTerm{close-range car} and \ConceptTerm{mid-range car} were detected. This discrepancy may arise because classifiers are trained on ImageNet, which contains object-level images. Moreover, classifiers typically has low-resolution activations in deeper layers. As a result, they may not capture the fine details of more distant objects.

Similar qualitative results were observed for other categories. In the \ConceptTerm{person} category, the \locVect{} of concept samples are grouped based on factors such as distance to the concept and its context (e.g., pose, activity, outfit, background). For example, smaller clusters were identified for people engaged in activities like \ConceptTerm{tennis}, \ConceptTerm{baseball}, \ConceptTerm{skiing}, and \ConceptTerm{surfing}. These findings align closely with the qualitative retrieval results in \cref{fig:retrieval-qualitative,fig:retrieval-qualitative-extra1,fig:retrieval-qualitative-extra2}, as both retrieval and hierarchical clustering rely on $L_2$ distances.

\subsection{Concept Confusion}
\label{sec:applications-concept-confusion}
In the following we first demonstrate how qualitative insights derived from the distribution visualizations can reveal concept confusion and potential causes thereof. Then we support the qualitative findings by additional quantitative analysis using the Concept Separation metric.

\begin{figure*}[tbh]
  \centering
  \includegraphics[width=\linewidth]{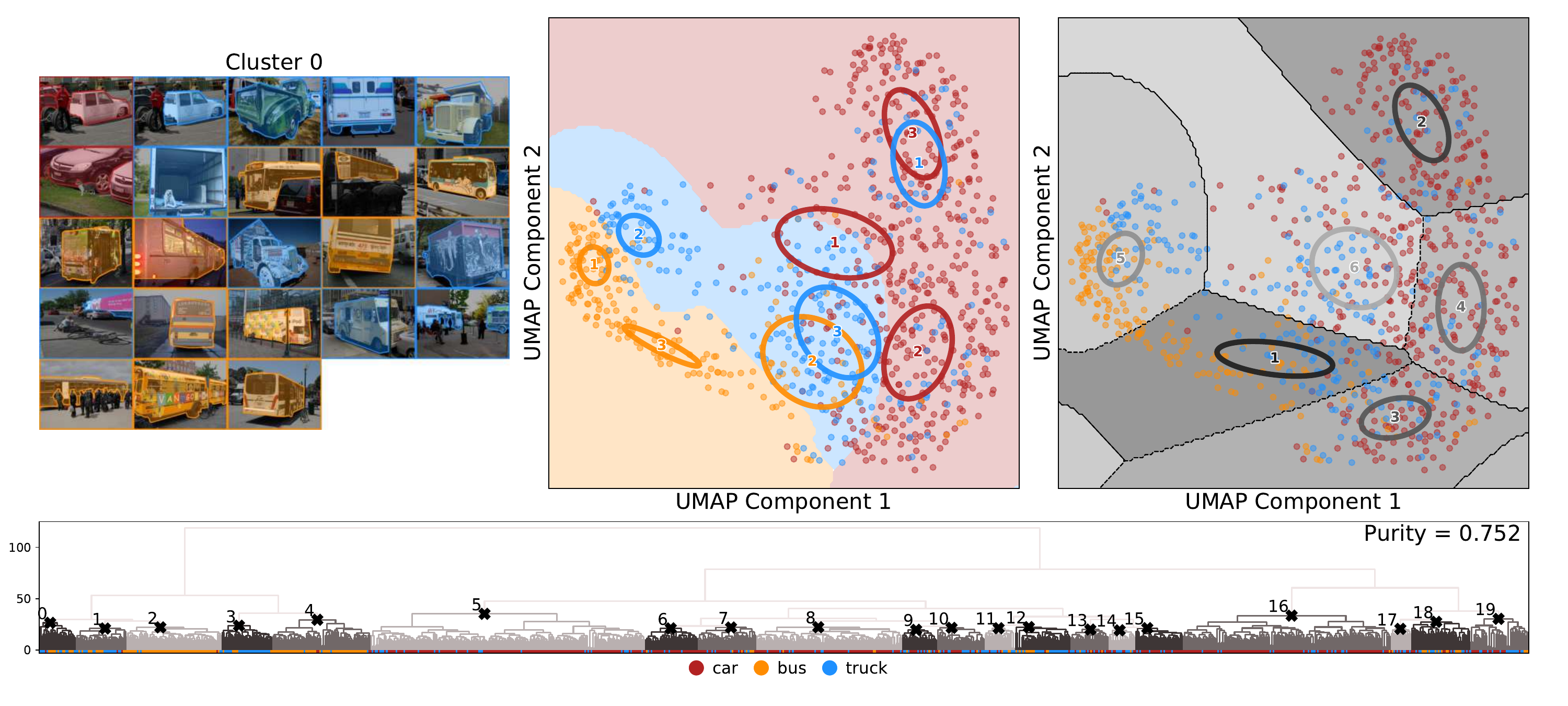}
  \caption{MS COCO: Concept confusion detected with \ConceptTerm{car}, \ConceptTerm{truck}, and \ConceptTerm{bus} \locVects{} in \texttt{model.encoder.layers.1} of DETR: content of a selected cluster of dendrogram (\emph{top-left}), GMMs fitted for 2D UMAP-reduced \locVect{} samples with regard to their labels (\emph{top-middle}), GMMs fitted for 2D UMAP-reduced \locVect{} samples regardless of their labels (\emph{top-right}), and \locVects{} dendrogram with clusters identified with \cref{alg:adaptive-clustering} (\emph{bottom}). Find similar visualizations for further models in \cref{fig:confusion-efficientnet,fig:confusion-swin}}
  \label{fig:confusion-detr}
\end{figure*}

\subsubsection{Qualitative Results}
\locVects{} provide a way to explain the semantic similarities (and low separation) of categories learned by a network by analyzing sample distribution densities. This can be achieved using \locVect{} generalization methods (\cref{sec:method-generalization}) and their corresponding visualizations (\cref{sec:experiments-generalization}).

Based on the findings in \cref{sec:applications-concept-separation}, we analyze the causes of low concept separation and identify concept confusion using the example of \ConceptTerm{car}, \ConceptTerm{truck}, and \ConceptTerm{bus}. \cref{fig:confusion-detr,fig:confusion-efficientnet,fig:confusion-swin} illustrate the confusion among these concepts in layer \texttt{model.encoder.layers.1} of DETR, layer \texttt{features.7.0} of EfficientNet, and layer \texttt{features.7} of SWIN.

The visualizations indicate category confusion in the selected networks, as evidenced by 
(a) the proximity or intersection of GMM components from different categories, and (b) samples from different categories being assigned to the same GMM components and tree clusters. 
Having a look at the contents of the identified sub-clusters reveals that each cluster groups visually similar samples (cf.~exemplary cluster illustrations in \cref{fig:confusion-detr,fig:confusion-efficientnet,fig:confusion-swin}). Through this analysis, we uncovered two reasons for the concept confusion:
\begin{itemize}
    \item \textbf{Labeling errors}: Some identical regions of the input space were assigned to segmentations from different categories, highlighting inconsistencies in the data.
    \item \textbf{Too large concept diversity}: The \ConceptTerm{truck} concept encompasses a wide range of vehicles, including \ConceptTerm{van}, \ConceptTerm{pickup}, and \ConceptTerm{semi-trailer} instances. Clusters distinguishing \ConceptTerm{car} and \ConceptTerm{bus} are better separated but contain \ConceptTerm{truck} infusions. 
\end{itemize}
Such confusion within the feature space might lead to detection errors. \Result{Insights derived from these results can inform data improvements, such as finding individual labeling errors and necessary category refinement} (e.g., separating \ConceptTerm{truck} into \ConceptTerm{van}, \ConceptTerm{pickup}, and \ConceptTerm{semi-trailer}).

\subsubsection{Quantitative Results: Concept Separation}
\label{sec:applications-concept-separation}

\begin{figure*}
  \centering
  \includegraphics[width=\linewidth]{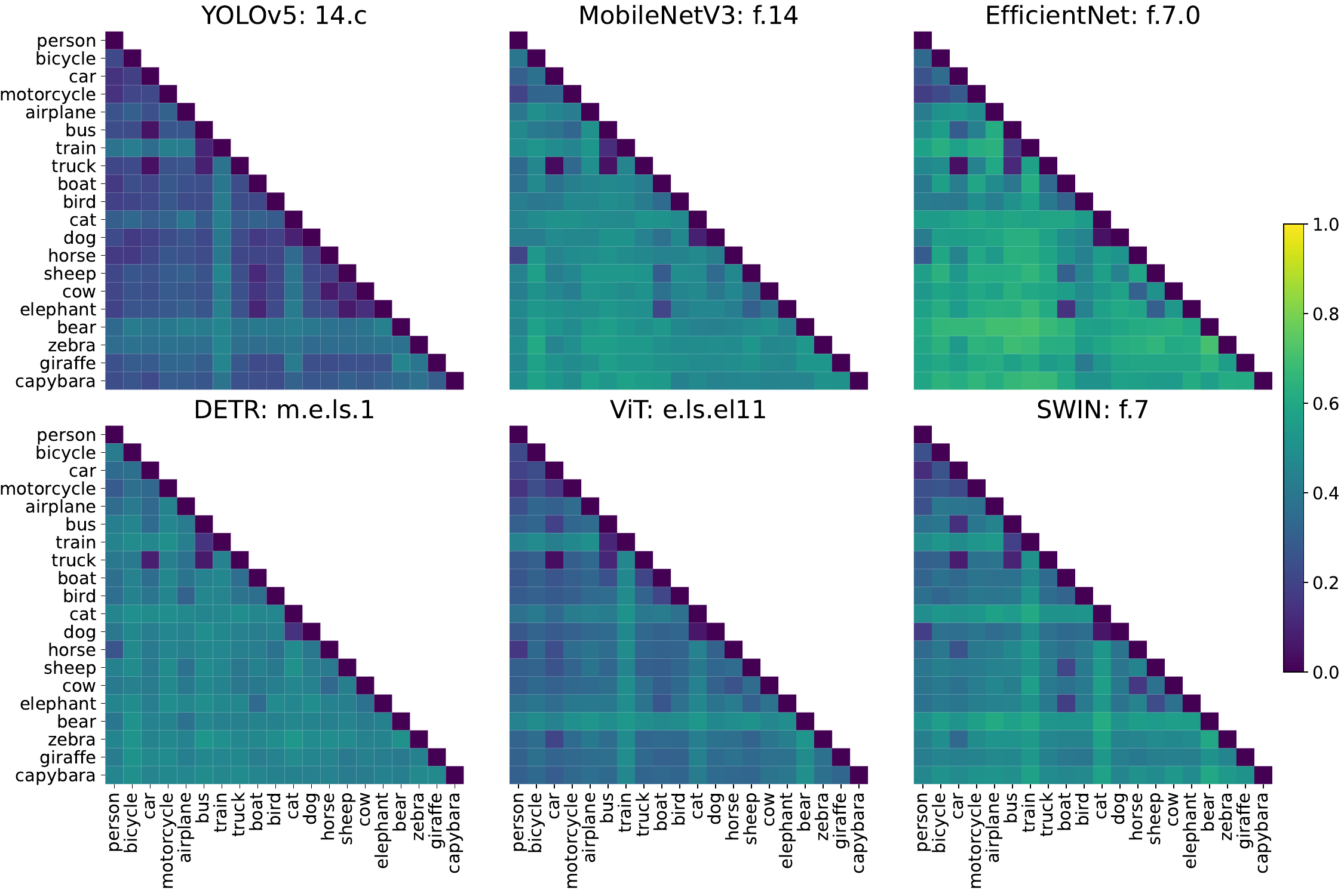}
  \caption{MS COCO \& Capybara Dataset: Pairwise Concept Separation (one-vs-one) results in \textbf{bold layers} of \cref{tab:layers}. See \cref{tab:separation} for Absolute Concept Separation Results.}
  \label{fig:separation}
\end{figure*}

Using \locVects{}, we can evaluate (1) how effectively the feature space of tested DNNs distinguishes between concepts, i.e., how well they are separated, and (2) identify concepts with low separation, indicating potential overlaps.

\cref{fig:separation,tab:separation} present the results of pairwise (one-vs-one) and absolute (one-vs-all) concept separation (see \cref{sec:method-applications}). \Result{Across all tested networks, we observe several recurring cases of low separation, which may signify biases in data}: (1) \ConceptTerm{car}, \ConceptTerm{truck}, and \ConceptTerm{bus}, (2) \ConceptTerm{bus} and \ConceptTerm{train}, and (3) \ConceptTerm{cat} and \ConceptTerm{dog}. From a human perspective, these categories are semantically similar and often appear in similar contexts within the tested data.

Additionally, some networks exhibit other regions of low separation due to semantic similarity (e.g., \ConceptTerm{sheep} and \ConceptTerm{horse} in EfficientNet and SWIN), contextual similarity (e.g., watery environments: \ConceptTerm{boat} and \ConceptTerm{elephant} in EfficientNet, SWIN, and MobileNet), or frequent co-occurrence of concepts (e.g., \ConceptTerm{person} and \ConceptTerm{horse} in EfficientNet, MobileNet, and DETR). These \Result{specific instances of low separation are not consistent across all tested networks, reflecting varying behaviors of their feature spaces}.

\subsection{Concept Outliers Detection}
\label{sec:applications-outliers}


\begin{figure*}[tbh]
  \centering
  \includegraphics[width=\linewidth]{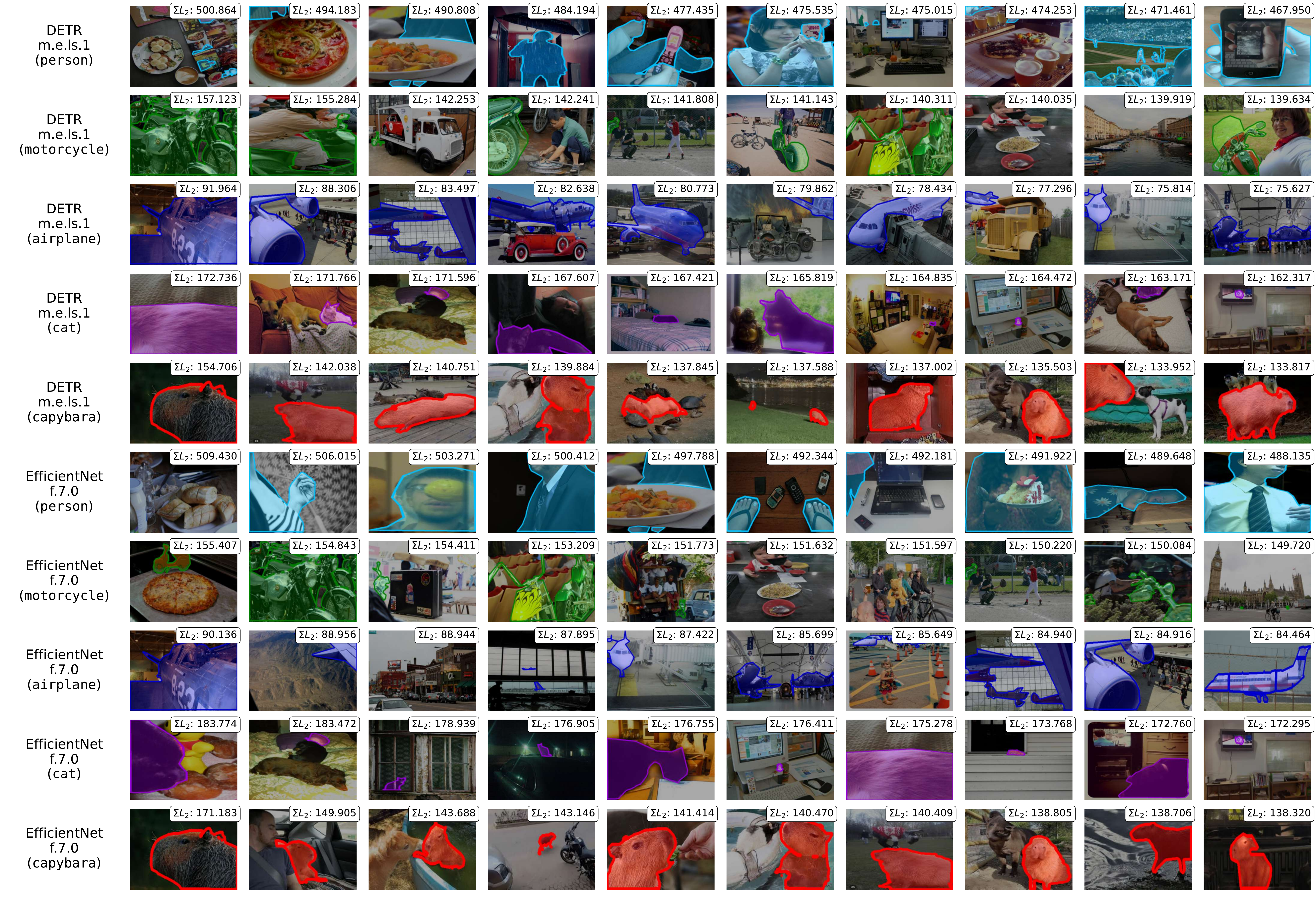}
  \caption{MS COCO \& Capybara Dataset: Top 10 outliers according to $\Sigma L_2$-distance (\emph{columns}) for one layer of best model DETR (\emph{top}) and worst model EfficientNet (\emph{bottom}) and for different concepts (\emph{rows}). Unique concepts are color-coded.
  Find similar visualizations for further models in \cref{fig:outliers-plot-all-extra1,fig:outliers-plot-all-extra2}.}
  \label{fig:outliers-plot-all}
\end{figure*}

Concept-level outliers refer to samples significantly differing from other concepts within the feature space, indicated by distinct \locVect{} values.
We detect these outliers based on distances of concept-related \locVects:
%
%
For each sample, we estimate the cumulative $L_2$-distances ($\Sigma L_2$) from the sample's \locVect{} $v_i \in V_\ConceptClass$ to the \locVects{} of other samples $v' \in V_\ConceptClass\setminus\{v\}$ within the same category $\ConceptClass$ (see \cref{def:outliers-sigma-l2}).
Samples with the highest cumulative distances are outliers, as their concept representations are the most distant from the others. 

We show top outliers across all networks for \ConceptTerm{person}, \ConceptTerm{motorcycle}, \ConceptTerm{airplane}, \ConceptTerm{cat}, and \ConceptTerm{capybara} categories in \cref{fig:outliers-plot-all} (also \cref{fig:outliers-plot-all-extra1,fig:outliers-plot-all-extra2}). Qualitatively, these outliers include samples with 
\begin{enumerate}[label=(\arabic*)]
    \item \textbf{broken target segmentations}, e.g., a capybara sample with only contour segmentation and no filling; 
    \item samples \textbf{missing crucial category features} such as very close-up concepts containing only a part of the concept, e.g., an image showing only the hand of a person;
    \item samples with \textbf{complex or novel contexts}, e.g., images showing only capybara heads with other animals or people in the background or blurred and distorted images of cats. In particular, contexts containing concepts of similar categories (e.g., cat and dog on a sofa) seem to require special features for concept-vs-background discrimination.
\end{enumerate}
This last aspect, in particular, can help to identify hard cases where the DNN might struggle to differentiate concepts of interest.
Such findings could \Result{help to determine measures for label fixing and concept-specific data augmentation}.

\subsection{Identification of semantically shortcutted layers}
\label{sec:applications-shortcut-detection}

As a last application, we showcase how quantitative comparison of concept separation across feature spaces can identify semantically shortcutted layers: Relatively \Result{low mAP results coincide with layers known to contribute less to the model's performance}, possibly indicating measures for architectural improvement. 

Concretely, from \cref{fig:retrieval-mAP-plot}, we observe that in complex branching architectures like DETR and YOLOv5, deeper layers ($l_{\text{last}}$) do not always outperform intermediate layers ($l_{\text{deep}}$). For instance, in DETR, the second transformer encoder captures object-level concepts more effectively than the sixth and final encoder. This phenomenon could be due to architectural shortcuts and connections allowing deeper layers to bypass certain computations, preserving their capacity to learn more specific tasks rather than broader concept representations.

This is supported by results from the original DETR paper by \citep{carion2020detr}. There, the authors tested DETR architectures with different numbers of model blocks to evaluate their effect on performance. This thorough testing revealed that adding more transformer encoder-decoder pairs had diminishing returns on model performance (see Tab.\,2 and Fig.\,4 in \cite{carion2020detr}), with optimal performance reached after 2-3 blocks. In our case, \texttt{m.e.ls.1} corresponds to the second encoder block and \texttt{m.e.ls.5} to the sixth. Hence, \Result{our decreasing mAP results with greater depth are consistent with the decreasing contribution to performance}.
We thus suggest that this semantic analysis of layers might serve as a valuable indication of the relevance of layers. In particular, our post-hoc approach using qualitative \locVects{} retrieval results (\cref{sec:applications-retrieval}) does not need extensive model retraining.
\Result{While this should only be taken as an initial indication, this poses a valuable direction for future research.}

\shortparagraph{Interpretation}
Given DETR's architecture, where each encoder is linked to a corresponding decoder at the same level (\cref{fig:network-architectures}), we assume that deeper encoders primarily target specific detections and features rather than contributing to general feature encoding. These deep encoders are often \enquote{shortcutted} by connections from earlier layers, allowing shallower encoder blocks like \texttt{m.e.ls.1} to handle the majority of feature encoding. As a result, these shallower layers tend to perform best in retrieval tasks (\cref{fig:retrieval-mAP-plot}) from the perspective of \Vect.

\section{Discussion}
\label{sec:discussion}

In this section, we discuss limitations (\cref{sec:discussion-limitations}) and potential future research directions (\cref{sec:discussion-future}).

\subsection{Limitations}
\label{sec:discussion-limitations}
We collect caveats about our method, mostly inherited from its XAI and data-driven nature, that users should be aware of.

\shortparagraph{Dataset scope}
While our experiments leverage the MS COCO, PASCAL VOC, and Capybara datasets (see ~\cref{sec:setup-data}), which provide a mix of general-purpose indoor and outdoor scenes and concepts, domain-specific, and out-of-distribution scenarios, our study is limited to these datasets. Future work should extend this investigation to broader datasets covering diverse domains and data modalities, such as medical imaging, satellite imagery, or fine-grained visual categorization. Exploring these additional datasets would provide a further understanding of the generalizability and robustness of our method across varied tasks and data types.

\shortparagraph{Data availability}
The proposed framework inherits the challenges associated with other post-hoc supervised concept-based approaches~\cite{kim2018interpretability,fong2018net2vec,bau2017network,crabbe2022concept}: \locVects{} require representative data with concept segmentation labels, a challenge that automatic labeling techniques could address. However, this solution may compromise the explanation quality.

\shortparagraph{Online usage}
It should be noted that a single \locVect{} is not expected to generalize well to samples other than the one it was optimized on. 
Application to unseen samples, i.e., online usage to predict a concept segmentation, only makes sense after averaging groups of \locVects{} (\cref{eq:generalization-centroid}). The respectively obtained \subglobVects{} and \globVects{}, however, attain generalization results competitive to other purely global concept embedding methods like Net2Vec (cf.\,\cref{fig:gcpv-vs-baselines}).

\shortparagraph{Quantification of explanations}
The \locVect{} method generates human-interpretable visual explanations, but, as is common in post-hoc XAI, quantitatively assessing the accuracy of these explanations remains challenging~\cite{longo2024explainable}. Furthermore, the absence of standardized methodologies and the diverse array of approaches employed in XAI make it difficult to conduct an objective and comprehensive comparison with other methods.

\shortparagraph{Verification of explanations}
Additionally, the quality of extracted concepts depends on various initial conditions, such as model architecture, data, and layer selection~\cite{mikriukov2023evaluating}: we observed instances where \locVect{} optimization failed, making comparing such samples unfeasible. To overcome this, we conducted ablation experiments (see~\cref{sec:experiments-ablation}), where we tested the impact of resolution and \locVect{} initialization on the final result. However, a deeper analysis of such cases may be required.

\shortparagraph{Human-machine knowledge alignment}
Unlike unsupervised approaches, supervised methods show \emph{user-defined} information~\cite{fong2018net2vec,kim2018interpretability}, whereas unsupervised methods reveal information \emph{seen by the model}~\cite{ghorbani2019towards,zhang2021invertible,fel2023craft}. It is challenging to quantify the disparity between human and machine perspectives. Therefore, a combined use of supervised methods like ours and unsupervised ones is recommended.

\shortparagraph{Computational expenses}
%
Currently, the \Vect{} method demands more computational resources (see~\cref{sec:method-complexity}) than alternative post-hoc baselines~\cite{fong2018net2vec,bau2017network}. This increased demand arises due to the need for \locVect{} generalization using hierarchical clustering, where these methods have a time complexity of $O(N^2)$ and memory complexity $\Omega(n^2)$~\cite{mullner2013fastcluster}. While these expenses only have to be employed once after model training, we acknowledge these computational bottlenecks as potential areas for future research.

\shortparagraph{Optimization techniques}
%
Finally, we recognize our methodology's potential for enhancing optimization algorithms and objectives. To maintain comparability to Net2Vec, their basic setup was little changed or optimized. Concretely, the use of AdamW~\cite{loshchilov2018decoupled} and the Net2Vec pseudo-BCE objective (see~\cref{eq:opt-loss}) makes this a non-convex optimization problem. An exploration of optimization techniques might further improve the resulting \locVects{} performance, e.g., going for Dice loss \cite{rabold2020expressive} or binary cross-entropy \cite{schwalbe2022enabling}.

\subsection{Future Topics}
\label{sec:discussion-future}

We here collect the most interesting research directions for the next steps and future directions of further improving and using \locVects{}.

\shortparagraph{Simultaneous analysis in multiple network layers}
The semantic information within deep neural networks (DNNs) is distributed across multiple layers, with increasing levels of abstraction as we move deeper into the network~\cite{fong2018net2vec,mikriukov2023quantified,schwalbe2021verification}. Distinctive differences in concept representations often emerge at different layers, especially in models with multiple parallel branches and prediction heads, such as object detectors. In these models, each branch may learn to capture different features. By merging information from different layers, we could gain more precise explanations of the learned representations~\cite{posada2022eclad,fel2023craft}.

In the future, this understanding can be integrated into our framework by either (1) concatenating \locVects{} $v_i$ from $i$ different layers $L_i$, with their respective channel dimensions $C_i$, resulting in a combined multi-layer \Vect{} (\multiLocVect{}) of dimensions $(\sum_i C_i) \times 1 \times 1$; or (2) optimizing \multiLocVects{} by stacking activations ($B \times (\sum_i C_i) \times H \times W$) from multiple layers to achieve richer feature representation.

\shortparagraph{Context-agnostic local concept embeddings}
In this work, \Vects{} capture both concept (foreground) and context (background) information, i.e., they are context-dependent: a \emph{specific} concept instance is discriminated against a \emph{specific} background. This differs from global methods like Net2Vec~\cite{fong2018net2vec}, which reduce context influence by averaging it across the entire dataset during optimization: \emph{any} concept instance is discriminated against \emph{any} background.

An intermediate way would be to train a local but context-agnostic (background-irrelevant) local concept embedding: A \emph{specific} concept instance is discriminated against \emph{any} background. To do so, one could optimize the respective representing vector similarly to \locVect{}, but placing the concept patch on different randomly selected backgrounds. This would effectively train a single vector across $N$ images with the same concept patch but different background contexts. To further enhance the robustness of the embedding, the concept patch can also be translated across different positions in each sample.

This approach may result in the loss of contextual information shown to be interesting for, e.g., image retrieval (\cref{sec:applications-retrieval}), and a potential decrease in performance for concept-level information retrieval (cf.~discrepancy between Net2Vec and \locVects{} performance in \cref{sec:experiments-baselines}). Instead, context-agnostic embeddings, due to reduced background influence, may offer other advantages, such as (1) improved or different identification of sub-concepts and (2) better concept generalization.

\shortparagraph{Correlation between concept outliers and network predictions}
\locVect{} enables the detection of outliers at the concept level, providing valuable insights into the semantic organization of the model's feature space. A promising direction for future work would be to extend this analysis to explore correlations between the success or failure of concept identification (or localization) and model predictions: Do concept-level and model-level outliers coincide? By conducting a detailed numerical analysis, we could uncover failure patterns and better understand how concept identification correlates or impacts the overall model performance. This could, in turn, help identify the model’s limitations and biases, providing a pathway to refining the training process and improving the model's robustness, particularly in handling ambiguous or challenging scenarios.

\shortparagraph{Alternative data types}
Expanding the exploration of \locVect{} to other data types, such as temporal (e.g., video and optical flow) and volumetric data (e.g., LIDAR) as relevant to domains like automated driving, is a possible future direction \cite{our_excv_paper}. These data types introduce new dimensions of concept representation, capturing temporal or spatial dynamics that static images cannot. For instance, temporal data can reveal motion-based concepts, such as walking or driving, that unfold over time, while volumetric data may contain three-dimensional structural patterns critical for understanding physical space. Exploring such data will help generalize the framework to identify more complex concepts (and their sub-concepts), enabling applications like autonomous driving, video analysis, and robotics.

\shortparagraph{Global constraints in local concept optimization}
Incorporating global concept constraints into the optimization process of local concepts would help ensure that \locVects{} aligns more closely with the global concept estimate. 
One idea would be to add an extra component $\mathcal{L}_{\text{Loc-Glob}}$ to the total loss function (\cref{eq:opt-loss}) that estimates the divergence between \locVect{} and global concept vectors (e.g., Net2Vec~\cite{fong2018net2vec}):
\begin{align}
    \mathcal{L}_{\text{Loc-Glob}} = \gamma \| v - v_{\text{Glob}} \|_2^2
    \label{eq:optimization-local-global-term}
\end{align}
where $v_{\text{Glob}}$ is the global vector (e.g., Net2Vec) of the relevant concept, and $\alpha$ is a hyperparameter controlling the magnitude of introduced constraint.

Alternatively, initializing \locVect{} with pre-trained global concept vector values (e.g., Net2Vec vectors) before optimization may provide a better starting point and potentially speed up convergence.

\shortparagraph{\locVect{} generalization techniques and their alignment}
In future work, exploring alternative \locVect{} generalization techniques presents an intriguing direction. It could include methods like GMM without UMAP preprocessing or techniques based on matrix factorization for computational efficiency.

While current approaches like UMAP-GMM offer clear visualizations and intuitive understanding for humans, they rely on dimensionality reduction, which distorts information. On the other hand, hierarchical clustering provides a comprehensive, higher-dimensional perspective but is harder to interpret visually and computationally costly. Evaluating the correlation between these techniques, especially under different conditions, can help reveal when their results align and diverge. 



\shortparagraph{More approaches for concept outlier detection}
In this work, we employed distance-based outlier detection (see \cref{sec:applications-outliers}). However, since \locVects{} represent the distribution and density of concepts, density-based methods can also be applied to detect outliers. For example, outliers can be identified through dimensionality reduction with UMAP and GMM fitting (see \cref{sec:method-generalization}). In this context, outliers can be detected by either (1) a low likelihood of Gaussian component membership or (2) by identifying samples associated with a Gaussian component that has a low importance (weight) score.

\shortparagraph{Instance-specific \locVects}
In this work, we focused on image-level segmentations of concepts, without distinguishing between separate instances of the same concept within an image. However, future work could extend \locVects{} to represent individual concept instances, allowing for even more precise representations. In particular, this would eliminate any subconcept averaging since one probing image would only contain a single instance. This refinement could enable a more detailed exploration of feature space representations, in particular, even better distinguishing of sub-concepts.
The main difficulty to tackle would be to provide for a given concept instance a concept-free background: Considering only images with a single concept instance would pose a selection bias, e.g., ruling out crowded scenes for concept \ConceptTerm{person}; ignoring other concept instances in the optimization might leak the other instances into the background due to receptive field sizes; and simply randomizing the background must be done carefully not to distort the background distribution. An augmentation scheme achieving the latter, which removes to-be-ignored concept instances from an image, for example, inpainting techniques, could be used. Alternatively, this could be combined with background randomization, as discussed above.

\section{Conclusion}
\label{sec:conclusion}

We introduced a novel post-hoc supervised local-to-global method for explaining how information about semantic concepts is distributed in DNN feature spaces. 
Using this, we showed that concept distributions are more complex than typically assumed by existing purely global C-XAI methods, frequently exhibiting subconcepts and concept confusion.

The method is based on local concept embeddings (\locVects{}) for concept segmentation that are obtained by a modification of existing global C-XAI methods. 
As experiments show, these concept embeddings enable accurate and competitive reconstruction of concept segmentations, both locally and globally, using clustering and cluster centroids as a means of generalization.
Furthermore, we demonstrate that \Vects{} are helpful for many relevant applications in DNN debugging, such as discovering sub-concepts, concept category confusion, semantic outliers, and retrieving similar images from a DNN perspective.
Additionally, we introduce an adaptation strategy that extends our method and existing concept segmentation techniques for use with vision transformers (ViTs).

Our results open up many future research directions. Most notable is further investigation of applications of \Vects{}, mainly for architecture debugging. Also, extending the approach with background randomization might be an interesting direction.
We hope our work will serve as a foundation for future research in concept-based explainability methods, complementing existing global methods by unraveling the concept distributions.

\FloatBarrier

\section*{Acknowledgments}  
\label{sec:acknowledgments}

This publication was supported with open access funding by Project DEAL and Anhalt University of Applied Sciences.

We would like to thank Dr. Christian Hellert, Dr. Jae Hee Lee, and Dr. Matthias Rottmann for their thoughtful critiques, valuable feedback, and support. Their insights were instrumental in the development of this work.

\section*{Data Availability Statement}
\label{sec:statements}

This work uses publicly available data, see \cref{sec:setup-data} for links.


\bibliography{main}

\clearpage

\appendix
\crefalias{section}{appendix}
\renewcommand{\appendixsection}[2]{\section{#2}}

\appendixsection{A}{Concept Distribution in Global Vectors}
\label{sec:appendix-b}


Opposite to proposed \locVect{}, global concept methods like Net2Vec are not well-suited for identifying sub-concepts (or outliers): rare instances are averaged out when training data is randomly sampled. Only by using individual images can specific concept instances be captured in isolation.

When random sampling is used, generating a concept vector that accurately represents a sub-concept becomes computationally impractical.
%
%
For instance, in a dataset of 100 concept-related samples, where only 10 belong to a specific sub-concept, the likelihood of randomly sampling all 10 sub-concept samples (with a sampling size of 10\%) without repetition is extremely low. The probability of this event is given by $1 / \binom{100}{10}$, $P(X=10) \approx 5.78 \times 10^{-14}$.
%
Even if we soften the requirement, expecting 7 or more samples to belong to the sub-concept, the probability increases only to about $P(X \geq 7) \approx 8.25 \times 10^{-7}$. 

As the dataset size increases, these probabilities drop even further. Consequently, the chances of deriving a vector corresponding to the sub-concept are extremely slim. 
Moreover, to adequately assess the differences in concept distributions, these global vectors would need to be compared against other randomly sampled global vectors, diminishing the likelihood of obtaining meaningful insights.

\appendixsection{B}{Handling Vision Transformer Models}
\label{sec:method-transformers}

Vision Transformers (ViTs) excel in computer vision tasks, but the way they work differs significantly from Convolutional Neural Networks (CNNs). CNNs use trainable convolutional filters applied to local regions of the input to extract different visual features.
This results in an alignment between neurons and spatial positions in the input. 
E.g. for 2D inputs, the resulting activations usually can be indexed by dimensions of 
$C \times H \times W$, representing channels, height, and width, respectively.
This spatial alignment was relaxed in the more recent ViT architectures in favor of more flexible connections: In CNNs, after training, the weights are fixed, determining how a neuron's value is influenced by the values of spatially proximate neurons from the previous layer(s).
In contrast, ViTs are sequence-to-sequence models that use self-attention within their blocks. This means that the model dynamically predicts the weights for the influence of previous layer neurons. That way, also, (spatially) far-away neurons may be picked, allowing global interactions between all input elements at any layer.
Regarding dimensions, ViTs often include a single convolutional patch projection~\cite{dosovitskiy2021vit} or partition~\cite{liu2021swin} layer (or a convolutional backbone with input projection~\cite{carion2020detr}) that converts a $C \times H \times W$ input into a flattened list of vectors representing (non-overlapping) image patches. That results in dimensions $HW \times C$, where $HW$ represents the number of patches (visual tokens), and $C$ is the patch's dimension length. The transformer blocks process these patch vectors as a sequence of tokens.

\begin{figure}[H]
  \centering
  \includegraphics[width=1.0\linewidth]{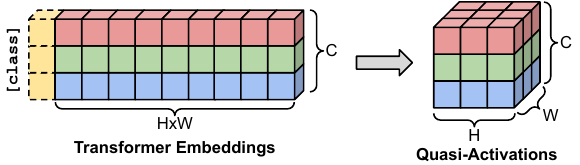}
  \caption{Conversion of sample's transformer embeddings to quasi-activations: Invert the flattening of the spatially arranged patches into a 1D list. Extra tokens (e.g. \texttt{[class]} token), if present, are omitted.
  Hence, \locVects{} can be applied to the quasi-activations extracted from ViTs.
  }
  \label{fig:transformer-activations}
\end{figure}

Inverting the operation that turns the spatially arranged patches into a list of patches, transformer embeddings can be converted into quasi-activations, as illustrated in \cref{fig:transformer-activations}. That way, an approximate spatial alignment of the neurons can be recovered, allowing them to be processed and analyzed similarly to CNN activations. Hence, our \locVect{} can be applied to the quasi-activations extracted for ViTs. It is important to note that some transformer architectures may include additional tokens that do not contain patch-local information (e.g., \texttt{[class]} token in ViT~\cite{dosovitskiy2021vit}), which we omit in our implementation to focus on the core visual tokens.


\begin{figure*}
  \centering
  \includegraphics[width=\linewidth]{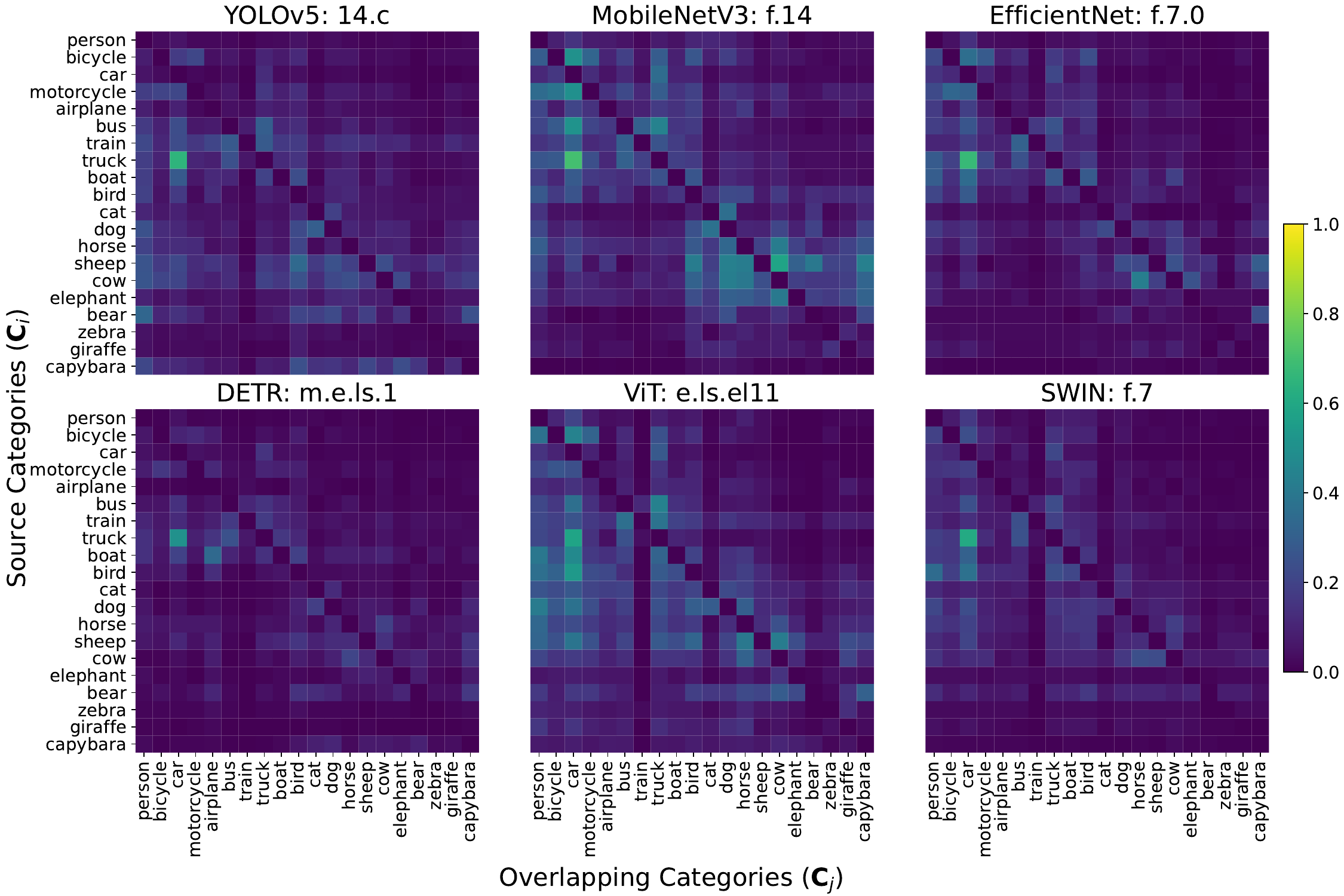}
  \caption{MS COCO \& Capybara Dataset: Concept Overlap results in \textbf{bold layers} of \cref{tab:layers}.}
  \label{fig:overlap}
\end{figure*}

\appendixsection{C}{Estimating Concept Overlap}
\label{sec:appendix-overlap}

An alternative to concept separation (\cref{eq:separation-index-absolute,eq:separation-index-pairwise}), which measures the distances between concepts (each represented by a \locVect{} cluster), is to evaluate the extent of overlap between concepts. Specifically, we aim to quantify the degree to which the overlapping concept $\ConceptClass_j$ intrudes into the region of the source concept $\ConceptClass_i$. Note that this is an asymmetric measure.

For this, we use the Overlap Ratio metric~\cite{borsos2018dealing}. The simplified form we use here\footnote{We set the parameters $k=1$ and $\theta=1$.} is defined as the fraction of \locVects{} in $\ConceptClass_i$ that are closer to any \locVect{} in $\ConceptClass_j$ than to their nearest neighbor in $\ConceptClass_i$:
\begin{align}
    \text{Overlap}(\ConceptClass_i, \ConceptClass_j) = \frac{1}{|\ConceptClass_i|} \sum_{x \in \ConceptClass_i} \delta(x, \ConceptClass_j),
    \label{eq:overlap-ratio}
\end{align}
where $\delta(x, \ConceptClass_j) = 1$ if $\min_{x' \in \ConceptClass_j} \|x - x'\|_2 < \min_{z \in \ConceptClass_i, z \neq x} \|x - z\|_2$, and $\delta(x, \ConceptClass_j) = 0$ otherwise.

A higher Overlap value indicates a more significant intrusion or overlap of $\ConceptClass_j$ into the feature space region of $\ConceptClass_i$.

\paragraph{Results: Concept Overlap}
\cref{fig:overlap} presents estimates of overlaps between tested categories across different models. The results indicate that concepts within the same supercategory (e.g., \ConceptTerm{vehicle}) often exhibit more overlap, as seen in lighter regions (e.g., MobileNet, DETR). The most prominent example is \ConceptTerm{car}, which intrudes into the feature space region of \ConceptTerm{truck} and, to a lesser extent, \ConceptTerm{bus}, and occasionally other categories. This is likely due to \ConceptTerm{car} having the highest number of samples among all categories (except for \ConceptTerm{person}, which remains highly distinct).

\appendixsection{D}{Time and Memory Complexity of \locVect}
\label{sec:appendix-a}\label{sec:appendix-complexity}

During the optimization process, the loss is calculated between the original segmentation mask $\Concept \in \Reals^{h\times w}$ and the concept projection mask $P(v; x) \in \Reals^{h\times w}$, where $P(v; x) = p(v \circ f_{\to L}(x))$.
Here, $v\in\Reals^{C\times 1\times 1}$ represents the \locVect{}, $f_{\to L}(x)\in\Reals^{C\times H\times W}$ denotes the sample activations, and $p(\cdot)$ is a scaling function $p\colon H\times W\to h\times w$. Given that $C\times H\times W$ typically exceeds $h\times w$, we ignore the scaling operation to evaluate time and memory complexities.

\shortparagraph{Time Complexity}
The time complexity for computing the loss components includes $O(C\times H\times W)$ for the projection, and $O(h\times w)$ for BCE and sigmoid operations. The time complexity for optimizing a single \locVect{} across one epoch per sample is $O(C\times H\times W)$.
For a dataset with $n$ samples and $e$ epochs, the total time complexity for optimization is $O(e\times n\times C\times H\times W)$.

Generalization using hierarchical clustering is generally more computationally expensive than generalization using UMAP and GMM fitting. The time complexity for hierarchical clustering is $O((n \times C)^2)$~\cite{mullner2013fastcluster}. At the same time, UMAP has a time complexity of $O(n \times C \times \log n)$ derived from nearest neighbors search, with empirical complexity $O((n \times C)^{1.14})$~\cite{mcinnes2018umap}, and GMM fitting has a time complexity of $O(n \times K \times C'^2)$, where $K$ is the number of GMM components, and $C'$ is reduced \locVect{} dimension ($C' \ll C$). Therefore, for comparison, we consider the worst-case scenario of hierarchical clustering.
Thus, the total time complexity for both optimization and generalization of \globVect{} being $O(e\times n\times C\times H\times W) + O((n\times C)^2)$.

In comparison, the time complexity for Net2Vec's global concept vector is $O(e\times n\times C\times H\times W)$, the same as that of \locVects{} calculation.

\shortparagraph{Memory Complexity}
The memory required for storing projections is $\Omega(n\times C\times H\times W)$, while the segmentation storage requires $\Omega(n\times H\times W)$. The memory needed for storing the concept vectors is $\Omega(n\times C)$.
%

Generalization using hierarchical clustering typically requires more memory than generalization using UMAP and GMM fitting. The memory complexity for hierarchical clustering is $\Omega((n \times C)^2)$~\cite{mullner2013fastcluster}. In contrast, UMAP has a memory complexity of $\Omega(n \times C)$ and GMM fitting has a memory complexity of $\Omega(n \times C \times K)$. Thus, for comparison, we consider the worst-case scenario of hierarchical clustering.

Consequently, the total memory complexity for \locVects{} is $\Omega((n\times C)^2) + \Omega(n\times C\times H\times W) + \Omega(n\times H\times W) + \Omega(n\times C)$.

In contrast, the total memory complexity for Net2Vec is $\Omega(n\times C\times H\times W) + \Omega(n\times H\times W) + \Omega(C)$.

\appendixsection{E}{Additional Results}
\label{sec:appendix-c}

This appendix presents supplementary tables and graphs that provide additional insights and details not included in \cref{sec:experiments} and \cref{sec:applications}. These results further support the findings discussed in the main text and offer a more comprehensive view of the analyzed datasets and methods.

\begin{figure*}
  \centering
  \includegraphics[width=\linewidth]{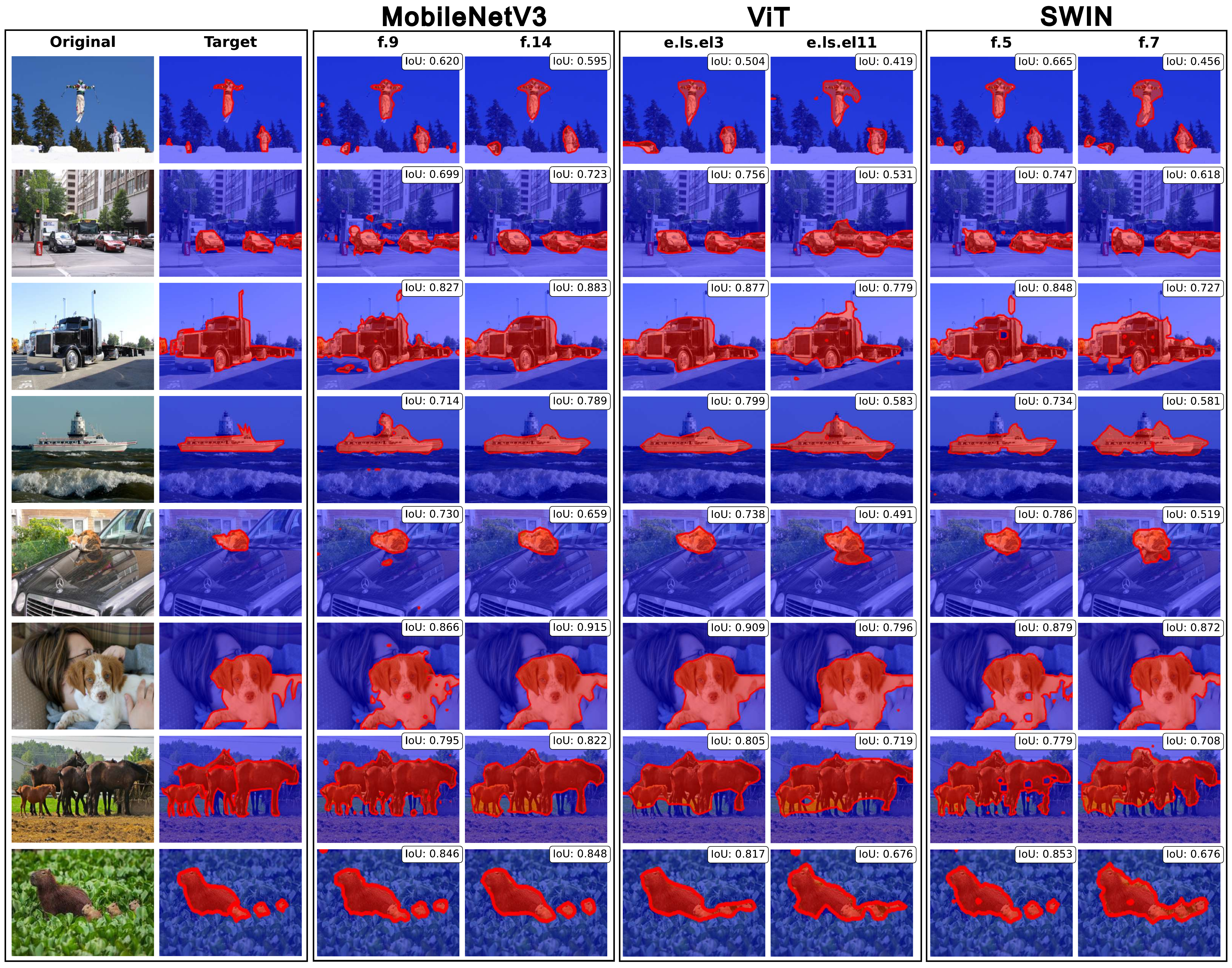}
  \caption{MS COCO \& Capybara Dataset: Exemplary concept segmentation masks predicted by optimized image-local concept models for different trained DNNs and layers (\emph{columns}), and different original input image and target concept segmentation mask (\emph{rows}). Examples chosen randomly.}
  \label{fig:gcpv-optimization-results-2}
\end{figure*}

\begin{table*}
	\centering
	\fontsize{8pt}{10pt}\selectfont
	\setlength{\tabcolsep}{2pt} 
         \caption{%
         MS COCO \& PASCAL VOC \& Capybara Dataset: Averaged concept segmentation performance for all tested samples in IoU of \locVects{} aggregated across different models (\emph{rows}), and layers of increasing depth (\emph{columns}). Models and layers are chosen as described in \cref{tab:layers-ablation}. \locVects{} optimization settings are chosen optimal as found in \cref{sec:experiments-ablation}.
        }
        	\label{tab:optimization-stats-aggregated-coco-voc-capy}
	\begin{tabular}{@{}|c|c|cccccccc|@{}}
            \hline
		\multirow{2}{*}{\textbf{Model}} & \multirow{2}{*}{\textbf{Dataset}} & \multicolumn{8}{c|}{\textbf{IoU}} \\
		                               &                 & $l_1$     & $l_2$     & $l_3$     & $l_4$     & $l_5$     & $l_6$     & $l_7$     & $l_8$     \\
		\hline
		\multirow{3}{*}{YOLOv5}        &   MS COCO   & 0.69±0.23 & 0.67±0.25 & 0.54±0.34 & 0.67±0.27 & 0.67±0.26 & 0.60±0.25 & 0.63±0.24 & 0.57±0.27 \\
									   &  PASCAL VOC & 0.76±0.20 & 0.74±0.22 & 0.62±0.32 & 0.74±0.24 & 0.73±0.25 & 0.65±0.26 & 0.69±0.23 & 0.64±0.25 \\
									   &  Capybara   & 0.86±0.10 & 0.85±0.12 & 0.78±0.19 & 0.87±0.11 & 0.85±0.10 & 0.78±0.14 & 0.81±0.12 & 0.76±0.13 \\
		\hline
		\multirow{3}{*}{MobileNetV3}   &   MS COCO   & 0.63±0.23 & 0.63±0.23 & 0.64±0.23 & 0.63±0.24 & 0.66±0.25 & 0.67±0.25 & 0.61±0.27 & ---       \\
									   &  PASCAL VOC & 0.69±0.21 & 0.70±0.20 & 0.71±0.20 & 0.71±0.21 & 0.73±0.22 & 0.74±0.22 & 0.69±0.24 & ---       \\
									   &  Capybara   & 0.79±0.13 & 0.80±0.12 & 0.82±0.12 & 0.83±0.11 & 0.86±0.11 & 0.86±0.11 & 0.81±0.15 & ---       \\
		\hline
		\multirow{3}{*}{EfficientNet}  &   MS COCO   & 0.60±0.25 & 0.63±0.23 & 0.64±0.22 & 0.65±0.22 & 0.63±0.26 & 0.62±0.26 & 0.61±0.26 & 0.54±0.28 \\
									   &  PASCAL VOC & 0.66±0.22 & 0.70±0.20 & 0.71±0.20 & 0.72±0.20 & 0.70±0.23 & 0.70±0.24 & 0.69±0.24 & 0.62±0.26 \\
									   &  Capybara   & 0.79±0.13 & 0.81±0.11 & 0.83±0.12 & 0.84±0.11 & 0.84±0.13 & 0.83±0.13 & 0.83±0.14 & 0.75±0.17 \\
		\hline
		\multirow{3}{*}{DETR}          &   MS COCO   & 0.75±0.21 & 0.64±0.25 & 0.68±0.23 & 0.68±0.23 & 0.68±0.23 & 0.66±0.23 & 0.66±0.24 & 0.64±0.24 \\
									   &  PASCAL VOC & 0.81±0.19 & 0.71±0.23 & 0.74±0.21 & 0.74±0.21 & 0.74±0.21 & 0.73±0.21 & 0.72±0.21 & 0.71±0.22 \\
									   &  Capybara   & 0.91±0.09 & 0.84±0.12 & 0.86±0.11 & 0.86±0.10 & 0.85±0.11 & 0.84±0.11 & 0.84±0.11 & 0.83±0.11 \\
		\hline
		\multirow{3}{*}{ViT}           &   MS COCO   & 0.65±0.25 & 0.66±0.26 & 0.65±0.27 & 0.63±0.26 & 0.59±0.28 & 0.52±0.29 & ---       & ---       \\
									   &  PASCAL VOC & 0.72±0.23 & 0.73±0.23 & 0.73±0.24 & 0.71±0.23 & 0.68±0.25 & 0.61±0.26 & ---       & ---       \\
									   &  Capybara   & 0.85±0.12 & 0.86±0.11 & 0.86±0.12 & 0.83±0.13 & 0.81±0.13 & 0.75±0.13 & ---       & ---       \\
		\hline
		\multirow{3}{*}{SWIN}          &   MS COCO   & 0.03±0.13 & 0.37±0.30 & 0.31±0.36 & 0.61±0.31 & 0.55±0.28 & ---       & ---       & ---       \\
									   &  PASCAL VOC & 0.06±0.20 & 0.43±0.31 & 0.30±0.38 & 0.67±0.27 & 0.62±0.25 & ---       & ---       & ---       \\
									   &  Capybara   & 0.13±0.30 & 0.59±0.26 & 0.21±0.37 & 0.84±0.09 & 0.73±0.14 & ---       & ---       & ---       \\
		\hline
	\end{tabular}
\end{table*}

\begin{figure*}
  \centering
  \includegraphics[width=\linewidth]{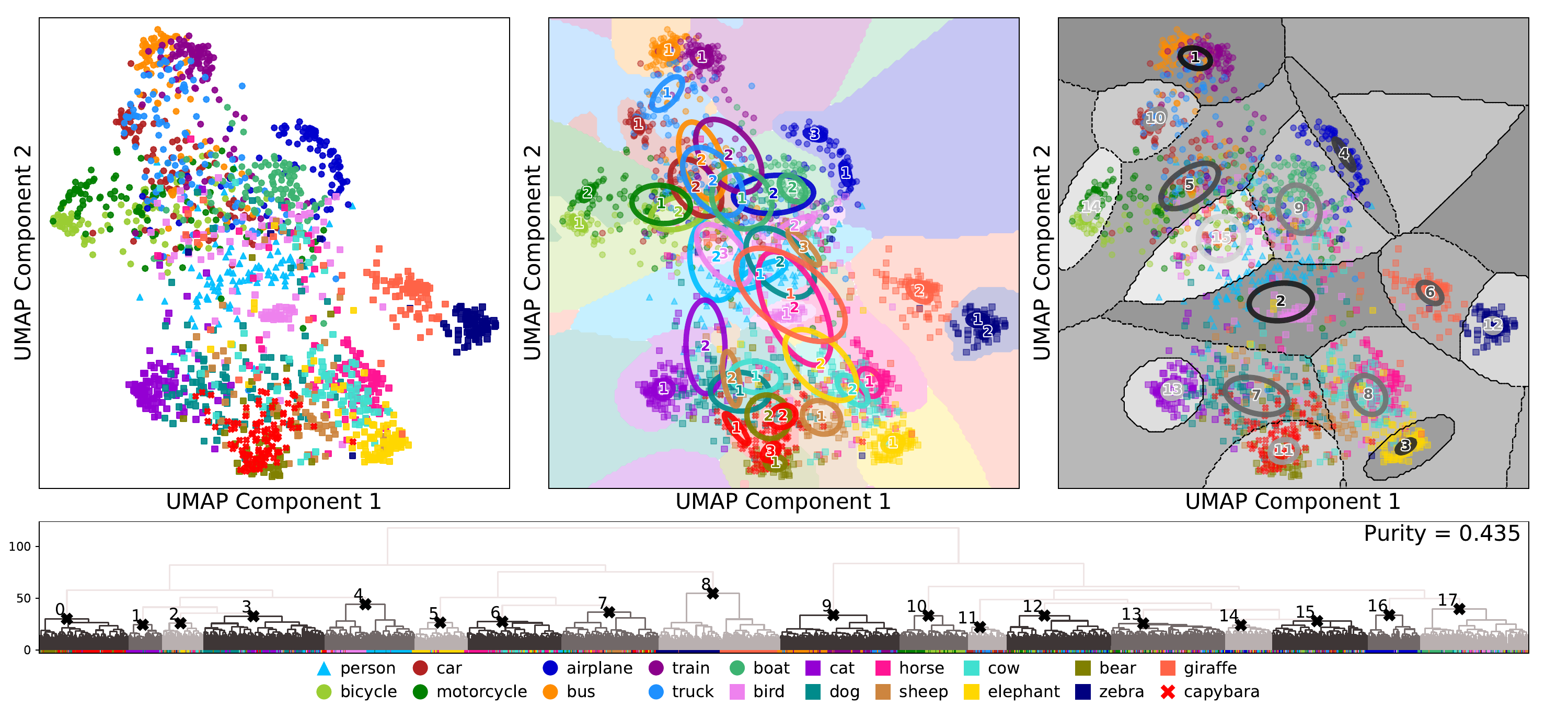}
  \caption{MS COCO \& Capybara Dataset: Generalization of all tested concepts \locVects{} in \texttt{features.7.0} of EfficientNet: 2D UMAP-reduced \locVects{} of every tested category (\emph{top-left}), GMMs fitted for samples with regard to their labels (\emph{top-middle}), GMMs fitted for all samples regardless of their labels (\emph{top-right}), and \locVects{} dendrogram with clusters identified with \cref{alg:adaptive-clustering} (\emph{bottom}).}
  \label{fig:gmm-dendrogram-efficientnet-f7}
\end{figure*}

\begin{figure*}
  \centering
  \includegraphics[width=\linewidth]{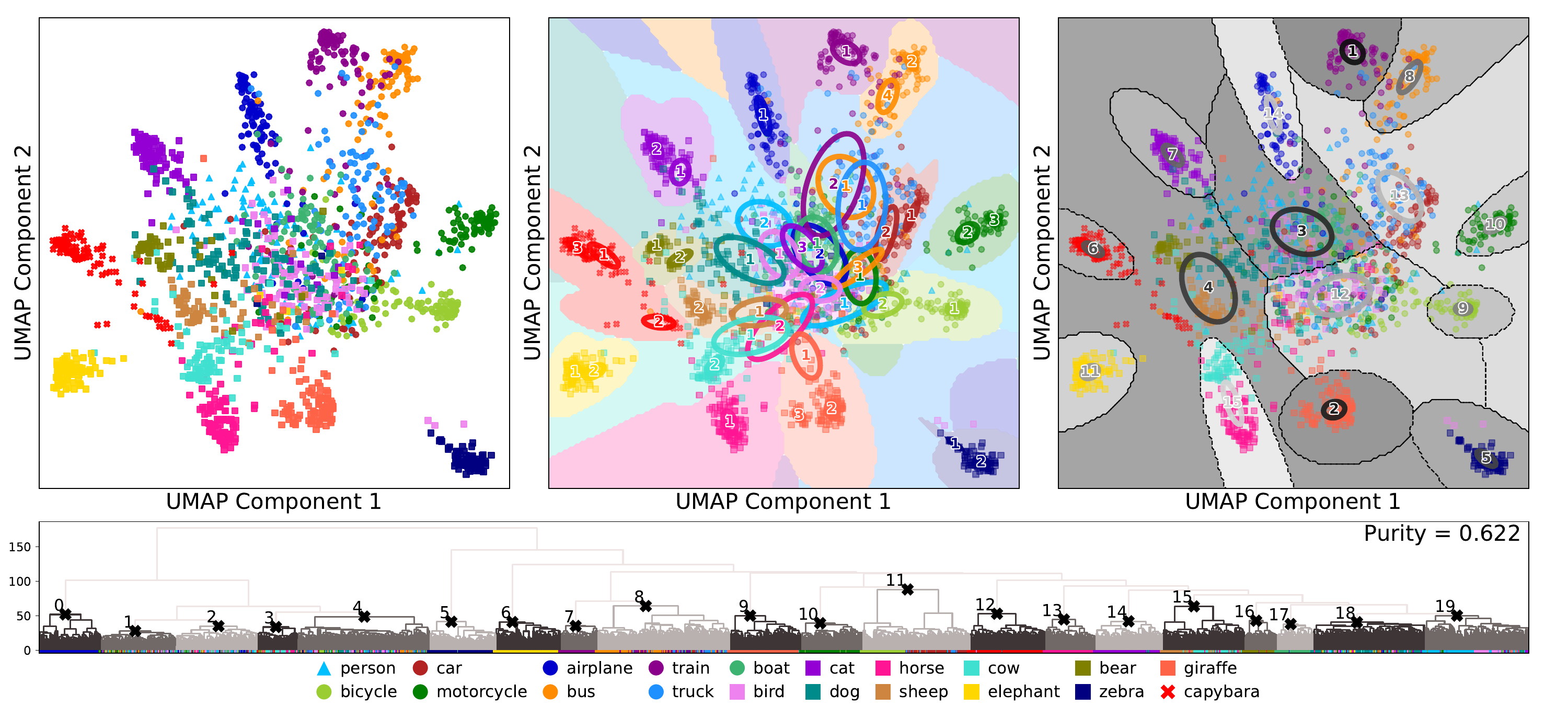}
  \caption{MS COCO \& Capybara Dataset: Generalization of all tested concepts \locVects{} in \texttt{features.7} of SWIN: 2D UMAP-reduced \locVects{} of every tested category (\emph{top-left}), GMMs fitted for samples with regard to their labels (\emph{top-middle}), GMMs fitted for all samples regardless of their labels (\emph{top-right}), and \locVects{} dendrogram with clusters identified with \cref{alg:adaptive-clustering} (\emph{bottom}).}
  \label{fig:gmm-dendrogram-swin-f7}
\end{figure*}

\begin{figure*}[tbh]
  \centering
  \includegraphics[width=\linewidth]{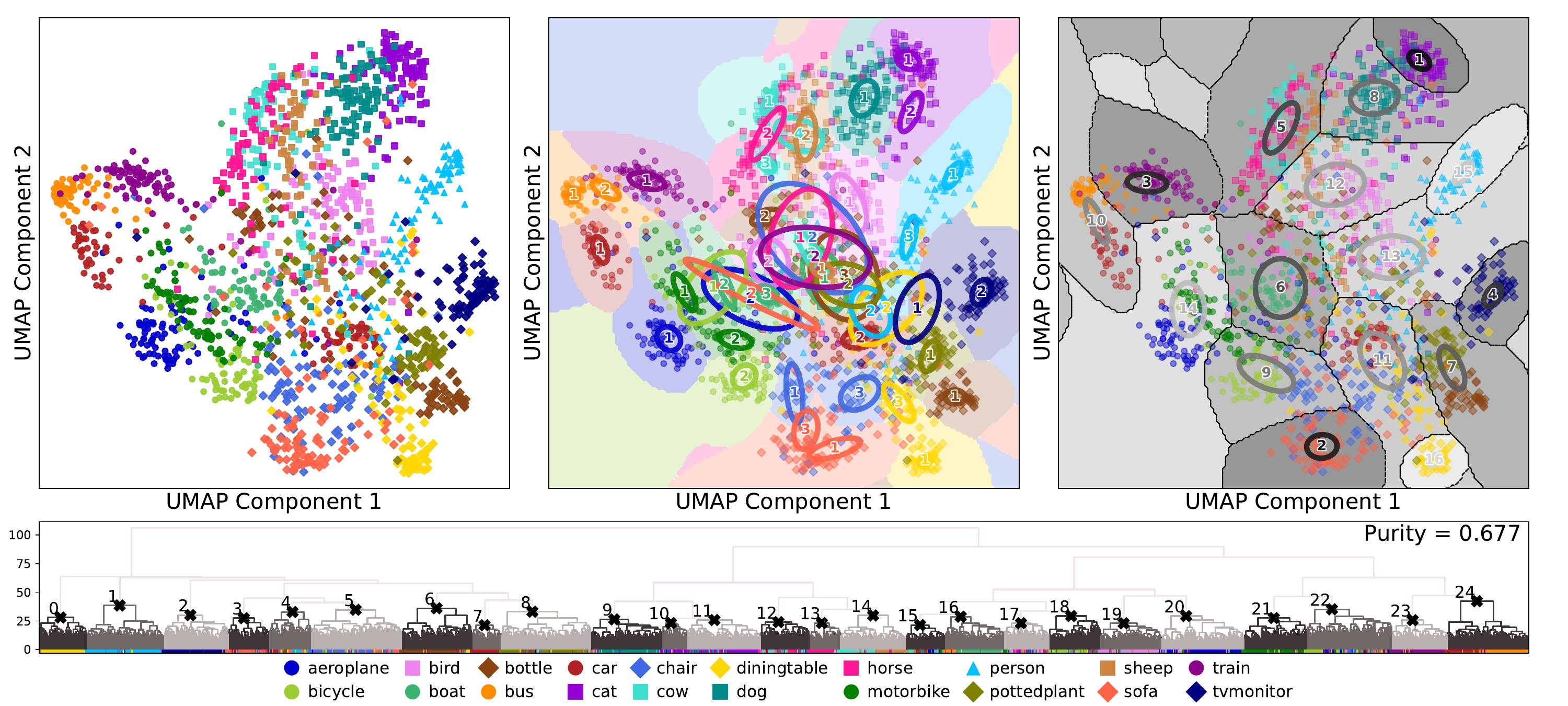}
  \caption{PASCAL VOC: Generalization of all tested concepts \locVects{} in \texttt{model.encoder.layers.1} of DETR: 2D UMAP-reduced \locVects{} of every tested category (\emph{top-left}), GMMs fitted for samples with regard to their labels (\emph{top-middle}), GMMs fitted for all samples regardless of their labels (\emph{top-right}), and \locVects{} dendrogram with clusters identified with \cref{alg:adaptive-clustering} (\emph{bottom}). }
  \label{fig:gmm-dendrogram-detr-e1-voc}
\end{figure*}

\begin{figure*}
  \centering
  \includegraphics[width=\linewidth]{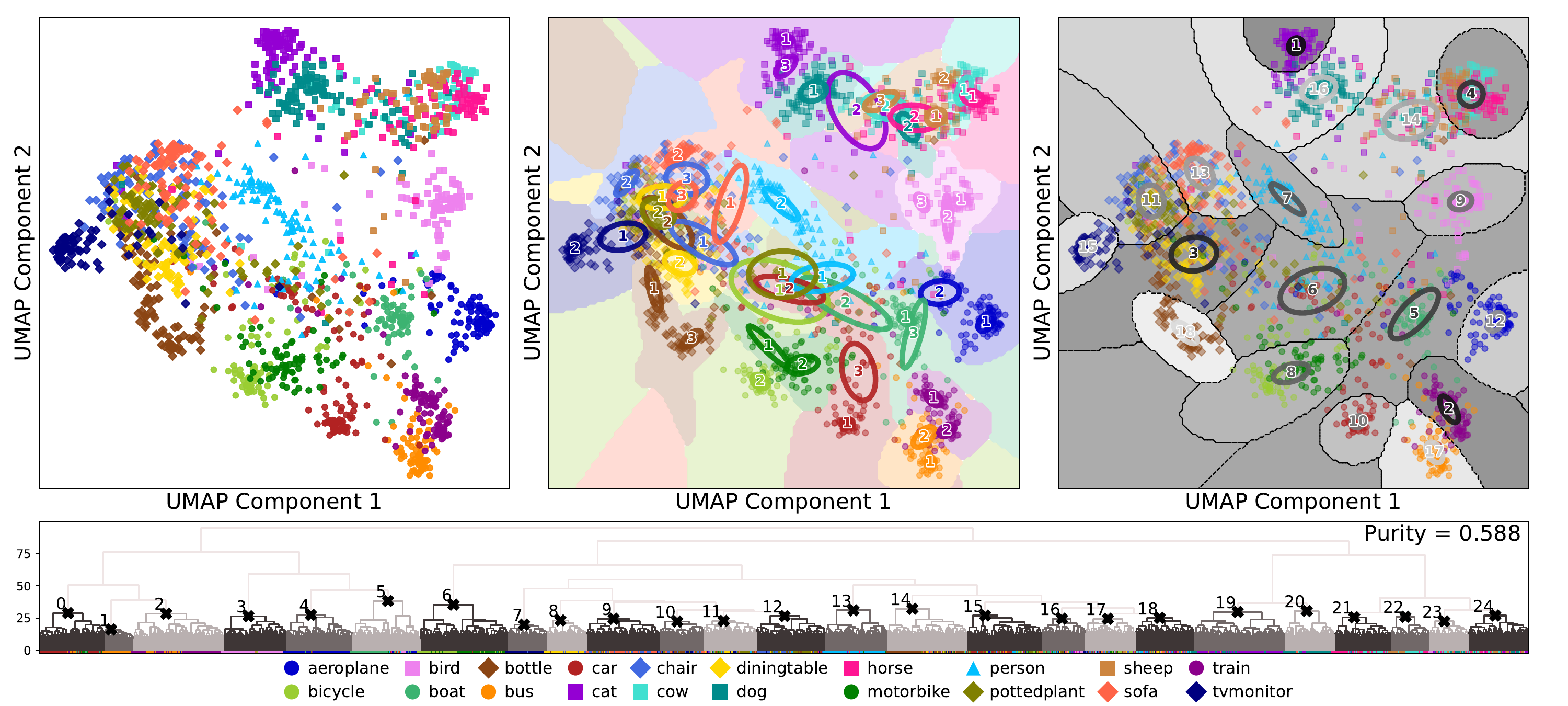}
  \caption{PASCAL VOC: Generalization of all tested concepts \locVects{} in \texttt{features.7.0} of EfficientNet: 2D UMAP-reduced \locVects{} of every tested category (\emph{top-left}), GMMs fitted for samples with regard to their labels (\emph{top-middle}), GMMs fitted for all samples regardless of their labels (\emph{top-right}), and \locVects{} dendrogram with clusters identified with \cref{alg:adaptive-clustering} (\emph{bottom}).}
  \label{fig:gmm-dendrogram-efficientnet-f7-voc}
\end{figure*}

\begin{figure*}
  \centering
  \includegraphics[width=\linewidth]{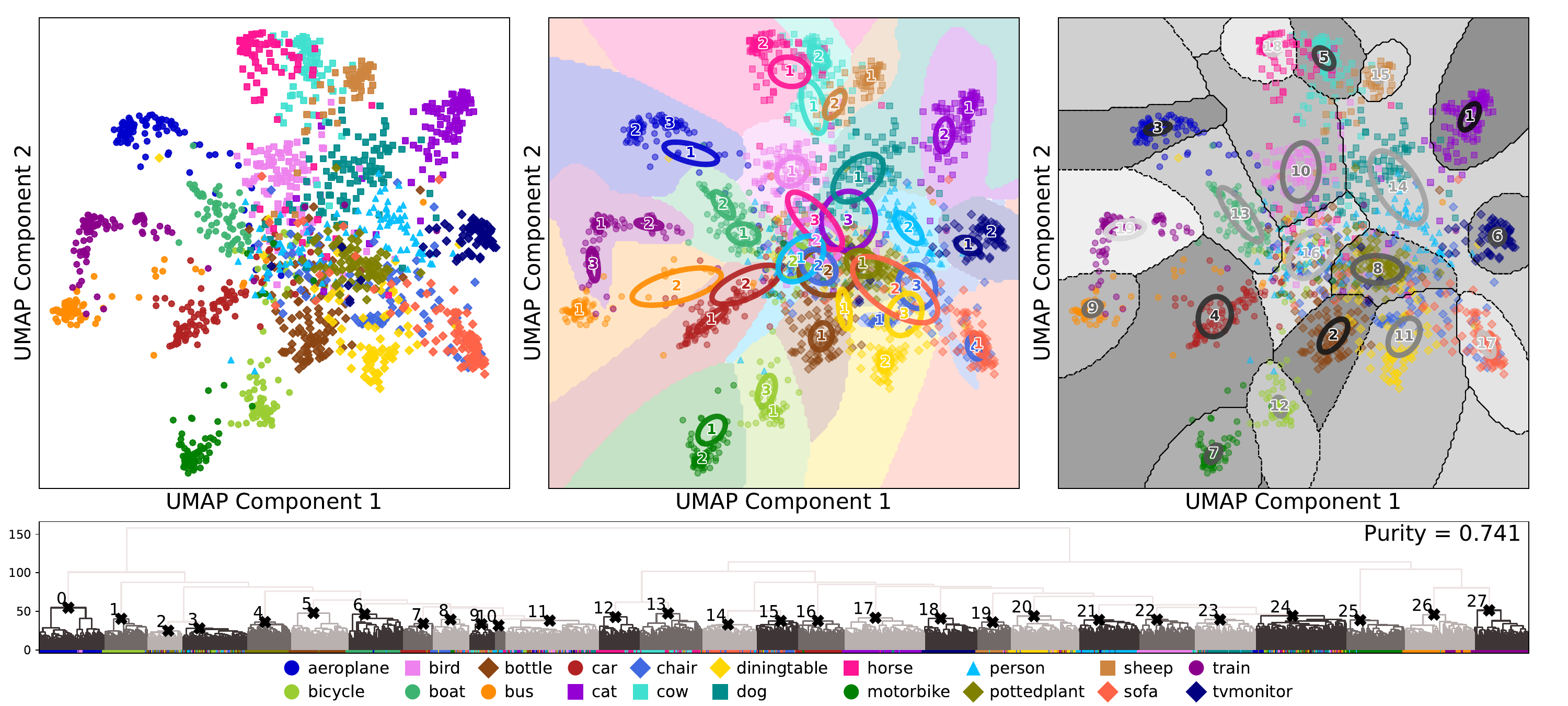}
  \caption{PASCAL VOC: Generalization of all tested concepts \locVects{} in \texttt{features.7} of SWIN: 2D UMAP-reduced \locVects{} of every tested category (\emph{top-left}), GMMs fitted for samples with regard to their labels (\emph{top-middle}), GMMs fitted for all samples regardless of their labels (\emph{top-right}), and \locVects{} dendrogram with clusters identified with \cref{alg:adaptive-clustering} (\emph{bottom}).}
  \label{fig:gmm-dendrogram-swin-f7-voc}
\end{figure*}

\begin{table*}
        \centering
        \fontsize{8pt}{10pt}\selectfont
        \setlength{\tabcolsep}{5pt}
        \caption{MS COCO: Concept segmentation performance comparison between the generalizations of \locVects{} to its (sub-)global variants (best matching \subglobVects{}, \globVects{}, see \cref{eq:generalization-centroid}), and direct global baseline methods (see \cref{sec:setup-baselines}).
        For reference, the average performance of individual \locVects{} on their respective training images is shown, marking the maximally achievable segmentation performance of our (sub-)global variants. Results are averaged across MS COCO supercategories (\ConceptTerm{person}, \ConceptTerm{vehicle}, \ConceptTerm{animal}) and all tested categories (\ConceptTerm{overall}).
        Layers are chosen as described in \cref{tab:layers}.
        Graphical visualizations of the results (\ConceptTerm{overall}) are shown in \cref{fig:gcpv-vs-baselines}.}
        \label{tab:gcpv-vs-baselines-final}
        \begin{tabular}{@{}|c|c|ccc|ccc|ccc|ccc|@{}}
                \hline
                \multirow{2}{*}{\textbf{Model}} & \multirow{2}{*}{\textbf{Method}} & \multicolumn{3}{c|}{\textbf{person (500)}} & \multicolumn{3}{c|}{\textbf{vehicle (1622)}} & \multicolumn{3}{c|}{\textbf{animal (1090)}} & \multicolumn{3}{c|}{\textbf{overall (3212)}} \\
                                              &                   &     $l_{mid}$  &     $l_{deep}$ &     $l_{last}$ &     $l_{mid}$  &     $l_{deep}$ &     $l_{last}$ &     $l_{mid}$  &     $l_{deep}$ &     $l_{last}$ &     $l_{mid}$  &     $l_{deep}$ &     $l_{last}$ \\
                \hline
                      \multirow{6}{*}{YOLOv5} &           Net2Vec &     0.51  &     0.56  &     0.54  &     0.35  &     0.44  &     0.44  &     0.52  &     0.57  &     0.56  &     0.43  &     0.50  &     0.50  \\
                                              &        Net2Vec-16 &     0.42  &     0.41  &     0.35  &     0.25  &     0.31  &     0.24  &     0.44  &     0.44  &     0.34  &     0.34  &     0.37  &     0.29  \\
                                              &        NetDissect &     0.19  &     0.21  &     0.32  &     0.20  &     0.22  &     0.19  &     0.16  &     0.31  &     0.22  &     0.18  &     0.25  &     0.22  \\
                                              &            \locVect{} &     0.71  &     0.64  &     0.64  &     0.64  &     0.59  &     0.60  &     0.74  &     0.68  &     0.68  &     0.69  &     0.63  &     0.63  \\
                                              &           \subglobVect{} &     0.56  &     0.61  &     0.59  &     0.45  &     0.54  &     0.51  &     0.58  &     0.63  &     0.58  &     0.51  &     0.58  &     0.55  \\
                                              &            \globVect{} &     0.37  &     0.30  &     0.35  &     0.24  &     0.32  &     0.31  &     0.42  &     0.46  &     0.43  &     0.32  &     0.36  &     0.36  \\
                \hline
                 \multirow{6}{*}{MobileNetV3} &           Net2Vec &     0.39  &     0.41  &     0.53  &     0.28  &     0.29  &     0.39  &     0.44  &     0.47  &     0.59  &     0.35  &     0.37  &     0.48  \\
                                              &        Net2Vec-16 &     0.30  &     0.32  &     0.39  &     0.22  &     0.21  &     0.26  &     0.33  &     0.37  &     0.40  &     0.27  &     0.28  &     0.32  \\
                                              &        NetDissect &     0.08  &     0.34  &     0.23  &     0.11  &     0.14  &     0.16  &     0.24  &     0.25  &     0.24  &     0.15  &     0.21  &     0.20  \\
                                              &            \locVect{} &     0.65  &     0.66  &     0.68  &     0.57  &     0.57  &     0.62  &     0.70  &     0.71  &     0.72  &     0.63  &     0.64  &     0.67  \\
                                              &           \subglobVect{} &     0.49  &     0.48  &     0.59  &     0.34  &     0.34  &     0.43  &     0.52  &     0.52  &     0.60  &     0.43  &     0.42  &     0.51  \\
                                              &            \globVect{} &     0.31  &     0.33  &     0.50  &     0.16  &     0.19  &     0.32  &     0.38  &     0.40  &     0.55  &     0.26  &     0.28  &     0.42  \\
                \hline
                \multirow{6}{*}{EfficientNet} &           Net2Vec &     0.44  &     0.55  &     0.57  &     0.32  &     0.41  &     0.44  &     0.50  &     0.62  &     0.64  &     0.40  &     0.50  &     0.53  \\
                                              &        Net2Vec-16 &     0.34  &     0.36  &     0.39  &     0.23  &     0.24  &     0.25  &     0.38  &     0.39  &     0.37  &     0.30  &     0.31  &     0.31  \\
                                              &        NetDissect &     0.24  &     0.23  &     0.19  &     0.16  &     0.15  &     0.16  &     0.24  &     0.25  &     0.24  &     0.20  &     0.20  &     0.19  \\
                                              &            \locVect{} &     0.66  &     0.63  &     0.57  &     0.60  &     0.57  &     0.50  &     0.71  &     0.67  &     0.58  &     0.65  &     0.61  &     0.54  \\
                                              &           \subglobVect{} &     0.51  &     0.60  &     0.60  &     0.37  &     0.45  &     0.46  &     0.55  &     0.64  &     0.63  &     0.45  &     0.54  &     0.54  \\
                                              &            \globVect{} &     0.36  &     0.48  &     0.50  &     0.19  &     0.29  &     0.38  &     0.42  &     0.57  &     0.58  &     0.29  &     0.41  &     0.47  \\
                \hline
                        \multirow{6}{*}{DETR} &           Net2Vec &     0.52  &     0.63  &     0.64  &     0.37  &     0.47  &     0.52  &     0.55  &     0.65  &     0.68  &     0.45  &     0.56  &     0.59  \\
                                              &        Net2Vec-16 &     0.27  &     0.36  &     0.28  &     0.15  &     0.25  &     0.22  &     0.27  &     0.43  &     0.36  &     0.21  &     0.33  &     0.28  \\
                                              &        NetDissect &     0.20  &     0.20  &     0.22  &     0.15  &     0.18  &     0.21  &     0.20  &     0.21  &     0.24  &     0.18  &     0.19  &     0.22  \\
                                              &            \locVect{} &     0.76  &     0.70  &     0.66  &     0.72  &     0.64  &     0.59  &     0.81  &     0.73  &     0.71  &     0.75  &     0.68  &     0.64  \\
                                              &           \subglobVect{} &     0.56  &     0.65  &     0.61  &     0.43  &     0.55  &     0.47  &     0.59  &     0.68  &     0.60  &     0.51  &     0.61  &     0.54  \\
                                              &            \globVect{} &     0.26  &     0.45  &     0.32  &     0.12  &     0.37  &     0.21  &     0.35  &     0.58  &     0.44  &     0.22  &     0.45  &     0.31  \\
                \hline
                         \multirow{6}{*}{ViT} &           Net2Vec &     0.43  &     0.56  &     0.54  &     0.35  &     0.45  &     0.43  &     0.52  &     0.63  &     0.62  &     0.42  &     0.53  &     0.52  \\
                                              &        Net2Vec-16 &     0.18  &     0.30  &     0.32  &     0.16  &     0.16  &     0.19  &     0.26  &     0.25  &     0.27  &     0.19  &     0.21  &     0.24  \\
                                              &        NetDissect &     0.24  &     0.24  &     0.26  &     0.17  &     0.17  &     0.19  &     0.27  &     0.24  &     0.28  &     0.22  &     0.20  &     0.23  \\
                                              &            \locVect{} &     0.66  &     0.63  &     0.54  &     0.63  &     0.60  &     0.46  &     0.72  &     0.69  &     0.60  &     0.66  &     0.63  &     0.52  \\
                                              &           \subglobVect{} &     0.49  &     0.58  &     0.51  &     0.36  &     0.42  &     0.36  &     0.56  &     0.59  &     0.51  &     0.45  &     0.50  &     0.43  \\
                                              &            \globVect{} &     0.16  &     0.35  &     0.16  &     0.09  &     0.17  &     0.20  &     0.31  &     0.43  &     0.39  &     0.18  &     0.29  &     0.26  \\
                \hline
                        \multirow{6}{*}{SWIN} &           Net2Vec &     ---   &     0.23  &     0.57  &     ---   &     0.36  &     0.46  &     ---   &     0.33  &     0.66  &     ---   &     0.33  &     0.54  \\
                                              &        Net2Vec-16 &     ---   &     0.19  &     0.27  &     ---   &     0.22  &     0.18  &     ---   &     0.26  &     0.27  &     ---   &     0.23  &     0.23  \\
                                              &        NetDissect &     ---   &     0.21  &     0.21  &     ---   &     0.16  &     0.16  &     ---   &     0.22  &     0.24  &     ---   &     0.19  &     0.20  \\
                                              &            \locVect{} &     ---   &     0.56  &     0.54  &     ---   &     0.56  &     0.50  &     ---   &     0.70  &     0.62  &     ---   &     0.61  &     0.55  \\
                                              &           \subglobVect{} &     ---   &     0.51  &     0.36  &     ---   &     0.40  &     0.41  &     ---   &     0.60  &     0.53  &     ---   &     0.49  &     0.44  \\
                                              &            \globVect{} &     ---   &     0.01  &     0.01  &     ---   &     0.08  &     0.18  &     ---   &     0.28  &     0.32  &     ---   &     0.14  &     0.20  \\
                \hline
        \end{tabular}
\end{table*}


\begin{figure*}
  \centering
  \includegraphics[width=\linewidth]{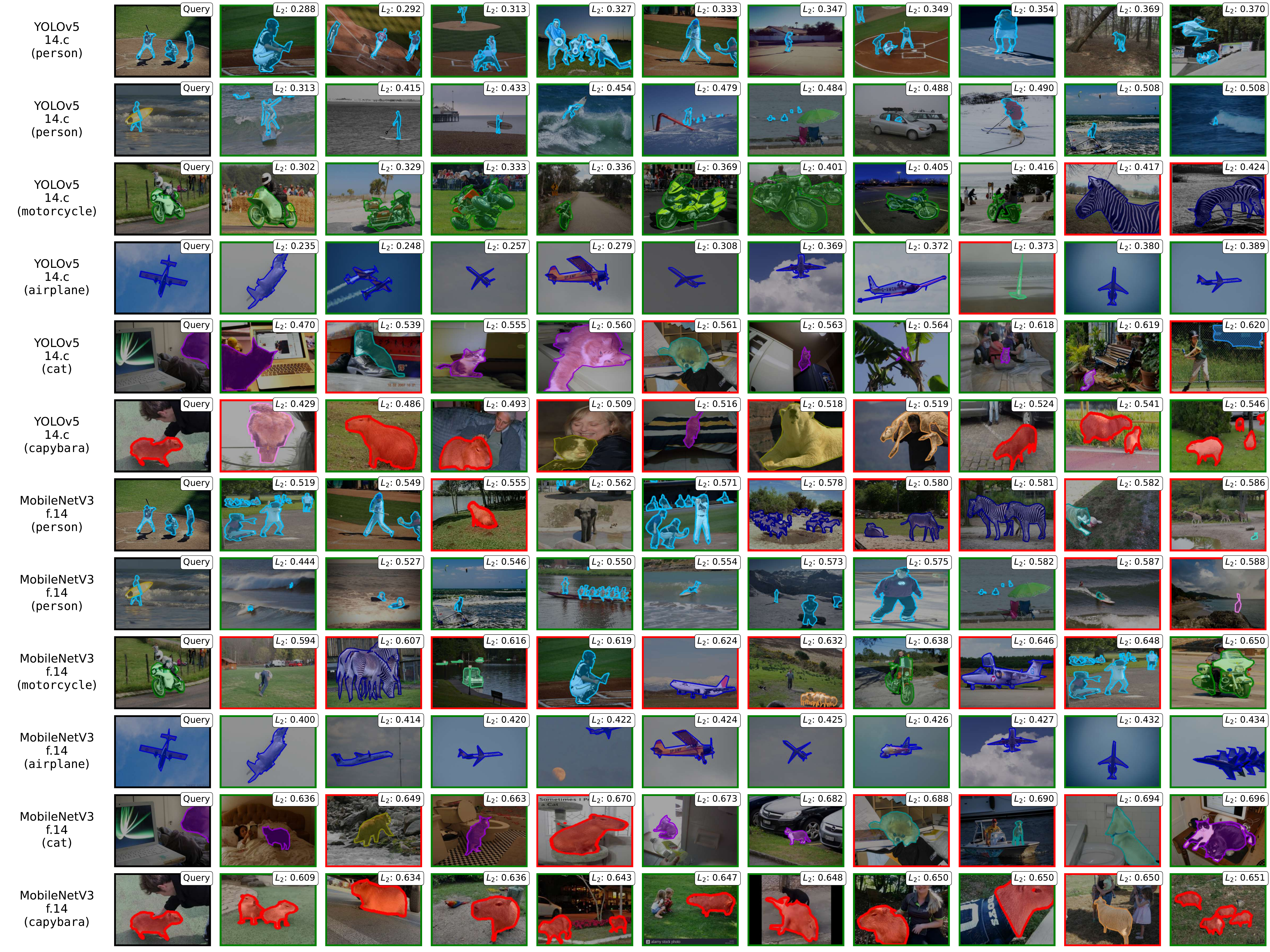}
  \caption{MS COCO \& Capybara Dataset: Top 10 retrieval results according to $L_2$-distance (\emph{columns}) for random \locVect{} queries (\emph{black frame, leftmost column}) of different categories. \emph{Green frame} - relevant sample. \emph{Red frame} - irrelevant sample. Unique concepts are color-coded.}
  \label{fig:retrieval-qualitative-extra1}
\end{figure*}

\begin{figure*}
  \centering
  \includegraphics[width=\linewidth]{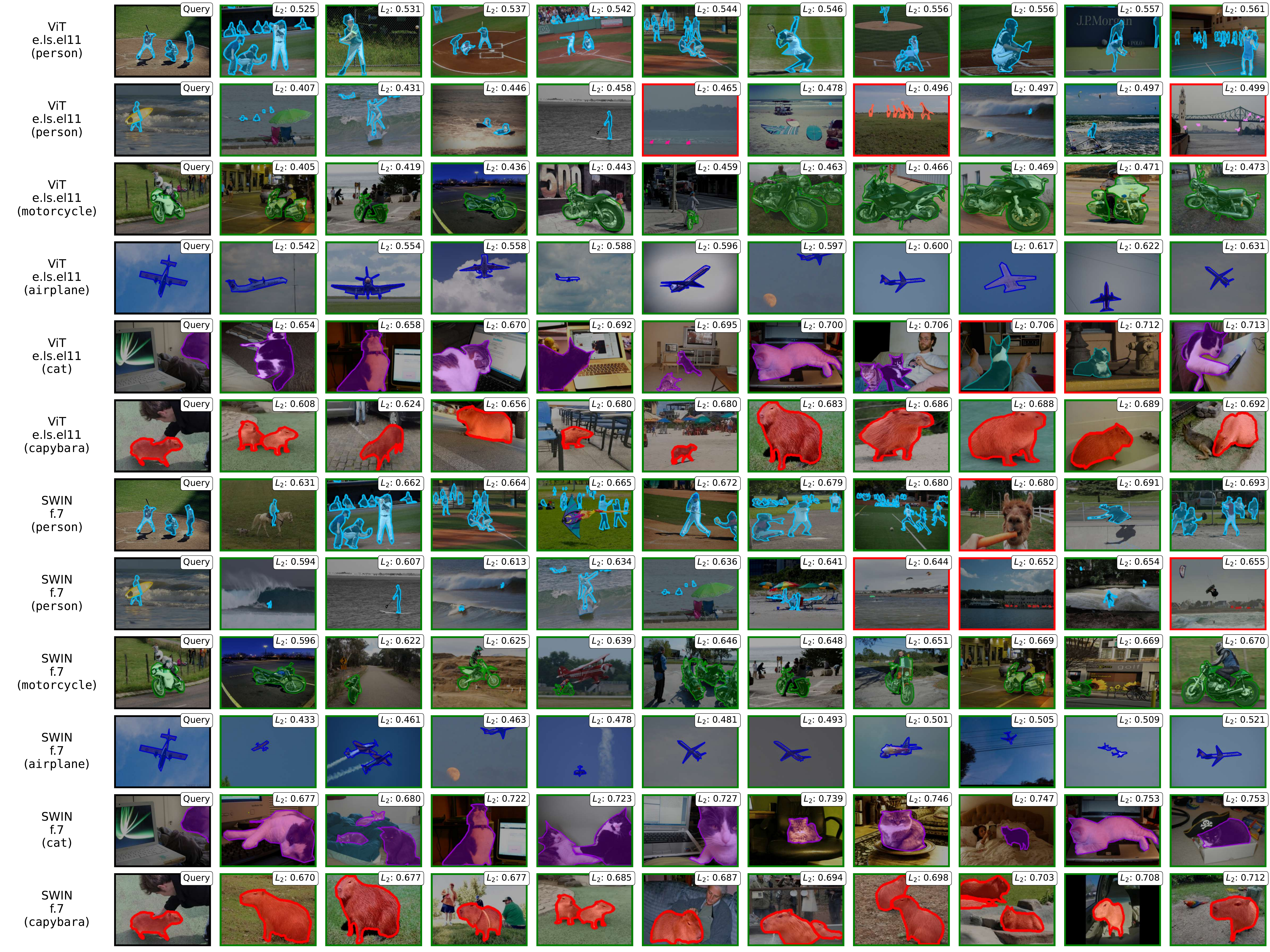}
  \caption{MS COCO \& Capybara Dataset: Top 10 retrieval results according to $L_2$-distance (\emph{columns}) for random \locVect{} queries (\emph{black frame, leftmost column}) of different categories. \emph{Green frame} - relevant sample. \emph{Red frame} - irrelevant sample. Unique concepts are color-coded.}
  \label{fig:retrieval-qualitative-extra2}
\end{figure*}


\begin{table*}
\centering
\fontsize{8pt}{10pt}\selectfont
\setlength{\tabcolsep}{5pt} 
\caption{MS COCO: Information retrieval performance in mAP@k of \locVects{} aggregated across different models and concept(-categories) (\emph{rows}), layers of increasing depth and number of samples $k$ (\emph{columns}). Results are averaged across supercategories (\ConceptTerm{person}, \ConceptTerm{vehicle}, \ConceptTerm{animal}) and all tested categories (\ConceptTerm{overall}). Models and layers are chosen as described in \cref{tab:layers}. Find graphical visualizations of the results (\ConceptTerm{overall}) in \cref{fig:retrieval-mAP-plot}.}
\label{tab:retrieval-map-all}
\begin{tabular}{@{}|c|c|cccc|cccc|cccc|@{}}
        \hline

        \multirow{2}{*}{\textbf{Model}} & \textbf{Category}  & \multicolumn{4}{c|}{\textbf{mAP in} $l_{mid}$} & \multicolumn{4}{c|}{\textbf{mAP in} $l_{deep}$} & \multicolumn{4}{c|}{\textbf{mAP in} $l_{last}$} \\
                                  & \textbf{(samples)}   &   \textbf{@5}    &   \textbf{@10}   &   \textbf{@20}   &   \textbf{@40}   &   \textbf{@5}    &   \textbf{@10}   &   \textbf{@20}   &   \textbf{@40}   &   \textbf{@5}    &   \textbf{@10}   &   \textbf{@20}   &   \textbf{@40}   \\
        \hline
        \multirow{4}{*}{YOLOv5}        & person  (500)     & 0.76  & 0.74  & 0.71  & 0.67  & 0.86  & 0.83  & 0.80  & 0.76  & 0.68  & 0.65  & 0.60  & 0.54 \\
                                       & vehicle (1622)    & 0.51  & 0.48  & 0.45  & 0.41  & 0.62  & 0.59  & 0.55  & 0.50  & 0.53  & 0.50  & 0.46  & 0.42 \\
                                       & animal  (1090)    & 0.43  & 0.39  & 0.35  & 0.29  & 0.60  & 0.55  & 0.49  & 0.41  & 0.46  & 0.41  & 0.36  & 0.29 \\
                                       & overall (3212)    & 0.52  & 0.49  & 0.46  & 0.41  & 0.65  & 0.61  & 0.57  & 0.51  & 0.53  & 0.49  & 0.45  & 0.40 \\
        \hline
        \multirow{4}{*}{MobileNetV3}   & person  (500)     & 0.41  & 0.38  & 0.35  & 0.32  & 0.40  & 0.38  & 0.36  & 0.33  & 0.60  & 0.58  & 0.53  & 0.49 \\
                                       & vehicle (1622)    & 0.30  & 0.28  & 0.26  & 0.24  & 0.31  & 0.30  & 0.28  & 0.25  & 0.37  & 0.36  & 0.33  & 0.30 \\
                                       & animal  (1090)    & 0.27  & 0.24  & 0.22  & 0.19  & 0.28  & 0.26  & 0.23  & 0.20  & 0.36  & 0.32  & 0.29  & 0.25 \\
                                       & overall (3212)    & 0.31  & 0.28  & 0.26  & 0.24  & 0.31  & 0.30  & 0.27  & 0.25  & 0.40  & 0.38  & 0.35  & 0.31 \\
        \hline
        \multirow{4}{*}{EfficientNet}  & person  (500)     & 0.46  & 0.44  & 0.41  & 0.38  & 0.69  & 0.67  & 0.63  & 0.59  & 0.77  & 0.77  & 0.74  & 0.72 \\
                                       & vehicle (1622)    & 0.34  & 0.32  & 0.29  & 0.27  & 0.43  & 0.41  & 0.38  & 0.35  & 0.54  & 0.52  & 0.50  & 0.46 \\
                                       & animal  (1090)    & 0.30  & 0.27  & 0.24  & 0.21  & 0.45  & 0.41  & 0.37  & 0.32  & 0.65  & 0.62  & 0.58  & 0.52 \\
                                       & overall (3212)    & 0.34  & 0.32  & 0.30  & 0.27  & 0.48  & 0.45  & 0.42  & 0.38  & 0.61  & 0.59  & 0.56  & 0.52 \\
        \hline
        \multirow{4}{*}{DETR}          & person  (500)     & 0.87  & 0.84  & 0.81  & 0.77  & 0.92  & 0.90  & 0.87  & 0.84  & 0.57  & 0.51  & 0.46  & 0.41 \\
                                       & vehicle (1622)    & 0.56  & 0.54  & 0.50  & 0.46  & 0.68  & 0.65  & 0.61  & 0.56  & 0.39  & 0.36  & 0.33  & 0.30 \\
                                       & animal  (1090)    & 0.48  & 0.44  & 0.39  & 0.34  & 0.71  & 0.66  & 0.61  & 0.51  & 0.32  & 0.28  & 0.24  & 0.20 \\
                                       & overall (3212)    & 0.58  & 0.55  & 0.51  & 0.46  & 0.73  & 0.70  & 0.65  & 0.59  & 0.39  & 0.36  & 0.32  & 0.28 \\
        \hline
        \multirow{4}{*}{ViT}           & person  (500)     & 0.35  & 0.33  & 0.30  & 0.28  & 0.61  & 0.58  & 0.55  & 0.50  & 0.71  & 0.69  & 0.67  & 0.64 \\
                                       & vehicle (1622)    & 0.28  & 0.27  & 0.25  & 0.23  & 0.38  & 0.37  & 0.34  & 0.32  & 0.51  & 0.49  & 0.47  & 0.44 \\
                                       & animal  (1090)    & 0.24  & 0.22  & 0.19  & 0.16  & 0.36  & 0.33  & 0.28  & 0.24  & 0.57  & 0.54  & 0.50  & 0.44 \\
                                       & overall (3212)    & 0.28  & 0.26  & 0.24  & 0.22  & 0.41  & 0.39  & 0.35  & 0.32  & 0.56  & 0.54  & 0.51  & 0.47 \\
        \hline
        \multirow{4}{*}{SWIN}          & person  (500)     &    ---     &    ---     &    ---     &    ---     & 0.80  & 0.78  & 0.75  & 0.71  & 0.76  & 0.72  & 0.68  & 0.62 \\
                                       & vehicle (1622)    &    ---     &    ---     &    ---     &    ---     & 0.53  & 0.50  & 0.47  & 0.43  & 0.65  & 0.63  & 0.61  & 0.58 \\
                                       & animal  (1090)    &    ---     &    ---     &    ---     &    ---     & 0.55  & 0.52  & 0.47  & 0.41  & 0.74  & 0.71  & 0.67  & 0.60 \\
                                       & overall (3212)    &    ---     &    ---     &    ---     &    ---     & 0.58  & 0.55  & 0.51  & 0.47  & 0.70  & 0.67  & 0.64  & 0.60 \\
        \hline
\end{tabular}
\end{table*}

\begin{figure*}
  \centering
  \includegraphics[width=\linewidth]{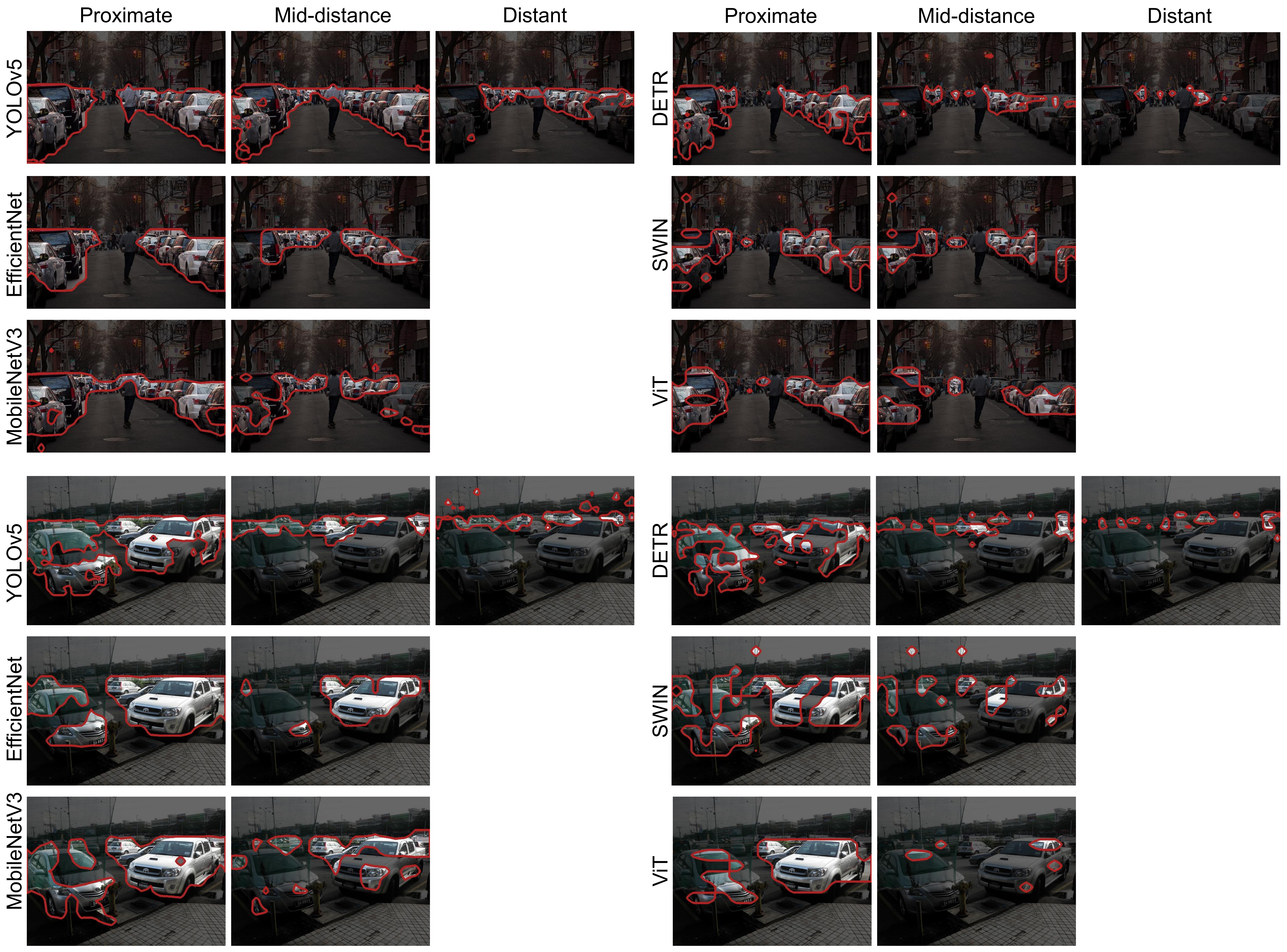}
  \caption{MS COCO: Detected distance sub-concepts \ConceptTerm{car} (\emph{columns}) in different networks (\emph{rows}). Sub-concept segmentations were obtained by projecting activations with sub-concept cluster centroids (\subglobVects, \cref{eq:generalization-centroid}). Sub-concept clusters were discovered in \locVect{} dendrograms using manual thresholding in \textbf{bold layers} of \cref{tab:layers}.}
  \label{fig:subconcepts-extra}
\end{figure*}

\begin{table*}
    \centering
    \fontsize{8pt}{10pt}\selectfont
    \setlength{\tabcolsep}{5pt} 
    \caption{MS COCO \& Capybara Dataset: Absolute Concept Separation (one-vs-all) results. Layers are chosen as described in \cref{tab:layers}.}
    \label{tab:separation}
        \begin{tabular}{@{}|c|c|ccc|ccc|@{}}
            \hline
            \multirow{2}{*}{\textbf{Dataset}} & \multirow{2}{*}{\textbf{Category}} & \textbf{YOLOv5} & \textbf{MobileNetV3} & \textbf{EfficientNet} & \textbf{DETR} & \textbf{ViT} & \textbf{SWIN} \\
            & & $l_{\text{deep}}$ & $l_{\text{last}}$ & $l_{\text{last}}$ & $l_{\text{deep}}$ & $l_{\text{last}}$ & $l_{\text{last}}$ \\
            \hline
\multirow{19}{*}{MS COCO}            
            & person & 0.23 & 0.30 & 0.24 & 0.39 & 0.23 & 0.20 \\
            & bicycle & 0.29 & 0.49 & 0.31 & 0.56 & 0.37 & 0.40 \\
            & car & 0.08 & 0.05 & 0.06 & 0.14 & 0.06 & 0.12 \\
            & motorcycle & 0.26 & 0.34 & 0.24 & 0.44 & 0.24 & 0.34 \\
            & airplane & 0.39 & 0.66 & 0.59 & 0.52 & 0.42 & 0.42 \\
            & bus & 0.08 & 0.07 & 0.15 & 0.11 & 0.14 & 0.15 \\
            & train & 0.16 & 0.18 & 0.23 & 0.23 & 0.14 & 0.25 \\
            & truck & 0.07 & 0.04 & 0.05 & 0.11 & 0.05 & 0.12 \\
            & boat & 0.16 & 0.31 & 0.19 & 0.53 & 0.31 & 0.28 \\
            & bird & 0.34 & 0.51 & 0.47 & 0.49 & 0.32 & 0.49 \\
            & cat & 0.14 & 0.14 & 0.07 & 0.22 & 0.09 & 0.09 \\
            & dog & 0.14 & 0.14 & 0.07 & 0.21 & 0.08 & 0.09 \\
            & horse & 0.11 & 0.31 & 0.39 & 0.41 & 0.25 & 0.23 \\
            & sheep & 0.11 & 0.42 & 0.38 & 0.59 & 0.38 & 0.33 \\
            & cow & 0.11 & 0.48 & 0.38 & 0.50 & 0.38 & 0.23 \\
            & elephant & 0.10 & 0.31 & 0.20 & 0.56 & 0.38 & 0.29 \\
            & bear & 0.53 & 0.65 & 0.73 & 0.59 & 0.60 & 0.73 \\
            & zebra & 0.56 & 0.67 & 0.78 & 0.73 & 0.36 & 0.62 \\
            & giraffe & 0.38 & 0.65 & 0.63 & 0.66 & 0.46 & 0.58 \\
            \hline
Capybara    
            & capybara & 0.35 & 0.65 & 0.74 & 0.60 & 0.42 & 0.66 \\
            \hline
            
            \hline
        \end{tabular}
\end{table*}

\begin{figure*}
  \centering
  \includegraphics[width=\linewidth]{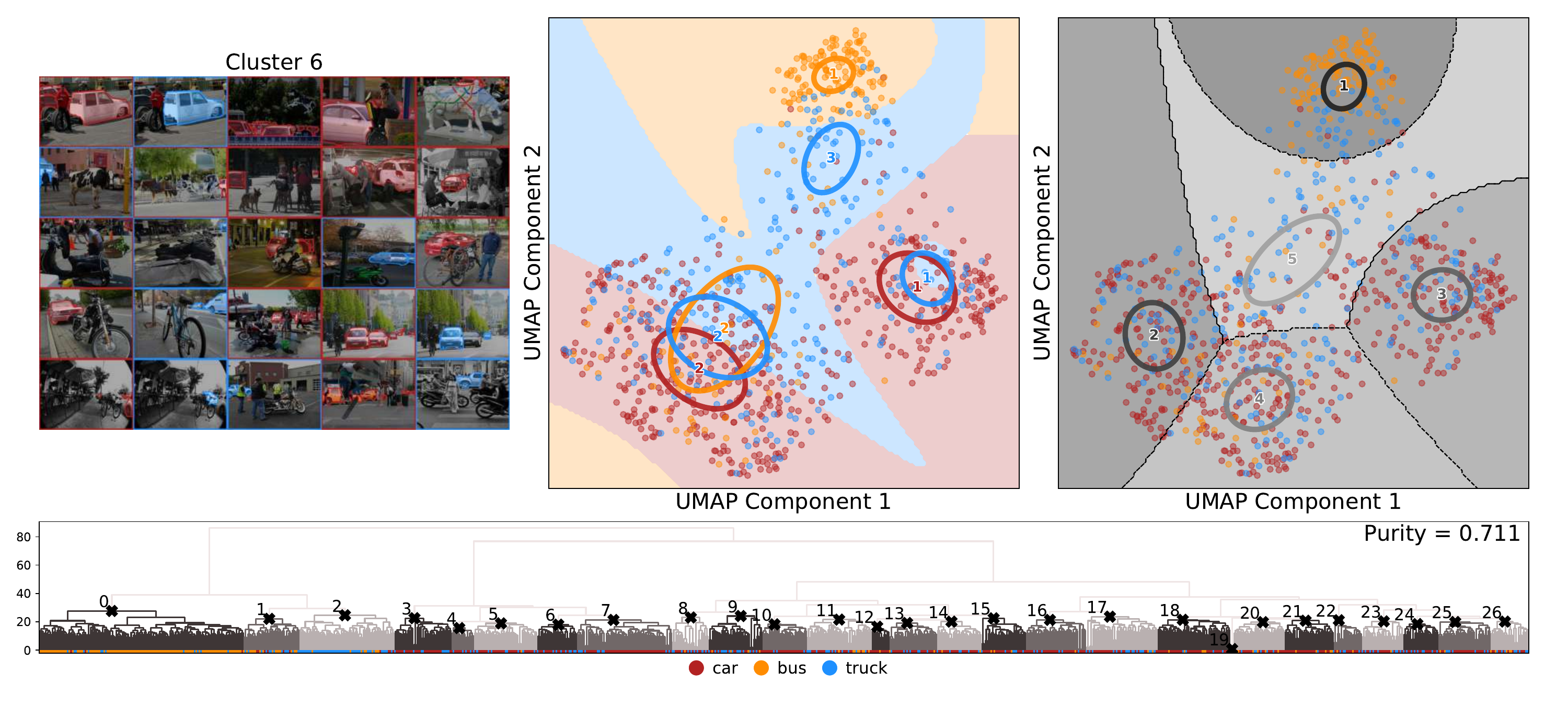}
  \caption{MS COCO: Concept confusion detected with \ConceptTerm{car}, \ConceptTerm{truck}, and \ConceptTerm{bus} \locVects{} in \texttt{features.7.0} of EfficientNet: content of a selected cluster of dendrogram (\emph{top-left}), GMMs fitted for 2D UMAP-reduced \locVect{} samples with regard to their labels (\emph{top-middle}), GMMs fitted for 2D UMAP-reduced \locVect{} samples regardless of their labels (\emph{top-right}), and \locVects{} dendrogram with clusters identified with \cref{alg:adaptive-clustering} (\emph{bottom}).}
  \label{fig:confusion-efficientnet}
\end{figure*}

\begin{figure*}
  \centering
  \includegraphics[width=\linewidth]{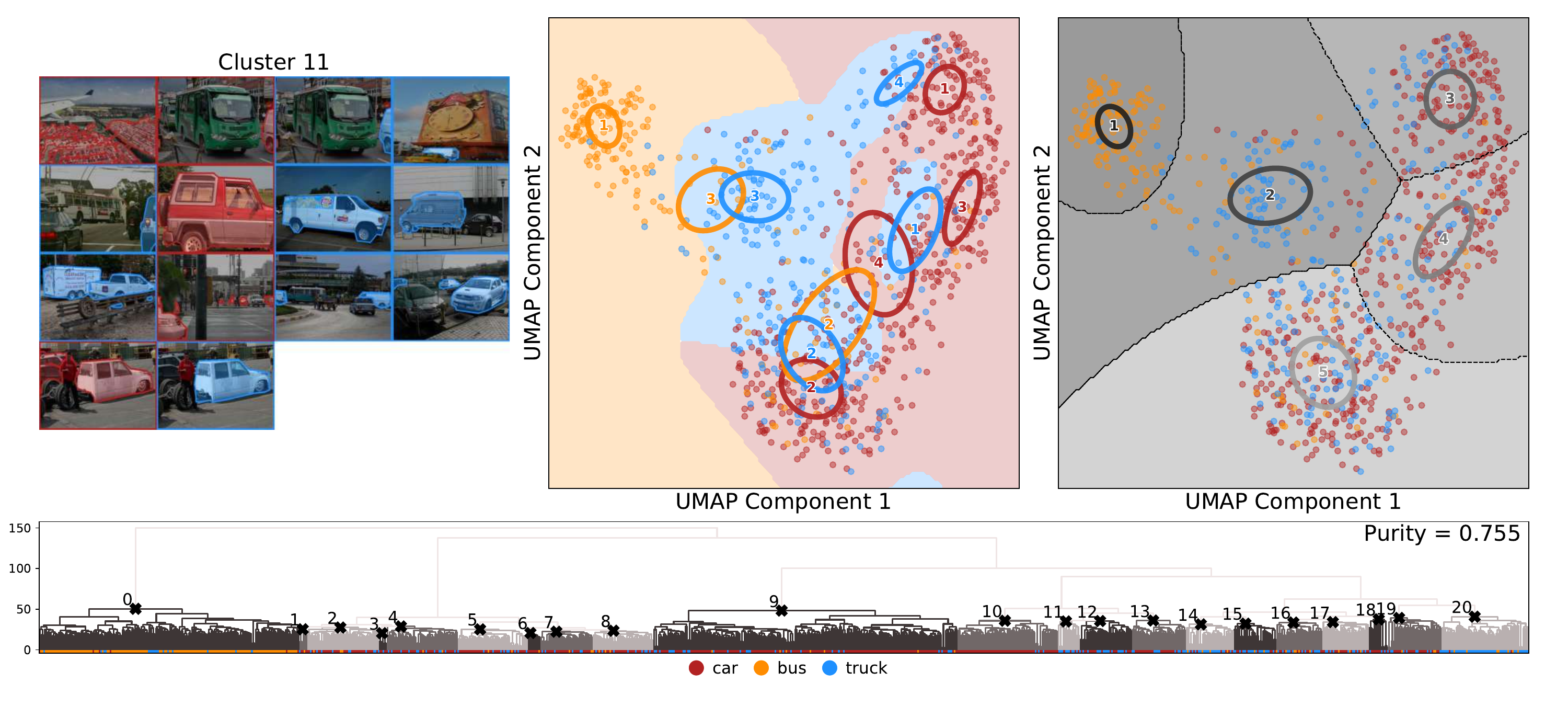}
  \caption{Concept confusion detected with \ConceptTerm{car}, \ConceptTerm{truck}, and \ConceptTerm{bus} \locVects{} in \texttt{features.7.0.0} of SWIN: content of a selected cluster of dendrogram (\emph{top-left}), GMMs fitted for 2D UMAP-reduced \locVect{} samples with regard to their labels (\emph{top-middle}), GMMs fitted for 2D UMAP-reduced \locVect{} samples regardless of their labels (\emph{top-right}), and \locVects{} dendrogram with clusters identified with \cref{alg:adaptive-clustering} (\emph{bottom}).}
  \label{fig:confusion-swin}
\end{figure*}

\begin{figure*}
  \centering
  \includegraphics[width=\linewidth]{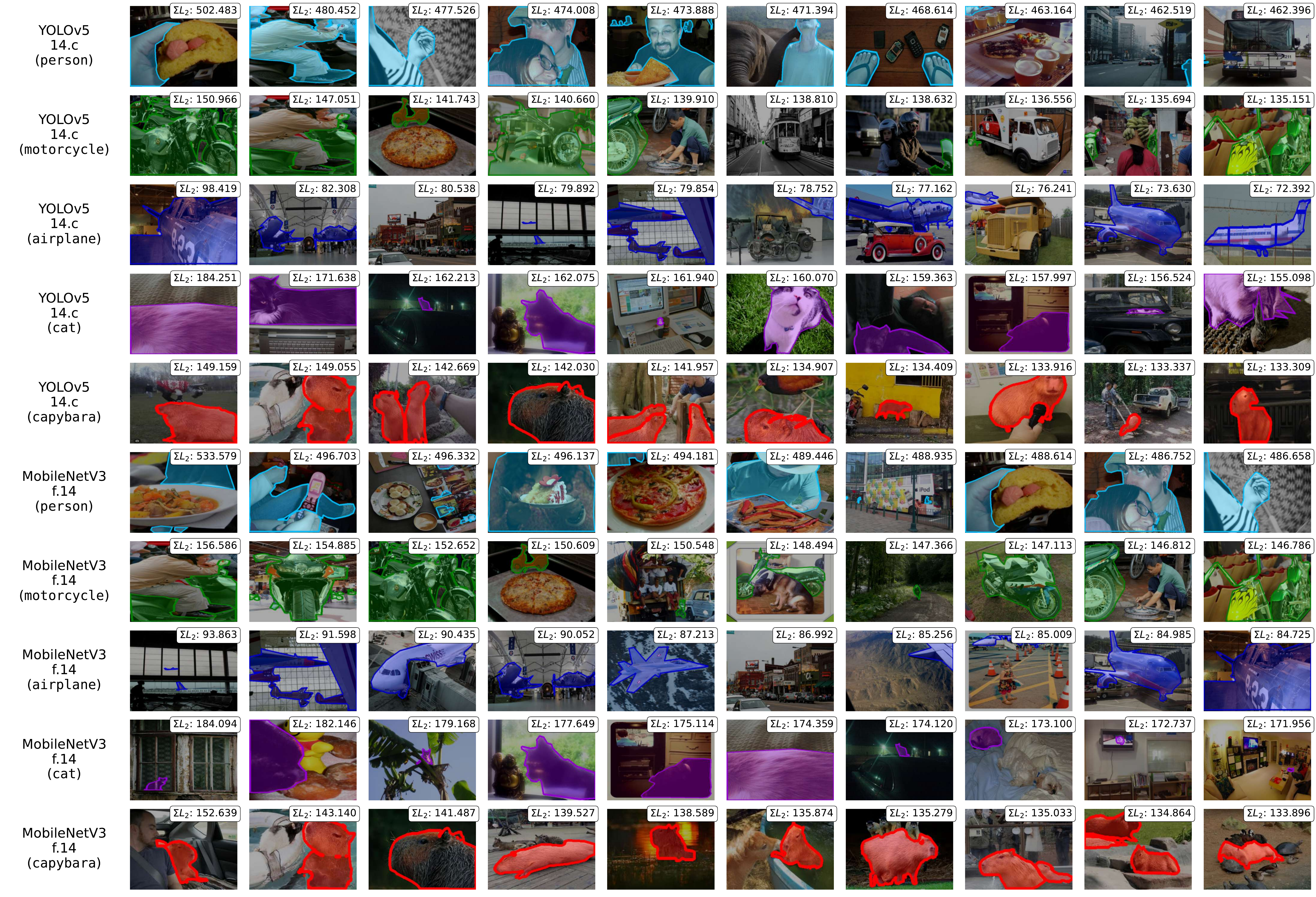}
  \caption{MS COCO \& Capybara Dataset: Top 10 outliers according to $\Sigma L_2$-distance (\emph{columns}) for one layer of models YOLOv5 and MobileNetV3 and for different concepts (\emph{rows}). Unique concepts are color-coded.}
  \label{fig:outliers-plot-all-extra1}
\end{figure*}

\begin{figure*}
  \centering
  \includegraphics[width=\linewidth]{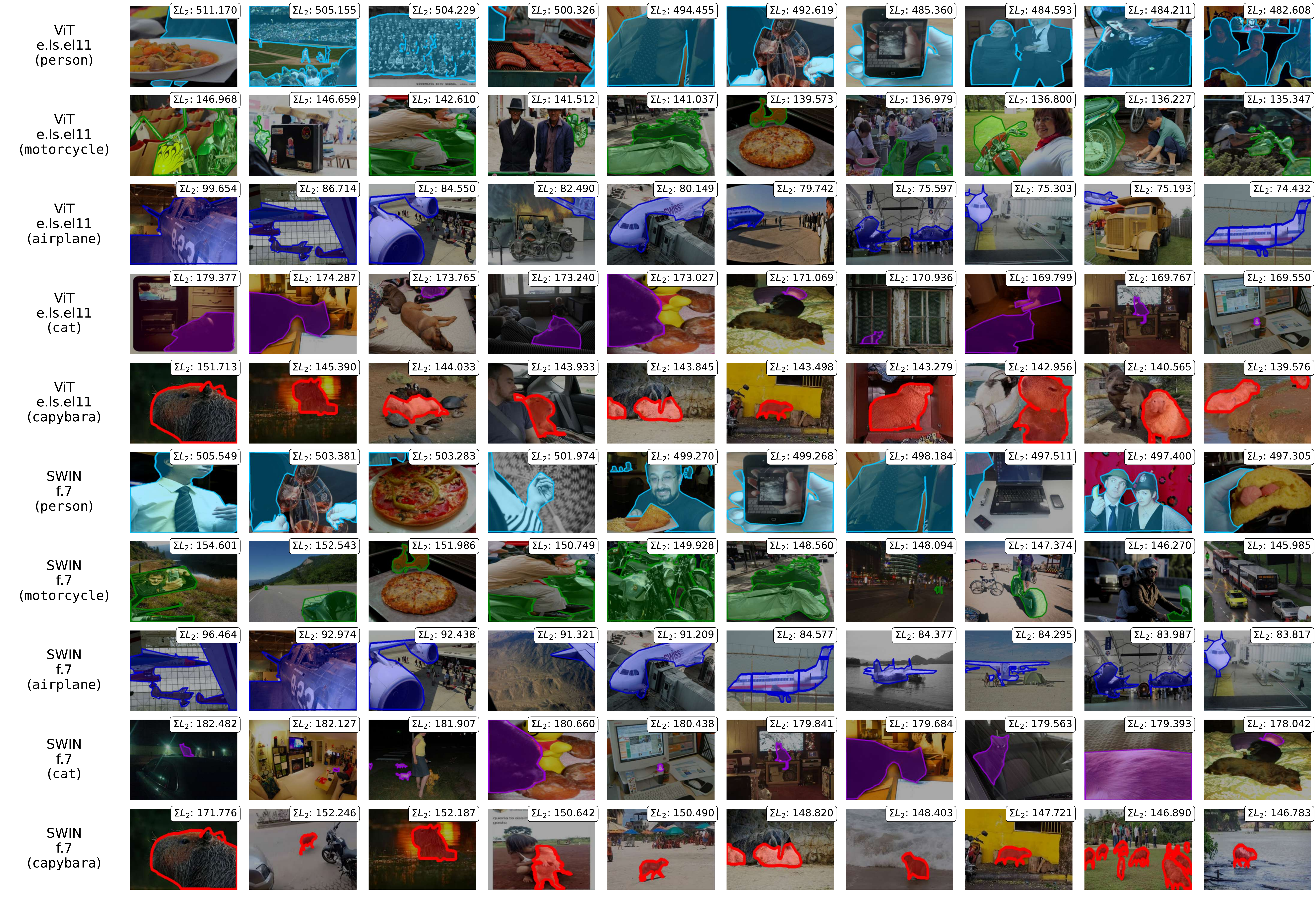}
  \caption{MS COCO \& Capybara Dataset: Top 10 outliers according to $\Sigma L_2$-distance (\emph{columns}) for one layer of classifiers ViT and SWIN and for different concepts (\emph{rows}). Unique concepts are color-coded.}
  \label{fig:outliers-plot-all-extra2}
\end{figure*}

\end{document}